%% file: main.tex
\documentclass{article}
\PassOptionsToPackage{dvipsnames}{xcolor}
\IfFileExists{iclr2027_conference.sty}%
  {\usepackage{iclr2027_conference,times}}%
  {\usepackage{iclr_fallback,times}}

\usepackage[utf8]{inputenc} % allow utf-8 input
\usepackage[T1]{fontenc}    % use 8-bit T1 fonts
\usepackage{microtype}      % microtypography

\usepackage{amsmath,amssymb,amsfonts}
\usepackage{bm}
\usepackage{bbm}
\usepackage{nicefrac}       % compact symbols for 1/2, etc.

\usepackage{xcolor}
\usepackage{colortbl}

\usepackage{booktabs}
\usepackage{graphicx}
\usepackage{multirow}
\usepackage{adjustbox}
\usepackage{caption}
\usepackage{subcaption}
\usepackage{tabularx}
\usepackage{makecell}
\usepackage{array}

\usepackage{enumitem}
\usepackage{gensymb}
\usepackage{marvosym}
\usepackage[normalem]{ulem} % \sout used by review macros below
\usepackage{algorithm,algorithmic}
\makeatletter
\renewcommand\fs@ruled{%
  \def\@fs@cfont{\bfseries}\let\@fs@capt\floatc@ruled
  \def\@fs@pre{{\color{black}\hrule height.8pt depth0pt}\kern2pt}%
  \def\@fs@post{\kern2pt{\color{black}\hrule}\relax}%
  \def\@fs@mid{\kern2pt{\color{black}\hrule}\kern2pt}%
  \let\@fs@iftopcapt\iftrue}
\makeatother
\floatstyle{ruled}\restylefloat{algorithm}
\usepackage{hyperref}
\usepackage{url}
\hypersetup{colorlinks=true}
\hypersetup{linktoc=all}
\hypersetup{citecolor=customblue}
\hypersetup{linkcolor=crimson}
\hypersetup{urlcolor=MidnightBlue}
\usepackage[all]{hypcap}
\usepackage[nameinlink]{cleveref}

\definecolor{Red}{rgb}{1,0,0}
\definecolor{Blue}{rgb}{0,0,0.8}
\definecolor{Green}{rgb}{0,0.7,0.2}
\definecolor{airforceblue}{rgb}{0.36, 0.54, 0.66}
\definecolor{ao(english)}{rgb}{0.0, 0.5, 0.0}
\definecolor{azure(colorwheel)}{rgb}{0.0, 0.5, 1.0}
\definecolor{crimson}{rgb}{0.86, 0.08, 0.24}
\definecolor{darkcerulean}{rgb}{0.03, 0.27, 0.49}
\definecolor{cobalt}{rgb}{0.0, 0.28, 0.67}
\definecolor{rosegold}{rgb}{0.72, 0.43, 0.47}
\definecolor{orange-red}{rgb}{1.0, 0.27, 0.0}
\definecolor{mountainmeadow}{rgb}{0.19, 0.73, 0.56}
\definecolor{malachite}{rgb}{0.04, 0.85, 0.32}
\definecolor{darkblue}{rgb}{0.0, 0.0, 0.55}
\definecolor{customblue}{rgb}{0.2, 0.35, 0.8}
\definecolor{gg}{gray}{0.9}

\renewcommand{\algorithmiccomment}[1]{\bgroup\hfill$\triangleright$~#1\egroup}

\makeatletter
\def\hlinewd#1{%
	\noalign{\ifnum0=`}\fi\hrule \@height #1 %
	\futurelet\reserved@a\@xhline}
\makeatother

\creflabelformat{equation}{#2\textup{#1}#3}
\crefname{assumption}{assumption}{assumptions}
\crefname{algorithm}{Algorithm}{Algorithms}
\Crefname{algorithm}{Algorithm}{Algorithms}

\newcommand{\eat}[1]{{}}

\definecolor{keyrow}{gray}{0.92}
\newcommand{\hi}{\rowcolor{keyrow}}
\newcommand{\snapkv}{SnapKV}
\newcommand{\dg}{\textsc{Dge}} % Draft-Guided Eviction; final name TBD
\newcommand{\betadec}{\beta_{\mathrm{dec}}}

\title{When to Evict, Not What to Keep:\\
Draft-Guided Eviction\\
for Training-Free KV-Cache Compression}

\author{%
\mdseries % the style wraps the author block in \bf; reset to medium weight
\begin{tabular}[t]{@{}l@{\hspace{5em}}l@{}}
Haeyong Kang\thanks{Corresponding author: hykang@duksung.ac.kr.} & Chang D. Yoo \\
Duksung Women's University & KAIST \\ & \\
\end{tabular}}

\iclrfinalcopy % arXiv/camera-ready: hides line numbers, shows authors

\begin{document}

\maketitle
% The style sets the running head inside \@maketitle's group, which does not
% survive with newer fancyhdr versions - reassert it globally here.
% Keep the year in sync with iclr2027_conference.sty's own \lhead.
\makeatletter
\@ifundefined{ificlrfinal}{}{%
  \ificlrfinal
    \fancyhead[L]{Preprint. Under review.}%
  \else
    \fancyhead[L]{Under review as a conference paper at ICLR 2027}%
  \fi
  \thispagestyle{fancy}}
\makeatother

\input{sections/abstract}
\input{sections/introduction}

\input{sections/related_work}
\input{sections/setup}
\input{sections/audit}

\input{sections/method}

\input{sections/experiments}
\input{sections/discussion}
\input{sections/conclusion}

% arXiv: no page limit, so the statements and references follow the body
% directly (the ICLR submission used \clearpage here to keep page 9 body-only).
\input{sections/statements}

\bibliography{references}
\IfFileExists{iclr2027_conference.bst}%
  {\bibliographystyle{iclr2027_conference}}%
  {\bibliographystyle{plainnat}}

\appendix
% Appendix-only float packing: the ICLR style sets \topfraction/\textfraction
% but leaves \bottomfraction at 0.3 and \floatpagefraction at 0.5, so wide
% tables cannot sit at the bottom of a page and leave large gaps. Scoped here
% so the 9-page body keeps its layout.
\renewcommand{\bottomfraction}{0.7}
\renewcommand{\floatpagefraction}{0.85}
\setcounter{topnumber}{3}
\setcounter{bottomnumber}{2}
\setcounter{totalnumber}{5}
\input{sections/appendix}

\end{document}

%% file: sections/abstract.tex
\begin{abstract}
Training-free KV-cache compression methods such as \snapkv{}, H2O, and PyramidKV evict tokens at the end of prefill, aiming to preserve the attention mass that future queries are expected to use---optimizing \emph{what} to keep. We show that this objective fails in two distinct ways. \textbf{(1) Compensation:} restoring the evicted attention mass can recover the attention-level target without recovering task quality. \textbf{(2) Selection:} covering more of the true decode-query mass can \emph{hurt} quality when the recovered mass is fragmented rather than concentrated in coherent spans. These failures share a common cause: eviction occurs before the queries that determine the answer trajectory exist. We propose \textbf{D}raft-\textbf{G}uided \textbf{E}viction (\dg{}), which defers eviction until after drafting the first $k{=}2$ answer tokens using the full cache---just one decode step beyond prefill. Because the draft is generated from the answer's own prefix, no cache entries are discarded before this trajectory signal becomes available. The per-head cache budget remains unchanged, and \dg{} can be applied directly to \snapkv{}, PyramidKV, H2O, and StreamingLLM without modifying their eviction scores. Unlike extra-pass methods, \dg{} changes \emph{when} eviction occurs rather than \emph{what} cache entries are selected. Extensive experiments demonstrate that \dg{} outperforms prior methods at every evaluated budget on five of six instruct-tuned backbones, achieving $44.2$ on LongBench, nearly matching FullKV at $44.3$. The timing-only control \dg{}-W achieves the same score, demonstrating that the gain comes from \emph{when} eviction occurs rather than \emph{what} is selected---an effect we term \emph{trajectory anchoring}.
\end{abstract}

%% file: sections/introduction.tex
\section{Introduction}
\label{sec:intro}

Serving long-context LLMs is a memory problem before it is a compute problem: the KV-Cache grows linearly with context length and batch size---roughly $16$\,GiB for an 8B model at 128K tokens, on par with the weights themselves---and every decode step reads the cache back. Training-free eviction is an attractive deployment strategy: keep a small per-head budget of $B$ past tokens, discard the rest, and change nothing else about the model. A productive line of work has refined \emph{which} tokens to keep---StreamingLLM keeps attention sinks and recency \citep{xiao2024streamingllm}, H2O accumulates attention mass \citep{zhang2023h2o}, \snapkv{} max-pools the scores of a late-prompt observation window \citep{li2024snapkv}, and PyramidKV schedules the budget across layers \citep{cai2024pyramidkv}---but these methods share one recipe: when prefill ends, score the past tokens, keep the top-$B$ per head, and evict the rest, permanently and \emph{before the first answer token is generated}.

\input{sections/figure_concept}

Two choices distinguish these training-free regimes: \emph{which} queries score the cache, and \emph{when} the cut fires. FullKV in \Cref{fig:concept}(a) uses the real decode queries and never evicts; prefill-time eviction in \Cref{fig:concept}(b) cuts before generation using only prefill-time scores; Lookahead Q-Cache (LAQ) in \Cref{fig:concept}(c) also cuts at prefill, but replaces future decode queries with synthesized \emph{pseudo}-queries \citep{ge2025lookahead}. Our \textbf{D}raft-\textbf{G}uided \textbf{E}viction (\dg{}), in \Cref{fig:concept}(d), instead drafts the first $k$ answer tokens on the full cache and evicts only afterward. It therefore observes the \emph{real} decode queries before making the irreversible cut: \dg{} changes \emph{when} the cut happens, rather than \emph{what} scores the cache.

Despite these differences, existing approaches optimize a common implicit objective: preserve the attention mass that future computation would place on the discarded tokens. \emph{Selection} maximizes captured mass, as in \snapkv{} and H2O \citep{li2024snapkv,zhang2023h2o}; \emph{compensation} attempts to restore what selection loses through merging entries or fitting biases and values \citep{zhang2024cam,wan2024lookm,zweiger2026fast}; and \emph{allocation} redistributes the budget across heads and layers \citep{feng2024adakv,cai2024pyramidkv}. This objective is plausible, and its attention-level gains are real. However, whether increasing attention-mass preservation reliably translates into task quality has not been systematically examined under controlled dosage.

Our controlled observations reveal two limitations. \textbf{(1) Compensation:} mass is fungible, but content is not. Restoring evicted attention mass through merging, value absorption, or per-token correction does not necessarily recover the information carried by the original tokens. \textbf{(2) Selection:} more captured attention mass does not necessarily improve task performance. When retained mass is fragmented rather than concentrated in coherent spans, an extractive answer can lose the information needed for reconstruction. We establish both effects under controlled dosage in \S\ref{sec:audit}, with the corresponding sweeps in \Cref{fig:headline}(a). Both limitations arise because eviction occurs before the queries that actually consume the cache exist: prefill-time signals can only estimate future decode queries, however they are constructed.

We propose \dg{}, which changes \emph{when} eviction occurs rather than \emph{what} is kept. Prior methods improve \emph{what} a prefill-time cut keeps, while leaving the cut at the end of prefill \citep{kim2025kvzip,zweiger2026fast,ge2025lookahead}. \dg{} instead drafts the first $k$ answer tokens on the full cache and evicts only afterward, allowing the decision to use real decode queries rather than estimates. The draft is the answer's own prefix, not an auxiliary pass, so the per-head budget is unchanged and eviction leaves the prefill critical path. A single decode step is often sufficient, and \S\ref{sec:experiments} shows that the deferral itself, rather than a sharper selection signal, drives the gain. We attribute this to what we call \emph{trajectory anchoring}: the initial answer tokens, produced on the full cache, fix the model's decoding trajectory before compression, and the compressed cache then carries the rest, as \Cref{fig:headline}(b) shows.

\paragraph{Contributions.}
\begin{itemize}[leftmargin=*]
\small
\item \textbf{Draft-guided eviction.} We introduce \dg{}, a training-free KV-cache compression framework that opens a new \emph{when}-to-evict axis: the model first drafts $k$ answer tokens with the full cache and evicts only afterward, keeping the per-head budget and leaving the scoring rule a free slot---the base evictor's own (the timing-only control \dg{}-W) or the draft's real queries (the default).

\item \textbf{Controlled analysis of attention-based eviction.} We identify two failure modes of optimizing attention mass---\emph{compensation} and \emph{selection}---and show that attention-level preservation can decouple from downstream task quality under controlled dosage.

\item \textbf{Empirical validation.} Across six instruct-tuned backbones from 3B to 14B parameters, \dg{} outperforms prior methods at every evaluated budget on five backbones and achieves an average LongBench score of $44.2$, compared with $40.1$ for LAQ and $35.3$ for \snapkv{}, within $0.2$ of uncompressed FullKV. The same deferral lifts PyramidKV, H2O, and StreamingLLM, while \dg{}-W also reaches $44.2$, isolating eviction timing as the source of the gain.
\end{itemize}

\input{sections/figure_headline}

%% file: sections/figure_concept.tex
% figure_concept: body figure, kept in its own file like the tables; edit here.
\begin{figure}[!t]
\centering
\captionsetup[subfigure]{skip=1pt}
\begin{subfigure}[b]{0.49\textwidth}
  \centering\includegraphics[width=\textwidth]{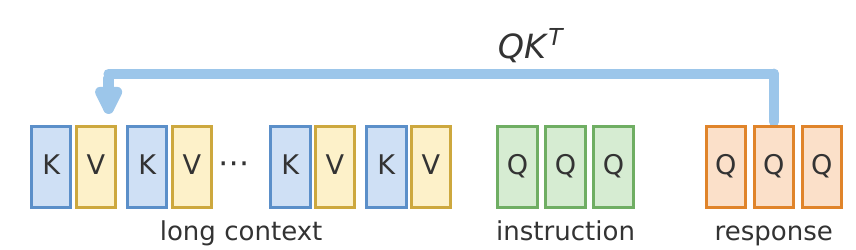}
  \caption{FullKV: real queries, no cut.}
  \label{fig:concept-a}
\end{subfigure}
\hfill
\begin{subfigure}[b]{0.49\textwidth}
  \centering\includegraphics[width=\textwidth]{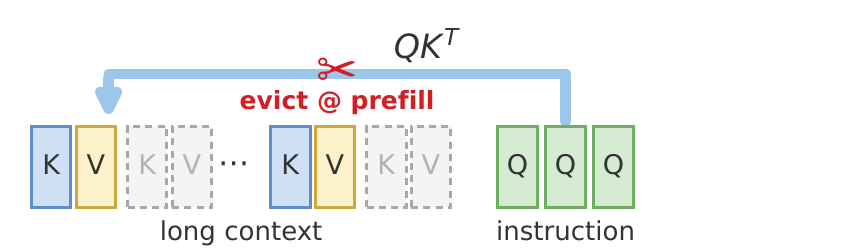}
  \caption{Prefill eviction: window scores.}
  \label{fig:concept-b}
\end{subfigure}

\begin{subfigure}[b]{0.49\textwidth}
  \centering\includegraphics[width=\textwidth]{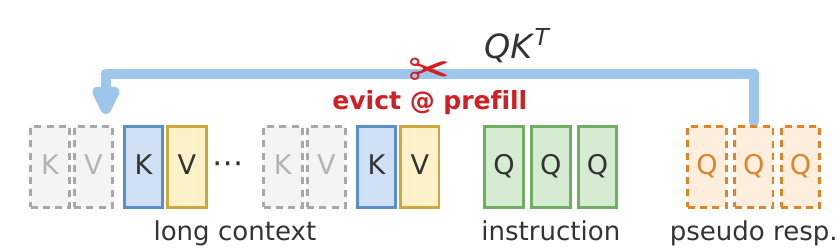}
  \caption{Lookahead: prefill, pseudo queries.}
  \label{fig:concept-c}
\end{subfigure}
\hfill
\begin{subfigure}[b]{0.49\textwidth}
  \centering\includegraphics[width=\textwidth]{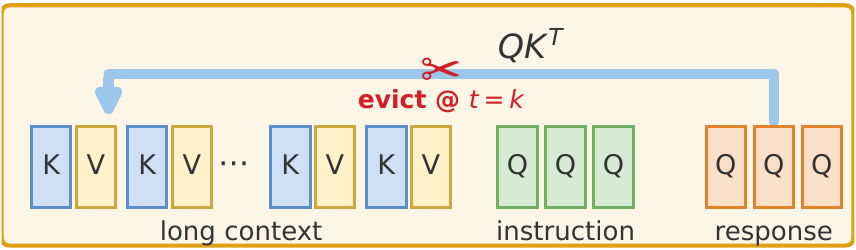}
  \caption{\textbf{\dg{}, ours}: defer the cut to $t{=}k$.}
  \label{fig:concept-d}
\end{subfigure}
\caption{\textbf{Four training-free KV-compression regimes:} which queries
drive the $QK^{\top}$ scoring, and when eviction fires, marked by red scissors. Only
\dg{}~(d) defers the cut; the window-score regime~(b) covers \snapkv{},
PyramidKV and their descendants.}
\label{fig:concept}
\end{figure}

%% file: sections/figure_headline.tex
% figure_headline: body figure, kept in its own file like the tables; edit here.
% Figure 2 (fig:headline): generated by paper/figures/make_fig1.py (data provenance in
% the script header); regenerate with
%   python paper/figures/make_fig1.py
\begin{figure}[!t]
\centering
\begin{subfigure}[b]{0.60\textwidth}
  \centering
  \includegraphics[width=\textwidth]{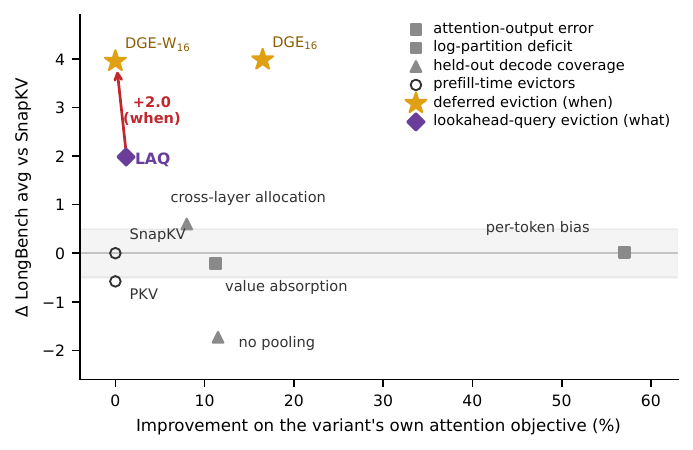}
  \caption{Moving \emph{when} beats improving \emph{what}.}
  \label{fig:headline-a}
\end{subfigure}
\hfill
\begin{subfigure}[b]{0.385\textwidth}
  \centering
  \includegraphics[width=\textwidth]{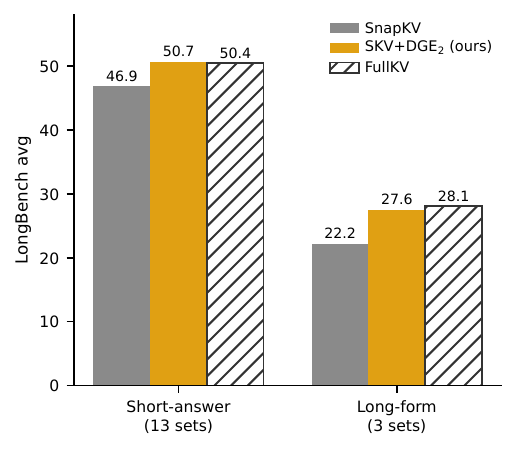}
  \caption{One decode step restores FullKV.}
  \label{fig:headline-b}
\end{subfigure}
\caption{\textbf{When beats what}, on LongBench with Llama-3.1-8B at $B{=}128$;
subscripts are the draft length $k$. \textbf{(a)}~Each \emph{what}-axis
mechanism's gain on its own attention objective against its LongBench $\Delta$
vs \snapkv{}; gray band: $\pm0.5$; hollow circles: \snapkv{} and PKV; LAQ
\citep{ge2025lookahead} at $m{=}16$. Both deferred markers defer the cut by
$16$ tokens, \dg{}-W$_{16}$ scoring with \snapkv{}'s window and \dg$_{16}$ with
the $16$ draft queries; the red arrow is deferral's gain over LAQ.
\textbf{(b)}~Short-answer ($13$ sets) vs long-form (the three summarization
sets) averages.}
\label{fig:headline}
\end{figure}

%% file: sections/related_work.tex
\section{Related Work}
\label{sec:related}

\paragraph{Selection and allocation: which tokens to keep.}
The dominant training-free family scores past tokens at the end of prefill under a per-head budget. \snapkv{} \citep{li2024snapkv} max-pools observation-window scores; H2O \citep{zhang2023h2o} and ScissorHands \citep{liu2023scissorhands} accumulate attention; FastGen \citep{ge2024fastgen} adapts budgets per head; PyramidKV \citep{cai2024pyramidkv} schedules them across layers; AdaKV \citep{feng2024adakv} and HeadKV \citep{fu2024headkv} reallocate them across heads; and StreamingLLM \citep{xiao2024streamingllm}, ThinK \citep{xu2024think}, $L_2$ selection \citep{devoto2024l2}, and Quest \citep{tang2024quest} use sinks, recency, or key statistics. These methods improve \emph{which} tokens survive, but commit before the first answer token exists. \dg{} leaves the per-head budget unchanged and can leave the scoring rule unchanged too, deferring the same selection decision until after the initial answer tokens. On \snapkv{}'s own scores, this timing change alone already outperforms the selection rules we evaluate.

\paragraph{Compensation: restoring what selection discarded.}
A second line of work attempts to repair the cut rather than improve selection. CaM \citep{zhang2024cam}, LOOK-M \citep{wan2024lookm}, KVMerger \citep{wang2024kvmerger}, and D2O \citep{wan2024d2o} merge evicted entries into retained ones, with the vision analogue ToMe \citep{bolya2023tome}; FAST \citep{zweiger2026fast} instead fits per-token biases and refits values. These methods can restore the attention mass targeted by the repair, but they cannot recover the original token-level content once it has been compressed away. Our controlled counterparts in \S\ref{sec:audit} expose this gap: attention-level compensation can improve without a corresponding gain in task quality. \dg{} avoids the repair problem altogether by generating the steering tokens while the cache is still intact, so the information needed to anchor the answer trajectory is never discarded.

\paragraph{Extra passes: sharpening the prefill-time signal.}
A third line spends additional computation to improve the eviction signal. KVzip \citep{kim2025kvzip} repeats the prefill pass; FAST, above, calibrates its biases and values on self-generated Q\&A; LAQ \citep{ge2025lookahead} synthesizes \emph{pseudo} future queries; and LookaheadKV \citep{ahn2026lookaheadkv} trains adapters to predict such scores. These methods improve \emph{what} to keep, but retain the same schedule: the cut still fires at prefill end, so the signal remains a proxy for the queries that actually consume the cache. Our experiments show that a matched LAQ run saturates with dose and, even at its best, trails a pure timing change. Other KV-efficiency methods change the representation, memory management, or decoding objective, including learned compression \citep{zhang2024beacon,nawrot2024dmc,mu2023gist}, quantization \citep{liu2024kivi,hooper2024kvquant}, PagedAttention \citep{kwon2023vllm}, InfiniGen \citep{lee2024infinigen}, and speculative decoding \citep{leviathan2023speculative}; concurrent analyses also question the attention-mass objective itself \citep{criticalkv2025,rethinking2026,fixedcontract2026}. In contrast, \dg{} changes only the \emph{timing of eviction}, using the answer's own prefix to expose real decode queries before the cut; App.~\ref{app:draftlike} contrasts it with the draft-like designs.

%% file: sections/setup.tex
\section{Preliminaries}
\label{sec:setup}

\paragraph{Notation.}
A prompt of length $L$ is prefilled; eviction keeps a per-head budget of $B$
tokens ($B{-}w$ selected past tokens plus the last $w$ window tokens,
$w{=}8$ throughout) per layer. For a future query $q$, the full attention output decomposes exactly as
\begin{equation}
O(q) \;=\; (1-\beta(q))\, O_{\mathrm{keep}}(q) \;+\; \beta(q)\,
V_{\mathrm{evict}}(q),
\qquad
\beta(q)=\textstyle\sum_{j\in E} a_j(q),
\label{eq:identity}
\end{equation}
where $E$ is the evicted set: $\beta$ is the \emph{missed mass} and
$V_{\mathrm{evict}}$ the attention-weighted centroid of evicted values.
Every compensation method estimates terms of Eq.~\ref{eq:identity}; every
selection method minimizes $\beta$.

\paragraph{Held-out instrumentation.}
Our diagnostics run the model once with the full cache, capture post-RoPE
$Q,K,V$ via forward pre-hooks, and replay \emph{real decode queries} against
candidate caches offline. Metrics, per App.~\ref{app:diagnostics} and~\ref{app:matchloss}: missed decode
mass $\betadec$ is the $\beta$ of Eq.~\ref{eq:identity} on a held-out half of
the decode steps, and a rule's \emph{held-out coverage} $\Delta$ is its
relative reduction in $\betadec$ vs \snapkv{}. Also relative attention-output
error, retained-mask run length, and a train-free attention-matching loss
$|\log Z-\log\hat Z|$, the log-partition deficit between the softmax
normalizer $Z$ over all keys and $\hat Z$ over the retained ones.

\paragraph{Evaluation protocol.}
LongBench \citep{bai2024longbench} (16 English datasets) on
Llama-3.1-8B-Instruct \citep{dubey2024llama}, budgets
$B\in\{32,64,128\}$ (larger only in the offline diagnostics), greedy decoding, the stop-token-corrected
Llama-3 chat protocol of App.~\ref{app:expdetails}; matched sample sets across methods. Backbone
checks span Mistral-7B-Instruct-v0.2 \citep{jiang2023mistral},
Qwen2.5-7B/14B-Instruct \citep{qwen2025qwen25},
Llama-3.2-3B-Instruct, and Llama-3-8B-Instruct, reported in
\Cref{tab:main} and App.~\ref{app:llama3}.
Re-runs are bit-identical, so the only uncertainty is sampling over the
$3{,}750$ examples; comparisons use a paired bootstrap, whose intervals on
differences are $\sim\pm0.3$ against $\pm1.1$ marginally, as
App.~\ref{app:ci} details.

\paragraph{Headline diagnostic.}
At a per-head budget of $B{=}128$ tokens, \snapkv{} misses
$\betadec = 16.7\%$ of decode attention mass
on average, with a median of $9.4\%$ and a $90$th percentile of $43\%$ over
examples, layers and heads, a heavy tail; the
window-query $\beta$ correlates $0.72$ with $\betadec$ ($0.68$--$0.76$
over $B{=}64$--$1024$)---the missed mass is apparent headroom that
\S\ref{sec:audit}'s analysis shows is \emph{not} convertible
into task quality by content-side means.

%% file: sections/audit.tex
\section{Attention-Mass Analysis}
\label{sec:audit}

We test the attention-mass objective under increasing dosage on each axis
(Llama-3.1-8B, $B{=}128$ unless a budget is named). \Cref{fig:headline}(a)
plots each mechanism's attention-level metric against its LongBench $\Delta$ vs
\snapkv{}: the four \emph{what}-axis mechanisms gain $8$--$57\%$ on their
objective yet at most $+0.6$ on LongBench, LAQ's $16$ pseudo queries buy
$+2.0$, and deferring the cut by the same $16$ tokens adds a further
$+2.0$---both \dg{} points near $2\times$ the best prior. Every family meets
its objective and none of it transfers (\Cref{tab:audit}(a) \emph{fit},
(b) \emph{select}).

\input{sections/table_audit}

\paragraph{Compensation restores mass, not content.}
\label{sec:audit-comp}
\Cref{tab:audit}(a) is ours throughout: compensation token and
multi-centroid following LOOK-M/KVMerger
\citep{wan2024lookm,wang2024kvmerger}, value absorption following CaM
\citep{zhang2024cam}, per-token bias following FAST
\citep{zweiger2026fast}. Mechanistically (App.~\ref{app:comp}), renormalized mass flows
\emph{multiplicatively} (inflow corr.\ $0.97$ with a token's own attention, so
merging by key similarity cannot route it), post-RoPE key averaging loses
$30\%$ of key norm to phase interference, and a per-token least-squares fit
overfits the eight window queries a prefill affords.

\paragraph{Selection trades coverage for contiguity.}
\label{sec:audit-sel}
\snapkv{}'s max-pooling \emph{loses} $9$--$22\%$ held-out coverage across
$B{=}64$--$512$ yet
\emph{wins} scores by buying span contiguity, and raw ``no pooling'' inverts
both at $B{=}128$ in \Cref{tab:audit}(b). Oracle bounds cap every practical
rule (token oracle $+31.8\%$; block and segment oracles in App.~\ref{app:comp}).
The only signal materially above the window is \emph{real decode queries}:
$+16.5\%$ in \Cref{tab:audit}(b) (\dg$_{16}$'s $k{=}16$ draft queries, matching
LAQ's $m{=}16$), $+30.4\%$ at $B{=}512$, against $+1.2\%$ for lookahead's
pseudo queries in App.~\ref{app:laqcov}. It is also the only
entry in \Cref{tab:audit} whose gain reaches the score: with runs as short as
no pooling's ($1.6$ vs $1.8$) it gains $+3.98$ where no pooling loses $1.73$---%
because it is a cut deferred until the real queries exist (\S\ref{sec:method}),
not a better prefill-time rule: the timing-only control, keeping \snapkv{}'s
long runs, gains the same ($+3.95$, \dg{}-W$_{16}$). \Cref{fig:headline}(b)
cashes it out: \snapkv{} collapses on long-form generation, and SKV+\dg$_2$
recovers FullKV in both regimes.

%% file: sections/table_audit.tex
% Condensed form of the two audit grids. The full versions stay in the appendix:
% App. tab:comp (five compensation mechanisms) and tab:sel (six selection rules).
\begin{table}[t]
\centering\small
\setlength{\tabcolsep}{3pt}
\caption{\textbf{The attention-mass objective is met and the score does not
follow} (Llama-3.1-8B, $B{=}128$; LongBench (LB) $\Delta$ vs \snapkv{};
$\uparrow$: higher is better; full grids in App.~\ref{app:comp}).
Attention-level: each mechanism's gain on its own objective (mass recovered,
error or deficit reduced); Cov.\ $\Delta$: relative gain in held-out coverage
over \snapkv{}; Run len.: mean contiguous run of retained tokens. All rows are
ours except \snapkv{} \citep{li2024snapkv}.}
\label{tab:audit}
\vspace{-5pt}
\begin{subtable}[t]{0.48\textwidth}
\centering\footnotesize
\caption{Compensation (\emph{fit})}
\adjustbox{max width=\linewidth}{%
\begin{tabular}{l r@{\,}l r}
\toprule
Mechanism & \multicolumn{2}{c}{Attention-level $\uparrow$} & LB $\Delta$ $\uparrow$ \\
\midrule
Compensation token & $+8.4\%$ & mass & $-0.03$ \\
Multi-centroid & $+35.7\%$ & mass & $-0.48$ \\
Value absorption & $+11.2\%$ & error & $-0.21$ \\
Per-token bias & $+57.0\%$ & deficit & $+0.02$ \\
\bottomrule
\end{tabular}}
\end{subtable}
\hfill
\begin{subtable}[t]{0.48\textwidth}
\centering\footnotesize
\caption{Selection (\emph{select})}
\adjustbox{max width=\linewidth}{%
\begin{tabular}{lrrr}
\toprule
Rule & Cov.\ $\Delta$ $\uparrow$ & Run len. & LB $\Delta$ $\uparrow$ \\
\midrule
\snapkv{} & -- & 10.4 & -- \\
No pooling & $+11.5\%$ & 1.8 & $-1.73$ \\
Decode queries (\dg$_{16}$) & $+16.5\%$ & 1.6 & $+3.98$ \\
Token oracle & $+31.8\%$ & 1.6 & (bound) \\
\bottomrule
\end{tabular}}
\end{subtable}
\end{table}

%% file: sections/method.tex
\section{Draft-Guided Eviction}
\label{sec:method}

The two problems of \S\ref{sec:audit} share a cause: the cut fires before the
queries that matter exist. \dg{} removes the cause rather than either symptom
by making the real decode queries exist \emph{before} the cut, and both
problems then dissolve: nothing is left to compensate, and the
coverage--contiguity trade-off is an artifact of pre-generation scoring---once
the queries reading the cache are the real ones, the spans an extractive answer
copies have already been read.

\paragraph{Policy.}
\dg{} is a scheduling change, not a new scorer: it defers eviction past the
opening of the generation, as App.~\ref{app:dg} sets out:
\begin{enumerate}[leftmargin=*]
\small
\item \textbf{Draft (phase A).} Prefill normally; generate the first $k$
answer tokens with the \emph{full} cache, capturing the $k$ real decode
queries per layer through forward pre-hooks---no extra pass, the draft being
the beginning of the final answer. Prefill's own forward emits token one, so
$k{=}1$ is the base evictor and $k{=}2$ the minimal dose, used throughout.
\item \textbf{Evict (phase B).} Score every past token, keep the per-head
budget, and edit the cache in place---once, irrevocably. The scorer is a free
slot. By default \dg{} uses the mean attention the $k$ draft queries pay to
each token, which the draft provides at no cost; the timing-only \dg{}-W keeps
the base evictor's own scorer---\snapkv{}'s window, PyramidKV's layer budgets,
H2O's accumulator or StreamingLLM's sinks---unchanged.
\item \textbf{Continue.} Resume on the compressed cache, keeping the draft
tokens as ordinary output.
\end{enumerate}
\S\ref{sec:experiments} finds the two scorers equivalent on average: the
contribution is \emph{when} the cut fires. The change is correspondingly
small to deploy: two stock \texttt{generate()} calls around one in-place cache
edit, with no training, no extra pass, and no hyper-parameter beyond $k$
(App.~\ref{app:dg}).

\paragraph{Cost.}
\dg{} leaves the asymptotic peak unchanged, as \Cref{fig:method}(a) sketches,
and the dose itself is free: total latency is flat in $k$. The cost is one
cache surgery, $1.45\times$ on short answers and $1.02\times$ on long in
\Cref{tab:evidence}(b), against a TTFT of $0.93\times$ \snapkv{}'s that no
score-based prior evictor there beats. If the answer terminates within the
draft, no eviction occurs and the output is bit-identical to FullKV's---the
rational policy, since the cache is freed at answer end.

\paragraph{A hypothesis for why \dg{} works: trajectory anchoring.}
The first $k$ tokens are the generation's high-information steering segment:
for extractive QA they contain the copied span; for summarization they fix
topic and structure. The draft segment is also precisely where prefill-time
eviction bites. Produced from the whole cache, the draft anchors a trajectory
a compressed cache then carries---consistent with the timing-only
\dg{}-W matching the $44.2$ average of \dg{} across \Cref{tab:main} and carrying
nearly all of the recovery in \Cref{fig:method}(b), as \S\ref{sec:experiments}
attributes.

\input{sections/figure_method}

%% file: sections/figure_method.tex
% figure_method: body figure, kept in its own file like the tables; edit here.
% Generated by paper/figures/make_fig_method.py
\begin{figure}[!t]
\centering
\begin{subfigure}[b]{0.605\textwidth}
  \centering
  \includegraphics[width=\textwidth]{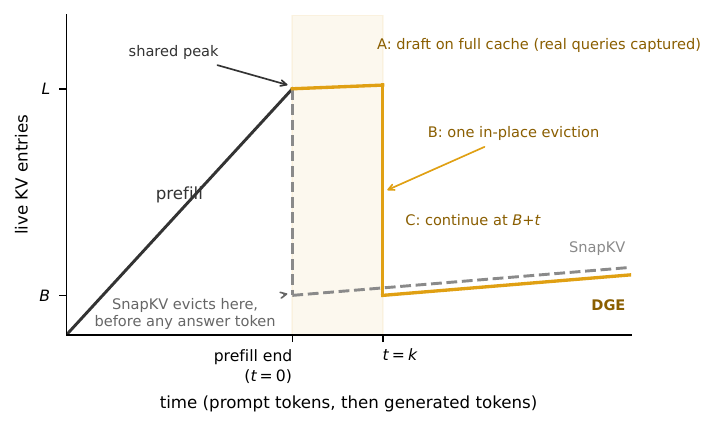}
  \caption{Same steady state, later eviction.}
  \label{fig:method-a}
\end{subfigure}
\hfill
\begin{subfigure}[b]{0.38\textwidth}
  \centering
  \includegraphics[width=\textwidth]{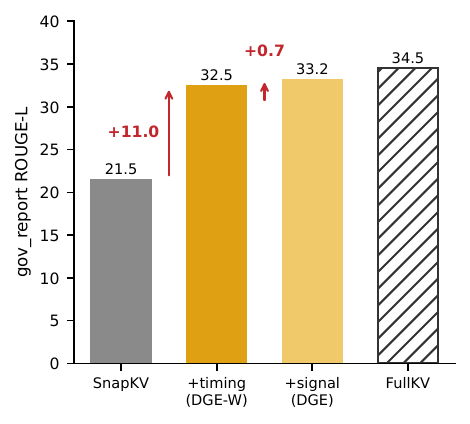}
  \caption{Timing carries the recovery.}
  \label{fig:method-b}
\end{subfigure}
\caption{\textbf{\dg{} overview and gain attribution.} \textbf{(a)}~Live KV
entries over time: one shared prefill ramp to $L$, the same $B{+}t$ steady
state, and the single eviction firing at prefill end for \snapkv{} but after
a $k$-token draft for \dg{}, which holds the full cache for those $k$ steps.
\textbf{(b)}~gov\_report ROUGE-L, Llama-3.1-8B, $B{=}128$, matched 50-sample
subset; FullKV reaches $34.5$.}
\label{fig:method}
\end{figure}

%% file: sections/experiments.tex
\section{Experiments}
\label{sec:experiments}

\S\ref{sec:audit} closed the \emph{what} axis; this section measures what
\S\ref{sec:method}'s deferral buys on the \emph{when} axis: a lead over every
prior method in every backbone $\times$ budget cell, FullKV parity within noise,
and, through a timing-only control, the deferral itself as the cause.

\subsection{Experimental setup}
\label{sec:exp-setup}

\paragraph{Backbone LLMs.}
Llama-3.1-8B-Instruct \citep{dubey2024llama} is the primary backbone, with
replications on Mistral-7B-Instruct-v0.2 \citep{jiang2023mistral},
Qwen2.5-7B/14B-Instruct \citep{qwen2025qwen25} and Llama-3.2-3B
in \Cref{tab:main}, and on Llama-3-8B in App.~\ref{app:llama3}; decoding
and prompt protocol follow \S\ref{sec:setup}'s preliminaries.

\paragraph{Datasets.}
We use LongBench \citep{bai2024longbench}, evaluating all 16 English
datasets: single-document QA (NarrativeQA, Qasper, MultiFieldQA;
\citealp{kocisky2018narrativeqa,dasigi2021qasper}), multi-document QA
(HotpotQA, 2WikiMultiHopQA, MuSiQue;
\citealp{yang2018hotpotqa,ho2020constructing,trivedi2022musique}),
summarization (GovReport, QMSum, MultiNews;
\citealp{huang2021govreport,zhong2021qmsum,fabbri2019multinews}),
few-shot learning (TREC, TriviaQA, SAMSum;
\citealp{li2002trec,joshi2017triviaqa,gliwa2019samsum}), synthetic
retrieval and counting, and code completion (LCC, RepoBench-P;
\citealp{guo2023longcoder,liu2024repobench}). Inputs average
1{,}235--18{,}409 tokens; longer prompts are middle-truncated per the official
protocol, and each set is scored by its official metric (F1, ROUGE-L,
accuracy, edit similarity).

\paragraph{Baselines.}
All baselines keep the same per-layer, per-head KV budget $B$ and differ only
in their retention strategy; our reproductions at $B{=}128$ rank as published
(\snapkv{} $42.2$, PyramidKV $41.6$, StreamingLLM $36.5$):
\begin{itemize}[leftmargin=*]
\small
\item \textbf{FullKV (FKV)} caches every token and is the uncompressed
reference.
\item \textbf{StreamingLLM (SLM)} \citep{xiao2024streamingllm} keeps
attention sinks and a recency window only.
\item \textbf{SnapKV (SKV)} \citep{li2024snapkv} scores past tokens with an
observation window and max-pools before per-head top-$B$; it is the base of
all our variants.
\item \textbf{PyramidKV (PKV)} \citep{cai2024pyramidkv} adds a layer-decaying
budget schedule.
\item \textbf{Lookahead Q-Cache (LAQ)} \citep{ge2025lookahead}, the closest
prior extra-pass baseline, at its best dose ($m{=}16$) in \Cref{tab:main},
across doses in \Cref{tab:evidence}(a) and over the full grid in
App.~\ref{app:tables}.
\end{itemize}

\paragraph{Our variants.}
\label{sec:exp-variants}
Each variant is prefixed with the base evictor it modifies and keeps its
budget; the suffixes isolate the two axes of \S\ref{sec:method} (pseudocode
in App.~\ref{app:dg}):
\begin{itemize}[leftmargin=*]
\small
\item \textbf{SKV+\dg$_{k}$} (\emph{eviction timing $+$ signal}) is the default
of \S\ref{sec:method}: the cut moves to decode step $k$ and is scored by the
$k$ real draft queries rather than the observation window. Budget and
allocation stay \snapkv{}'s, so any difference from the SKV row comes from
those two changes alone.
\item \textbf{SKV+\dg{}-W$_{k}$} (\emph{timing only}) defers eviction
identically but scores with \snapkv{}'s unmodified window recipe, so \dg{}-W
differs from the SKV row solely in the step at which eviction happens.
\end{itemize}
The deferral composes with the other retention rules the same way
(PKV+/H2O+/SLM+\dg{} rows, App.~\ref{app:tables}).

Unless noted otherwise: budget $B{=}128$ for single-budget results
($B\in\{32,64,128\}$ in \Cref{tab:main}),
observation window $w{=}8$, pooling kernel 7, draft length $k{=}2$ ($k$ up to
$32$ in the dose grid),
matched sample sets across methods, one run per configuration (greedy; CIs in
App.~\ref{app:ci}).

\subsection{Main results}
\label{sec:exp-main}

\input{sections/table_summary}

\paragraph{\dg{} leads every prior method at every budget, near FullKV.}
Averaged over all fifteen backbone $\times$ budget cells, as in the Avg.\ column
of \Cref{tab:main}, SKV+\dg$_{2}$ scores $44.17$ against $40.13$ for the
strongest prior method, SKV+LAQ at its best dose, and $35.34$ for \snapkv{}; both \dg{} variants beat every prior
method in all fifteen cells, by at least $1.01$, and stay within $0.77$ of
FullKV---matching or exceeding it in five cells---where the strongest prior
method trails FullKV by $1.5$--$9.0$ (per-dataset scores: App.~\ref{app:tables},
except Mistral-7B and Llama-3.2-3B at $B{=}32$). The deferral is essentially
\emph{budget-invariant}: averaged over backbones, as $B$ shrinks from $128$ to
$32$, SKV+\dg$_{2}$ moves only from $44.30$ to $44.04$, while \snapkv{} falls
from $39.06$ to $31.06$, so \dg{}'s lead over the best prior method grows from
$1.1$--$2.2$ at $B{=}128$ to $5.7$--$8.6$ at $B{=}32$. On the
sixth backbone, Llama-3-8B, a $k{=}32$ draft closes $72\%$ of the \snapkv{} gap
to FullKV at $B{=}128$, as App.~\ref{app:llama3} shows. The deferral also
composes with other base evictors: PKV/H2O/SLM+\dg$_{2}$ land within $0.9$ of
FullKV in the three settings App.~\ref{app:analysis} reports, SLM gaining
$+12.3$ on average (PKV's non-monotone Qwen columns: a discrete-schedule
artifact, explained there too).

\paragraph{Where the gain lives.}
Draft termination (answers ending inside the draft) is
rare at the headline dose---$11.2\%$ of samples at $k{=}2$, $0\%$ on the
long-generation categories, per App.~\ref{app:tables}---so the gain must come from
generation on the compressed cache. On gov\_report, where $\sim$99\% of tokens
decode after eviction, SKV+\dg$_{2}$ lifts ROUGE-L from $22.3$ to $32.6$
against FullKV's $33.8$ (\Cref{fig:dose}(b)); on qmsum both deferred variants
beat FullKV (App.~\ref{app:tables}). The opening tokens, not the retained content,
carry the generation. Off LongBench, Needle-in-a-Haystack (Qwen2.5-7B,
$B{=}32$)---where a query-aware selector should simply keep the needle---makes
the same point: \snapkv{} recovers $74.7$ and H2O $42.1$, while \dg$_{2}$
returns $100.0$ against FullKV's $98.9$ (App.~\ref{app:needle}).

\subsection{Attribution ablation}

The attribution rests on the timing-only control \dg{}-W of
\S\ref{sec:exp-variants}, which splits \dg{}'s recovery into a pure timing
term, \snapkv{} to \dg{}-W, and a pure signal term, \dg{}-W to \dg{}. On the matched 50-sample gov\_report split of
\Cref{fig:method}(b):
$\Delta(\text{timing}) = \dg\text{-W} - \snapkv = +11.0$ and
$\Delta(\text{signal}) = \dg - \dg\text{-W} = +0.7$---timing accounts for
$94\%$ of the recovery. The split holds across \Cref{tab:main}:
averaged over all fifteen cells, deferring eviction alone lifts \snapkv{} from
$35.34$ to \dg{}-W's $44.17$, and swapping in the real decode queries adds
nothing further, \dg{} also averaging $44.17$. On Llama-3.1-8B at $B{=}128$,
App.~\ref{app:ci}'s paired bootstrap separates the timing gain of $+4.11$ from
zero but not the $0.04$ signal term. The eviction
signal---the axis every selection
method optimizes---is thus worth at most $0.44$ LongBench points in any cell of
\Cref{tab:main} once the cut is deferred: the deferral is the effect. Hence
\S\ref{sec:method}'s free scorer slot: the draft queries cost nothing, and the
base evictor's own scores do as well on average---in the few cells where
App.~\ref{app:ci}'s bootstrap separates the two, \dg{}-W is ahead
(Qwen2.5-14B at $B{\le}64$).

\subsection{Dose-response in $k$}

\input{sections/figure_dose}

One decode step is the whole dose, as \Cref{fig:dose}(a) shows on every
backbone but Llama-3-8B (App.~\ref{app:llama3}; why the dose is
backbone-dependent: App.~\ref{app:analysis}). At $B{=}128$ on
Llama-3.1-8B, the one backbone the full grid of \Cref{tab:ablation}
(App.~\ref{app:tables}) covers, $k{=}2/4/8/16/32$ score
$46.37/46.29/46.12/46.19/46.13$. The minimal
dose $k{=}2$ (\S\ref{sec:method}) already jumps $+4.2$ from \snapkv{} and
stays flat within noise thereafter, with or without the draft-query scores
(\dg{}-W$_{16}$: $46.17$). The pattern
is budget-invariant: at $B{=}32$ the dose is flat from $k{=}2$
(\Cref{tab:ablation}) despite \snapkv{} collapsing to $35.85$. This flatness, and \dg{}-W tracking \dg{}
although its scorer never sees the draft, are consistent with
App.~\ref{app:whymath}'s eviction-mass argument.

\input{sections/table_evidence}

\subsection{Lookahead-query eviction}
\label{sec:exp-lookahead}
LAQ \citep{ge2025lookahead} scores with $m$ \emph{pseudo}
future queries but still evicts at prefill end---the \emph{what} axis at its
limit. Run in its own code base under our matched
protocol, its dose saturates $\sim$2 points below FullKV. \dg{}-W$_{2}$ defers one
decode step and clears LAQ's best dose by $+2.1$ at $B{=}128$ in
\Cref{tab:evidence}(a), and by $+5.8$ at $B{=}32$ (App.~\ref{app:analysis}):
tighter budgets widen the gap, and
\Cref{fig:dose}(b) places the margin on the hard summarization and code sets. Across all fifteen
backbone $\times$ budget cells, the Avg.\ column of \Cref{tab:main} puts the
margin at $+4.0$.

\subsection{Complexity and overheads}
\Cref{tab:evidence}(b) prices the deferral ($d$: head dimension): TTFT improves
to $0.93\times$ \snapkv{}'s, since eviction leaves the prefill path, and the
one-off cache surgery costs $1.45\times$ on $32$-token answers (H2O's
accumulator: $2.15\times$), amortising to $1.02\times$ by $512$.

%% file: sections/table_summary.tex
% Compact body table: LongBench averages only. Full 16-dataset breakdown and
% the dose grid live in the appendix (tab:main-full, tab:ablation).
% Backbones ordered by parameter count: 3B < Mistral-7B < Qwen2.5-7B < Llama-3.1-8B < Qwen2.5-14B.
\begin{table}[t]
\centering
\caption{\textbf{One decode step of deferral leads every prior method at every
budget} (LongBench average over 16 sets; five backbones ordered by size; KV
budgets $\{32,64,128\}$; FullKV is budget-independent; Avg.\ is the mean over
all fifteen backbone $\times$ budget cells; \textbf{bold} is the best
compression method per column, FullKV excluded as the uncompressed
reference). Both deferred rows are ours: the shaded
\textbf{bold} row is the default \dg{}, \dg{}-W$_{2}$ the timing-only control (same
deferral, \snapkv{}'s scores). LAQ \citep{ge2025lookahead} is at its best
dose, $m{=}16$.}
\label{tab:main}
\vspace{2pt}
\setlength{\tabcolsep}{3pt}
\small
\adjustbox{max width=\textwidth}{%
\begin{tabular}{l ccc ccc ccc ccc ccc c}
\toprule
& \multicolumn{3}{c}{Llama-3.2-3B} & \multicolumn{3}{c}{Mistral-7B-v0.2}
& \multicolumn{3}{c}{Qwen2.5-7B} & \multicolumn{3}{c}{Llama-3.1-8B}
& \multicolumn{3}{c}{Qwen2.5-14B} & \multirow{2}{*}{Avg.} \\
\cmidrule(lr){2-4}\cmidrule(lr){5-7}\cmidrule(lr){8-10}\cmidrule(lr){11-13}\cmidrule(lr){14-16}
Method & $32$ & $64$ & $128$ & $32$ & $64$ & $128$ & $32$ & $64$ & $128$
       & $32$ & $64$ & $128$ & $32$ & $64$ & $128$ & \\
\midrule
FKV & \multicolumn{3}{c}{$42.45$} & \multicolumn{3}{c}{$41.35$}
    & \multicolumn{3}{c}{$45.54$} & \multicolumn{3}{c}{$46.24$}
    & \multicolumn{3}{c}{$46.05$} & $44.33$ \\
\midrule
SLM & 28.14 & 29.30 & 30.86 & 24.58 & 25.68 & 27.10 & 27.11 & 28.90 & 30.19 & 33.44 & 35.14 & 36.51 & 28.63 & 30.40 & 31.82 & 29.85  \\
PKV & 33.07 & 33.50 & 37.68 & 27.87 & 31.83 & 34.91 & 32.69 & 30.74 & 38.38 & 36.54 & 39.58 & 41.64 & 35.07 & 34.64 & 40.41 & 35.24  \\
SKV & 31.57 & 35.64 & 38.07 & 26.37 & 31.35 & 34.53 & 31.56 & 36.77 & 39.98 & 35.85 & 39.38 & 42.22 & 29.94 & 36.42 & 40.51 & 35.34  \\
SKV+LAQ & 35.79 & 39.13 & 40.90 & 33.94 & 37.21 & 39.86 & 37.81 & 41.59 & 43.47 & 40.65 & 43.17 & 44.20 & 37.10 & 42.63 & 44.46 & 40.13  \\
\midrule
SKV+\dg{}-W$_{2}$ & 42.10 & 42.30 & 42.32 & 40.58 & \textbf{41.10} & 40.87 & \textbf{45.17} & 45.20 & \textbf{45.36} & \textbf{46.40} & 46.20 & 46.33 & \textbf{46.18} & \textbf{46.31} & 46.18 & \textbf{44.17}  \\
\hi \textbf{SKV+\dg$_{2}$ (Ours)} & \textbf{42.32} & \textbf{42.36} & \textbf{42.56} & \textbf{40.72} & 40.90 & \textbf{41.00} & 45.05 & \textbf{45.32} & 45.32 & 46.39 & \textbf{46.36} & \textbf{46.37} & 45.74 & 45.94 & \textbf{46.27} & \textbf{44.17}  \\
\bottomrule
\end{tabular}}
\end{table}

%% file: sections/figure_dose.tex
% figure_dose: body figure, kept in its own file like the tables; edit here.
\begin{figure}[!t]
\centering
\begin{subfigure}[b]{0.49\textwidth}
  \centering
  \includegraphics[width=\textwidth]{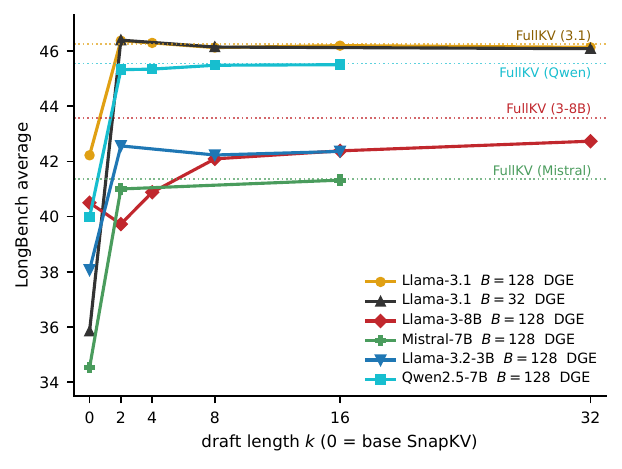}
  \caption{Minimal dose is backbone-dependent.}
  \label{fig:dose-a}
\end{subfigure}
\hfill
\begin{subfigure}[b]{0.49\textwidth}
  \centering
  \includegraphics[width=\textwidth]{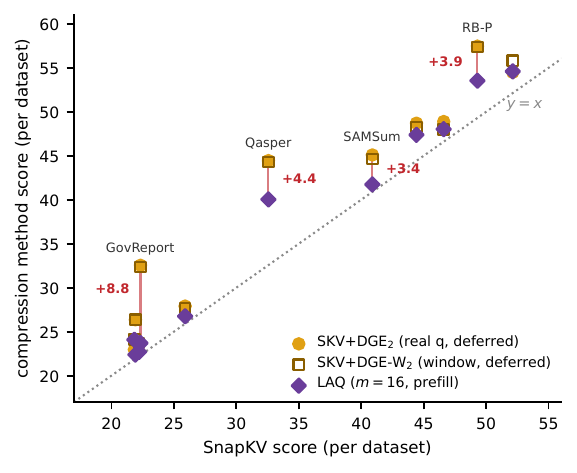}
  \caption{Timing rescues hard sets; LAQ does not.}
  \label{fig:dose-b}
\end{subfigure}
\caption{\textbf{When, not what: dose and per-dataset view}
(Llama-3.1-8B unless noted; LongBench).
\textbf{(a)}~Four of the five swept backbones saturate at $k{=}2$ ($k{=}0$:
base \snapkv{}; dotted: FullKV; Llama-3.1-8B also at $B{=}32$); the older
Llama-3-8B climbs through $k{=}32$, as App.~\ref{app:llama3} analyzes.
\textbf{(b)}~Per-dataset scores at $B{=}128$ ($y{>}x$ beats \snapkv{}), zoomed
to the sets where methods diverge; red connectors mark the lead over Lookahead
Q-Cache \citep{ge2025lookahead}, $+3.4$ to $+8.8$ on the labelled sets.}
\label{fig:dose}
\end{figure}

%% file: sections/table_evidence.tex
% Condensed form of the two grids §6.5 and §6.6 rely on. The full versions stay in
% the appendix: App. tab:lookahead (LAQ dose m=4..32 at B=64/128) and
% tab:complexity (all six baselines, peak/decode KV, footnote on H2O).
\begin{table}[t]
\centering\small
\caption{\textbf{One deferral step clears the \emph{what}-axis ceiling at no
prefill cost} (Llama-3.1-8B; matched protocol). \dg{} rows are ours;
App.~\ref{app:tables} and~\ref{app:complexity} give the full grids. Latency is wall
clock vs \snapkv{} at a pinned $32$ new tokens for short, $512$ for long;
TTFT uses one fixed $7.8$k-token prompt.}
\label{tab:evidence}
\vspace{2pt}
\begin{subtable}[t]{0.545\textwidth}
\centering\footnotesize
\caption{Pushing \emph{what} to its limit (LongBench avg.)}
\setlength{\tabcolsep}{4pt}
\adjustbox{max width=\textwidth}{%
\begin{tabular}{llcc}
\toprule
Method & Evicts / signal & $B{=}64$ & $B{=}128$ \\
\midrule
FKV & never & 46.24 & 46.24 \\
SKV & prefill / window & 39.38 & 42.22 \\
\midrule
LAQ$_{m=4}$ & prefill / pseudo & 42.75 & 43.81 \\
LAQ$_{m=8}$ & prefill / pseudo & 43.11 & 44.13 \\
LAQ$_{m=16}$ & prefill / pseudo & 43.17 & 44.20 \\
\midrule
SKV+\dg{}-W$_{2}$ & step 2 / window & 46.20 & 46.33 \\
SKV+\dg$_{2}$ & step 2 / draft & \textbf{46.36} & \textbf{46.37} \\
\bottomrule
\end{tabular}}
\end{subtable}\hfill
\begin{subtable}[t]{0.435\textwidth}
\centering\footnotesize
\caption{What deferral costs}
\setlength{\tabcolsep}{4pt}
\adjustbox{max width=\textwidth}{%
\begin{tabular}{lccc}
\toprule
Method & Extra scoring & TTFT & \makecell{Latency\\[-1pt]\scriptsize short/long} \\
\midrule
FKV & -- & 0.91$\times$ & 0.89/0.83 \\
SKV & $O(Lwd)$ & 1.00$\times$ & 1.00/1.00 \\
LAQ$_{m{=}16}$ & $O(mLd)$ & 1.42$\times$ & 0.95/0.83 \\
H2O & $O(L^{2}d)$ & 3.45$\times$ & 2.15/1.06 \\
\midrule
SKV+\dg{}-W$_{k}$ & $O(Lwd)$ & 0.92$\times$ & 1.45/1.02 \\
SKV+\dg$_{k}$ & $O(kLd)$ & 0.93$\times$ & 1.45/1.02 \\
\bottomrule
\end{tabular}}
\end{subtable}
\end{table}

%% file: sections/discussion.tex
\section{Discussion}
\label{sec:discussion}

\paragraph{Sharper benchmarking.}
The standard protocol---evict at end-of-prefill, then generate---folds two
decisions into one number; one deferred-eviction row separates them, and
averaged over backbones and budgets a timing-only deferral closes almost all of
the \snapkv{}-to-FullKV gap of \Cref{tab:main}: the reported penalty is timing,
not selection quality.

\paragraph{What the objective measures.}
Attention-output fidelity is a geometric quantity \S\ref{sec:audit}'s mechanisms
already deliver near their ceilings; task quality asks instead for retrievable
content and a stable trajectory.

\paragraph{Limitations and research directions.}
Trajectory anchoring is inferred from ablations, not isolated, and decoding
is greedy throughout, so how far it survives sampling is open. LongBench's short answers cap the doses it can
probe---at $k{=}16$ a third of Llama-3.1-8B's answers terminate inside the draft---pointing at
long-form, agentic and multilingual workloads. The draft holds the full cache,
so the prototype's working set is $2\times$ FullKV's ($1.13\times$ in the native
GQA layout, App.~\ref{app:complexity}); batching it in a paged-attention server
would turn the TTFT gain into throughput.

%% file: sections/conclusion.tex
\section{Conclusion}
\label{sec:conclusion}

Training-free KV-cache compression evicts at the end of prefill to preserve
the attention mass future queries will use---optimizing \emph{what} to
keep---and that objective fails in two ways: restored mass does not restore
task quality, and extra coverage can hurt when it arrives fragmented. Both
occur because eviction fires before the queries that will read the cache
exist. \dg{} removes that cause by deferring eviction until after
drafting the first $k{=}2$ answer tokens on the full cache; it drops into
\snapkv{}, PyramidKV, H2O and StreamingLLM alike, changing \emph{when}
eviction occurs, not \emph{what} is selected. It leads every prior method at
every budget on five of six backbones ($44.2$ on LongBench vs FullKV's
$44.3$), and the timing-only control \dg{}-W reaches the same score: the gain
is \emph{trajectory anchoring}, from the deferral itself. Before optimizing
\emph{what} to keep, fix \emph{when} to evict.

%% file: sections/statements.tex
% Statements required or recommended by the ICLR 2027 template
% (iclr2027_conference.tex). They sit after the main text and before the
% references, and do not count toward the page limit.

\subsection*{AI use statement}
We used generative AI tools solely for language editing, including grammar correction and concise rewriting. All research, ideas, and manuscript content were developed by the authors. We reviewed all AI-assisted edits and take full responsibility for the final content.

%% file: sections/appendix.tex
% Tight list margins for the appendix only (enumitem is loaded in main.tex).
\setlist[itemize]{leftmargin=1.1em,itemsep=1pt,topsep=2pt,parsep=0pt}

\noindent\textbf{Appendix map.} \Cref{app:expdetails,app:diagnostics} fix the
protocol behind every number: the LongBench run itself, then the offline
diagnostic harness. \Cref{app:comp} derives the compensation and selection
mechanisms \S\ref{sec:audit} analyzes and holds their two full grids, and
\Cref{app:matchloss} the attention-matching loss they optimize.
\Cref{app:dg} specifies \dg{} as pseudocode and derives a per-step
eviction-mass bound that motivates why deferral can recover FullKV; \Cref{app:draftlike}
contrasts it with the other passes that also run before the cut, and
\Cref{app:complexity} prices it in latency, TTFT and memory.
\Cref{app:llama3,app:needle} are the two settings the body reports only in
summary---the sixth backbone and retrieval---and \Cref{app:ci} the paired
bootstrap behind every interval. \Cref{app:tables} holds the per-dataset
score tables, \Cref{app:analysis} the remaining ablations, and
\Cref{app:negative} the diagnostics that ruled a hypothesis out.

\section{Experimental details}
\label{app:expdetails}

\paragraph{Backbones and context.}
We evaluate six instruct backbones. Llama-3.1-8B-Instruct \citep{dubey2024llama}
is the primary model (128K native context via RoPE scaling); backbone checks use
Mistral-7B-Instruct-v0.2 \citep{jiang2023mistral} (32K context, a different
architecture family), Qwen2.5-7B-Instruct \citep{qwen2025qwen25} (a third
architecture family with 28 layers and 4 KV heads, ChatML template),
Qwen2.5-14B-Instruct (the largest backbone, 48 layers and 8 KV heads, same
ChatML protocol), Llama-3.2-3B-Instruct (a smaller model of the same instruct
lineage, for a scale check), and Llama-3-8B-Instruct (8K native context, no RoPE
scaling; the shorter-context predecessor of the primary backbone). All share the
per-head KV-Cache geometry \dg{} operates on and differ only in long-context
adaptation and instruction-tuning generation, contrasted in \Cref{tab:backbones}.

\paragraph{Datasets and metrics.}
LongBench \citep{bai2024longbench}, all 16 English datasets: single-document QA
(NarrativeQA, Qasper, MultiFieldQA), multi-document QA (HotpotQA, 2WikiMQA,
MuSiQue), summarization (GovReport, QMSum, MultiNews), few-shot learning (TREC,
TriviaQA, SAMSum), synthetic retrieval/counting (PassageCount,
PassageRetrieval), and code completion (Lcc, RepoBench-P). We report the official
per-dataset metric---F1 for QA, ROUGE-L for summarization, classification
accuracy for TREC and the synthetic tasks, and edit similarity for code---and
the official per-dataset max-new-token budget. Prompts exceeding the truncation
limit are middle-truncated (first half $+$ last half) following the official
protocol; the limit is $7{,}500$ tokens for the Llama and Qwen backbones
(matched across methods) and $31{,}500$ for Mistral.

\paragraph{Decoding and prompt protocol.}
Greedy decoding throughout, so results are deterministic and there is no
seed variance to average over. We use the repository's corrected Llama-3 chat
template and stop-token set (adding \texttt{<|eot\_id|>} and a per-dataset
newline stop for the few-shot sets, which otherwise over-generate); Mistral uses
its \texttt{[INST]} wrapping. Every method sees the identical decoding pipeline
and matched sample sets.

\paragraph{Compression settings.}
Per-head budget $B\in\{32,64,128,512\}$ ($B{=}128$ unless noted), observation
window $w{=}8$, max-pool kernel $7$ (the \snapkv{} recipe). The compressed cache
is stored per query head (GQA layout). \dg{} draft length $k\in\{2,\dots,32\}$
($k{=}2$ for both \dg{} and the timing-only \dg{}-W in the main tables,
$k{=}16$ for the two deferred markers of \Cref{fig:headline}(a)); at $k{=}1$
the policy coincides with the base evictor by construction (the first token comes
from the prefill forward), so $k{=}2$---one full-cache decode step---is the
smallest dose the method admits; whether it is also \emph{sufficient} is
backbone-dependent, as \S\ref{app:llama3} shows.

\paragraph{Software and hardware.}
PyTorch with \texttt{transformers} 4.44.2; flash-attention 2.6.3 for the
window/positional scorers and \dg{}, and the SDPA backend for the compensation
variants ($\beta$-bias and value absorption apply a decode-time attention-mask
bias). Each run uses a single 80\,GB GPU; all reported cells are our own
runs (H2O is omitted on the Mistral-7B backbone).

\paragraph{Lookahead Q-Cache comparison.}
We run LAQ \citep{ge2025lookahead} in its own released code base on the same
backbone, datasets, truncation, and greedy decoding, porting in only our Llama-3
chat template and stop tokens. Without this alignment LAQ's verbose, unterminated
generations score $\sim$5 points lower (e.g.\ Qasper $10.96\!\to\!39.37$), so the
alignment is required for the matched comparison of \Cref{tab:lookahead}
and \S\ref{sec:exp-lookahead}.

\section{Diagnostic details}
\label{app:diagnostics}
All diagnostics share one harness (released with the code).
The model runs as plain FullKV; forward \emph{pre}-hooks on every attention
layer recompute post-RoPE $Q,K$ from the layer input, so the probe observes
exactly what the eviction code would see without altering the forward pass.
Selection replicates \snapkv{}'s scoring path operation for operation---%
float32 softmax over the last $w$ queries, score sum over past keys, maxpool
(kernel $7$), per-head top-$B$---so a rule's diagnostic score and its
deployed behavior cannot diverge through dtype or pooling drift.

The key design choice is the \emph{held-out split}: every decode-step query is
replayed against the full uncompressed $K$, giving the attention mass a FullKV
run would place on the tokens a rule evicts ($\beta_{\mathrm{dec}}$,
\S\ref{sec:audit}). In-window quantities ($\beta_{\mathrm{win}}$) are what the
rule can see; decode-side quantities are what it is graded on. Every selection
variant in the ladder of \S\ref{sec:audit-sel} (pooling ablations, window
sizes, layer-decay, key-norm re-weighting, block granularities $8/16/32$,
cross-layer sharing, repeat-prefill) is scored by this held-out decode
coverage under identical budgets; \Cref{tab:sel} reports the resulting
granularity ladder and \Cref{app:negative} logs the rules the diagnostic
ruled out.

\section{Compensation and selection details}
\label{app:comp}

\begin{figure}[!htbp]
\centering
\begin{subfigure}[b]{0.49\textwidth}
  \centering\includegraphics[width=\textwidth]{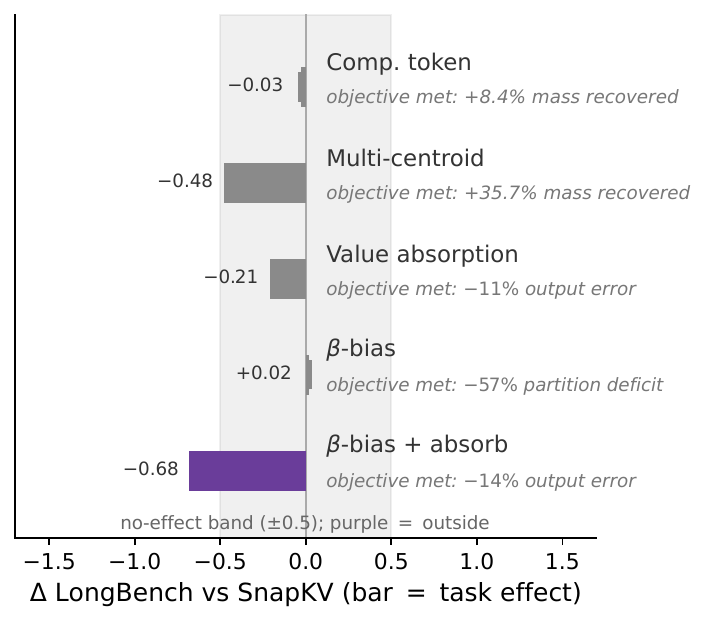}
  \caption{Compensation: objective met, score flat.}
  \label{fig:audit-a}
\end{subfigure}
\hfill
\begin{subfigure}[b]{0.49\textwidth}
  \centering\includegraphics[width=\textwidth]{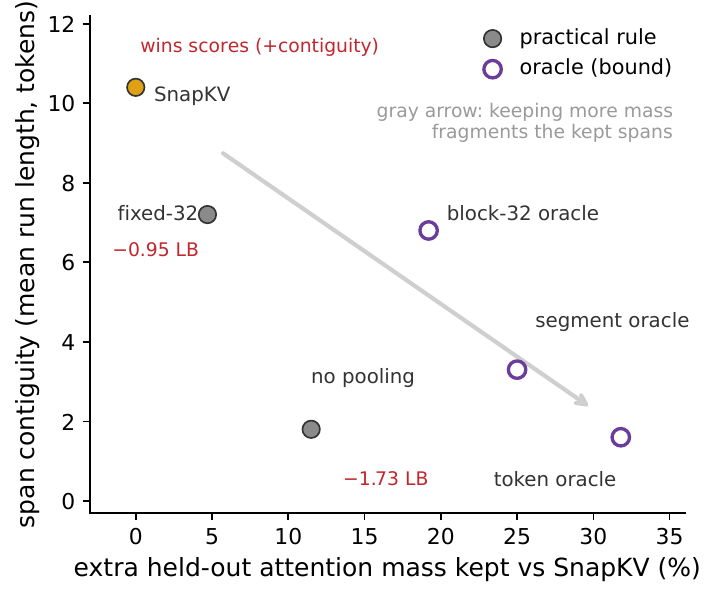}
  \caption{Selection: coverage vs contiguity.}
  \label{fig:audit-b}
\end{subfigure}
\caption{\textbf{The attention-mass analysis, visualized} (data from
\Cref{tab:comp,tab:sel}): \textbf{(a)}~one row per compensation mechanism,
\textbf{(b)}~one point per selection rule.}
\label{fig:audit}
\end{figure}

\Cref{fig:audit} summarizes the two key points of this section:
compensation mechanisms meet their attention-level objective yet move
LongBench by $\sim$0 in \Cref{tab:comp}, and selection rules trade
held-out coverage against span contiguity---\snapkv{} keeps the least raw
mass but the longest contiguous runs and wins scores, while raw-mass rules
and oracles buy coverage by fragmenting spans in \Cref{tab:sel}.

\paragraph{Mechanisms (all train-free).}
(i)~a single attention-matched compensation token per head (value centroid,
phase-aligned key centroid, window-estimated mass target matched by key-scale
search; FA2-compatible); (ii)~multi-centroid variant ($c$ spherical-$k$means
cluster tokens); (iii)~value absorption along the empirical window
co-attention transport; (iv)~per-token bias in the style of FAST
\citep{zweiger2026fast}, $\beta_i{=}\log(1+m_i/s_i)$ applied through the
attention mask.

\begin{table}[htbp]
\centering\small
\caption{\textbf{Compensation dose-response.} Weak positives appear only on
diffuse tasks (narrativeqa $+1.1$, gov\_report $+0.7$, qmsum $+1.3$). All five rows are our
own train-free implementations, not runs of the published methods: the
compensation-token and multi-centroid rows fold evicted entries into retained
ones in the style of \citet{wan2024lookm} and \citet{wang2024kvmerger}, and
value absorption in the style of \citet{zhang2024cam}; the per-token bias is FAST-style
(\citealp{zweiger2026fast}) but a closed form $\beta_i{=}\log(1{+}m_i/s_i)$
that replaces FAST's NNLS fit; the last row combines that bias with
absorption. All five are derived below. Effects here are signed
measurements, so a reduction reads negative; \Cref{tab:audit}(a) reports the
same quantities as improvements, where higher is better throughout.}
\label{tab:comp}
\vspace{2pt}
\begin{tabular}{lcc}
\toprule
Variant & Attention-level effect & LongBench $\Delta$ \\
\midrule
Comp.\ token (v1) & recovers 8.4\% of lost mass & $-0.03$ (16 ds) \\
Multi-centroid $c{=}8$ & recovers 35.7\% of lost mass & $-0.48$ (16 ds) \\
Value absorption & output error $-11\ldots{-14}\%$ & $-0.21$ (16 ds) \\
Per-token bias & partition deficit $-57\%$ & $+0.02$ (16 ds) \\
Bias + absorption & output error $-14\ldots{-17}\%$ & $-0.68$ (16 ds) \\
\bottomrule
\end{tabular}
\end{table}

\paragraph{Baseline.} Every LongBench $\Delta$ tabulated in
\Cref{tab:comp,tab:sel}, and each of the four \emph{what}-axis points of
\Cref{fig:headline}(a), save the no-pooling entry, is measured against one
\snapkv{} run on all sixteen
sets, whose average is $42.15$. That is $0.06$ below the $42.22$
\Cref{tab:evidence}(a) and \Cref{tab:main} print ($42.1535$ against
$42.2161$), and the gap is one of sample size rather than of runs: this run
stops at $50$ of the $200$ gov\_report and qmsum prompts, its predictions
there are a prefix of the master sweep's, and the two agree exactly on the
other fourteen sets. Each $\Delta$ here is that variant's own sixteen-set
average minus this $42.1535$---the no-pooling entry, in \Cref{tab:sel} and in \Cref{fig:headline}(a) alike, excepted,
its retained $-1.73$ coming from a 3-dataset $\times$ 50 judgment run rather
than a sixteen-set average---so the variant's own number is untouched and
every $\Delta$ is exactly $0.0626$ more generous than the same variant scored
against the $42.2161$ run, whatever that variant's own sample counts are (value
absorption and the per-token bias are short on sets other than gov\_report and qmsum). The other five points of \Cref{fig:headline}(a) are measured against
that $42.22$, but only \snapkv{} and the two \dg{} markers come from the
master sweep: PKV is its own $200$-prompt run and LAQ the separate code-base
run \Cref{tab:lookahead} reports.

\paragraph{Mechanistic findings.}
(a)~Renormalized mass flows \emph{multiplicatively}: inflow at a retained
token correlates $0.97$ with its own attention, not key similarity---so
similarity-merging (LOOK-M/KVMerger; \citealp{wan2024lookm,wang2024kvmerger})
cannot route mass correctly (kernel-transport flow corr.\ $\approx 0$).
(b)~Post-RoPE arithmetic key averaging loses $30\%$ of key norm to phase
interference (mean per-pair resultant $R{\approx}0.33$, the whole-key norm
ratio being $0.70$); a circular-mean construction preserves it.
(c)~At $B{=}128$, an unstructured least-squares value correction---the
counterpart of FAST's ordinary-least-squares $C_v$ fit, not of its NNLS mass
fit---ridge-fit on
half of the real decode queries, does not transfer to the held-out half: its
residual is $1.08\times$ the uncorrected error, so an oracle fit with
$B{\times}d$ free parameters ends up worse than no correction at all.
Structured closed forms have no fit to overfit.
(d)~Repeat-prefill queries follow induction/copy patterns (run length $1.2$)
and score selection \emph{worse} than the window ($-13\%$ held-out coverage).

\begin{table}[htbp]
\centering\small
\caption{\textbf{Selection decomposition.} The \snapkv{} row is the
published method \citep{li2024snapkv}; every rule below it is ours---two
variants of its selection (block granularity, pooling removed) and three
oracle bounds---so no external method is run here. The real-decode-query row
of \Cref{tab:audit}(b) is absent because it is not a prefill-time rule: it
needs queries that do not exist until decoding, which is what
\S\ref{sec:method} defers the cut to obtain. Its LongBench entry, $+3.98$
against the $42.22$ sweep run, is therefore the SKV+\dg$_{16}$ row of the
dose grid of \Cref{tab:ablation} ($46.19$ printed; the difference is taken
on unrounded averages)---the same $k{=}16$ \dg{} marker as
\Cref{fig:headline}(a)---not a rule scored on this ladder; the coverage and run
length beside it are measured offline like every other row here.}
\label{tab:sel}
\vspace{2pt}
\begin{tabular}{lccc}
\toprule
Rule ($B{=}128$) & held-out coverage $\Delta$ $\uparrow$ & run len. & LongBench $\Delta$ $\uparrow$ \\
\midrule
\snapkv{} (window+maxpool) & -- & 10.4 & -- \\
fixed 32-blocks & $+4.7\%$ & 7.2 & $-0.95$ \\
no pooling & $+11.5\%$ & 1.8 & $-1.73$ \\
\midrule
token oracle & $+31.8\%$ & 1.6 & (bound) \\
block-32 oracle & $+19.2\%$ & 6.8 & (bound) \\
segment oracle & $+25.0\%$ & 3.3 & (bound) \\
\bottomrule
\end{tabular}
\end{table}

\paragraph{Compensation token (mass matching).} For evicted set $E$ with
window-score weights $w_j$ ($\sum_{j\in E} w_j{=}1$), the token's value is the
centroid $v^\ast=\sum_j w_j v_j$. Its key direction is the \emph{phase-aligned}
centroid: each RoPE rotate-half pair $(d, d{+}64)$ is treated as a complex
number $z_j$, and the weighted complex sum is renormalized to the weighted
magnitude, $\tilde z = \frac{\sum_j w_j z_j}{|\sum_j w_j z_j|}\sum_j w_j|z_j|$,
so the circular-mean phase is kept and no norm is lost to phase interference.
(A naive arithmetic mean shrinks by the resultant length
$R=|\sum_j w_j z_j| / \sum_j w_j |z_j| \in[0,1]$; $R$ is also our phase-dispersion
measure.) The token's target mass is $\beta^\ast=\beta_{\mathrm{win}}\cdot R$
(dispersed phases shrink the target), clamped to $[10^{-4},0.5]$. With
retained log-partition $\log\hat Z_i$ per window query $i$ and $c_i=q_i^\top u/\sqrt d$ for
unit key direction $u$, the mass the token receives at key scale $s$ is
$\overline{\sigma(s\,c_i - \log\hat Z_i)}$; $s$ is chosen on a $129$-point grid over
$[0,\,25/\max_i|c_i|]$ (exp-safe) to match $\beta^\ast$. Pure key
\emph{scaling}---no logit bias---so the token runs unchanged under
FlashAttention-2. The multi-centroid variant first splits $E$ by weighted
spherical $k$-means (deterministic position-stratified init) and builds one
token per cluster.

\paragraph{Value absorption and the closed-form bias.} Both use the same
empirical transport: with window-attention matrix $A$, co-attention
$C_{ij}=\sum_{w} A_{wi}A_{wj}$ (retained $i$, evicted $j$), column-normalized
to $T_{j\to i}$. Absorption folds evicted values into retained ones,
$c_i = v_i + \bigl(\sum_j \mu_j T_{j\to i} v_j - m_i v_i\bigr)/(s_i{+}m_i)$
with transported mass $m_i=\sum_j \mu_j T_{j\to i}$---closed form, keys and
cache shape untouched, decode unchanged. The per-token bias is the closed form
$\beta_i=\log(1+m_i/s_i)$, clamped at $10$, added to the decode attention mask,
which grows token $i$'s softmax mass from $s_i$ toward $s_i{+}m_i$. This
replaces \citet{zweiger2026fast}'s NNLS fit, which FAST solves on thousands of
reference queries per KV-head per context---a reservoir cap of $50{,}000$,
averaging ${\sim}16{,}000$ per head on QuALITY. Refit train-free on the $w{=}8$
observation queries a single prefill affords, that regression can drive
\emph{in-window} error to zero by memorizing them, precisely the failure mode
the held-out decode replay is built to catch (cf.\ repeat-prefill in \Cref{app:negative}); the closed form has no
fit to overfit, and offline replay measured $11$--$14\%$ attention-output
error reduction for absorption at $B{=}128$--$512$ against $-1.5\%$ for the
bias alone, as \Cref{app:negative} logs.

\section{The sixth backbone: Llama-3-8B and the minimal steering dose}
\label{app:llama3}

We repeat the main experiment on Llama-3-8B-Instruct, the predecessor of
our primary backbone. The two models share their entire architecture and
KV geometry---so \dg{} operates on an identical per-head cache---and differ
only in long-context adaptation: Llama-3.1 adds a $16\times$ longer context
via RoPE scaling and a stronger, chattier instruction-tuning round,
itemized in \Cref{tab:backbones}. This isolates the effect of the backbone's
generation behavior on the deferral.

\input{sections/table_backbones}

\paragraph{The minimal steering dose is backbone-dependent.}
On Llama-3 the \emph{single}-step deferral ($k{=}2$) that erases the gap on
Llama-3.1 needs a longer draft: SKV+\dg$_{2}$ recovers $+2.1$ of the
$5.8$-point gap at $B{=}64$ (already the best compression method in the
block) and trails \snapkv{} by $0.77$ at $B{=}128$. Crucially this is not degeneration---the
anchored openings are fluent and often \emph{more} complete answers---but a
\emph{longer draft recovers the gap}: the dose climbs monotonically
$39.73\!\to\!40.88\!\to\!42.09\!\to\!42.38\!\to\!42.73$ at
$k{=}2/4/8/16/32$. There is no knee---each extra anchor token buys a
little more, with $79\%$ of the total climb done by $k{=}8$ and the
remaining $0.64$ spread over $k{=}8\!\to\!32$---and the ramp only
overtakes \snapkv{} itself between $k{=}2$ ($-0.77$ against the $40.50$
baseline) and $k{=}4$ ($+0.38$). SKV+\dg$_{32}$ ($42.73$) is the best compression method in the block,
above SnapKV $40.50$ and recovering $72\%$ of the $3.08$-point gap to
FullKV $43.58$ (large per-dataset jumps: Qasper $35.1\!\to\!42.5$, SAMSum
$39.0\!\to\!42.7$). Two consequences: the headline
$k{=}2$ result is a property of decisive-opening backbones (the modern
long-context generation), and the \S\ref{sec:audit} analysis is unaffected,
having been measured on Llama-3.1. An answer-length-controlled rescore
separates the verbosity artifact from the real deficit on the worst case,
TriviaQA: SKV$+$\dg$_2$'s first-line score is $75.8$ against $89.8$
(\snapkv{}) and $90.6$ (\dg$_{16}$)---the $k{=}2$ draft leaves
Llama-3 mid-preamble, and its chattier full-sentence openings (``United States
of America.'' where the reference is ``United States'') are scored down by
extractive F1. Truncating each first line to the longest reference length
($+2$ words) recovers $+5.9$ for \dg$_2$ ($75.8\!\to\!81.7$) while moving
every other method $\le\!1.0$ and Llama-3.1's \dg$_2$ by $0.1$
($92.5\!\to\!92.6$): roughly $40\%$ of the deficit is phrasing, and the
residual $\sim$$9$ points is what \Cref{tab:llama3}'s longer drafts close
($90.3$ at $k{=}8$).

\paragraph{What the dose actually measures.}
A larger $k$ changes two things at once: the scorer sees more real decode
queries, and more tokens are generated before the cut. Three probes
separate them, and only the second of the two survives.

\emph{(i) It is not a slower start.} The natural reading of ``longer
steering segment'' is that Llama-3 takes more tokens to reach its answer.
It does not. Scoring the FullKV generations of both backbones for the token
position at which a gold answer string first appears---six extractive
LongBench sets, $505$ and $550$ predictions in which a gold string appears
at all---gives a median onset of $0$ tokens on \emph{both}, with $92\%$
(Llama-3) and $98\%$ (Llama-3.1) of answers begun within two tokens. The
chattier phrasing documented above is real, but it does not delay the
answer.

\emph{(ii) It is not slower selection convergence.} If the dose bought
scoring information, Llama-3's chosen tokens should keep moving for longer.
We replayed \dg{}'s own scoring path---mean softmax attention of the first
$k$ draft queries over past keys, group-meaned across GQA heads,
per-head top-$B$---and measured the overlap between the set kept at $k$ and
the set kept at $k{=}32$, over six datasets ($63$ and $64$ prompts).
The two backbones are indistinguishable: $0.605$ (Llama-3) vs $0.611$
(Llama-3.1) at $k{=}2$, and $0.858$ vs $0.838$ at $k{=}16$. Selection
converges at the same rate on both.

\emph{(iii) It is consistent with anchoring.} The discriminating run is the timing-only
control at the same doses. \dg{}-W$_{2}$, \dg{}-W$_{16}$ and \dg{}-W$_{32}$ keep
\snapkv{}'s window scores---the draft contributes \emph{no} scoring
information whatsoever---yet they climb $39.81\!\to\!42.19\!\to\!42.61$ at
$k{=}2/16/32$, landing within $0.08$, $0.19$ and $0.12$ of full \dg{} at each dose
in \Cref{tab:llama3}. The dose does not buy a better eviction signal; it
buys tokens generated on the full cache before the cut. Llama-3's
trajectory simply takes longer to become robust to eviction, which is a
statement about \emph{when} the cut can safely happen, not about
\emph{what} it should keep. Even the one parameter that looked like a scoring
choice turns out to be timing.

% fig:present is declared here (before tab:llama3) only to control float
% stacking: both land on the same page and LaTeX stacks [t] floats in
% declaration order, so the figure renders ABOVE the table.
\begin{figure}[htbp]
\centering
\begin{subfigure}{0.49\textwidth}
  \centering\includegraphics[width=\textwidth]{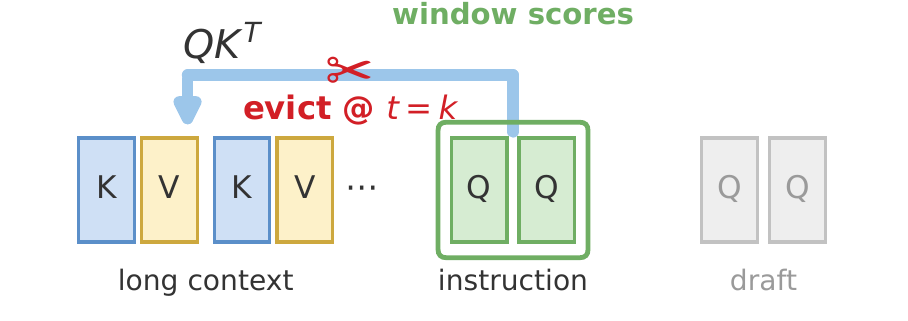}
  \caption{\snapkv{}+\dg{}-W$_{k}$: window scores, deferred cut.}
  \label{fig:present-dgew}
\end{subfigure}
\hfill
\begin{subfigure}{0.49\textwidth}
  \centering\includegraphics[width=\textwidth]{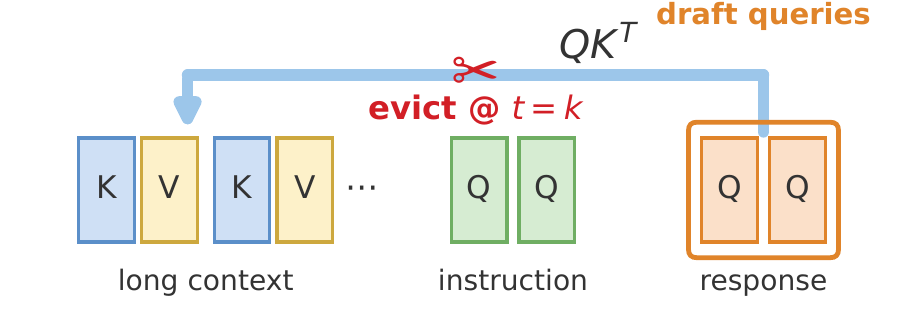}
  \caption{\snapkv{}+\dg$_{k}$: real draft queries, deferred cut.}
  \label{fig:present-dge}
\end{subfigure}
\caption{\textbf{\dg{}-W vs \dg{}: same \emph{when}, different \emph{what}},
drawn in \Cref{fig:concept}'s visual language. The two variants differ
\emph{only} in the scoring source; the draft is generated on the full cache
either way, though faded in~(a).}
\label{fig:present}
\end{figure}

\input{sections/table_llama3}

\section{Draft-guided eviction implementation}
\label{app:dg}

\begin{algorithm}[htbp]
\caption{Draft-Guided Eviction with its two scorers: \dg$_{k}$ (draft queries,
the default) and the timing-only \dg{}-W. Both wrap any prefill-time evictor $E$ (retention rule $R_E$, observation
window $w$), draft $k$ tokens on the full cache (\textbf{A}), and evict at
$t{=}k$ (\textbf{B}--\textbf{C}), differing only at line~\textbf{B}
(cf.\ \Cref{fig:present}). The baseline $E$
instead evicts at $t{=}0$.}
\label{alg:dge}
\small
\begin{algorithmic}[1]
\REQUIRE prompt $x_{1:L}$, per-head budget $B$, draft length $k$,
         base evictor $E$, generation budget $T$
\ENSURE answer $y$
\STATE $\mathcal{C},\ y_1,\ \{q_1^{(\ell,h)}\} \leftarrow \mathrm{prefill}(x_{1:L})$
       \COMMENT{full cache --- the peak $E$ also holds; prefill emits $y_1$}
\FOR{$t = 2$ \TO $\min(k, T)$}
    \STATE $y_t,\ \{q_t^{(\ell,h)}\} \leftarrow \mathrm{decode}(\mathcal{C})$;\quad
           append $K,V$ of $y_t$
           \COMMENT{\textbf{A: draft} + capture real queries}
    \IF{$y_t = \mathrm{EOS}$}
        \RETURN $y_{1:t}$ \COMMENT{answer ended; cache is freed --- never evict}
    \ENDIF
\ENDFOR
\FOR{layer $\ell = 1$ \TO $N$,\ head $h = 1$ \TO $H$}
    \STATE $s \leftarrow \tfrac{1}{k}\sum_{t \le k}
           \mathrm{softmax}\big(q_t^{(\ell,h)} K^{(\ell,h)\top} / \sqrt{d}\,\big)$
           \COMMENT{\textbf{B: score} --- \dg{}: the $k$ real draft queries}
    \STATE \hspace{1.1em}\emph{or}\ \ $s \leftarrow s_E^{(\ell,h)}$
           \COMMENT{\dg{}-W: $E$'s window scores --- \emph{only this line differs}}
    \STATE $I \leftarrow R_E\big(s,\ B - w\big)$
           \COMMENT{top-$B$ (SKV), layer budgets (PKV), sinks (SLM)}
    \STATE $\mathcal{C}^{(\ell,h)} \leftarrow
           \mathcal{C}^{(\ell,h)}\big[\,I \cup \mathrm{window} \cup \mathrm{draft}\,\big]$
           \COMMENT{one in-place edit, temporal order kept}
\ENDFOR
\FOR{$t = k{+}1$ \TO $T$ \textbf{or} EOS}
    \STATE $y_t \leftarrow \mathrm{decode}(\mathcal{C})$;\quad append $K,V$
           \COMMENT{\textbf{C: continue} at $E$'s decode memory}
\ENDFOR
\RETURN $y$
\end{algorithmic}
\end{algorithm}
\paragraph{Implementation notes.} \Cref{alg:dge}'s prototype is two stock
\texttt{generate()} calls around one cache edit. \emph{Phase A} lifts every
layer's \texttt{max\_capacity\_prompt} to $2^{30}$ (compression disabled) and
registers forward pre-hooks that recompute post-RoPE queries from each layer's
input: per-step decode queries (the eviction evidence), the last-$w$ prompt
queries (for the \dg{}-W control), and, for the H2O base only, all prompt
queries (its accumulator; freed per layer as consumed, $\sim$60\,MB/layer).
\emph{Phase B} edits the cache in place, per layer: score, top-$B$,
\emph{sort indices to preserve temporal order}, then gather
[retained past]\,$\|$\,[window $+$ draft tail]; the continuation call resumes
from the edited cache (retained tokens' RoPE positions are already baked into
their cached keys, so no re-indexing is needed).
\emph{Early termination:} if the draft ends (EOS, or fewer than $k$ tokens
produced), phase A's output is returned as-is and no eviction ever fires; this
is the FullKV-identical path whose per-dataset frequency \Cref{tab:term}
accounts. Latency and memory of the two-phase prototype are measured in
App.~\ref{app:complexity}: $+9.9\%$ over \snapkv{} end-to-end, flat in $k$,
with TTFT $0.93\times$.

\subsection{Why deferral can recover FullKV: an eviction-mass argument}
\label{app:whymath}
Write the per-head decode output at step $t$ as $o_t=\sum_{j} a_{t,j}\,v_j$
with weights $a_{t,j}=\mathrm{softmax}_j(q_t^{\top}k_j/\sqrt d)$. Split the keys
into the retained set $R$ and the evicted set $E$, and let
$m_t(E)=\sum_{j\in E} a_{t,j}$ be the \emph{evicted mass}---the attention the
step-$t$ query places on discarded tokens. Renormalizing over $R$ gives the
compressed output $\tilde o_t=\sum_{j\in R}\tfrac{a_{t,j}}{1-m_t}v_j=\mu_R^t$;
writing $\mu_E^t,\mu_R^t$ for the attention-weighted value means over $E,R$ we
have $o_t=(1-m_t)\mu_R^t+m_t\mu_E^t$, hence
\begin{equation}
\|o_t-\tilde o_t\| \;=\; m_t(E)\,\bigl\|\mu_E^t-\mu_R^t\bigr\|
\;\le\; 2\,m_t(E)\,\rho_v,\qquad \rho_v=\max_j\|v_j-\bar v\| .
\label{eq:evict-mass}
\end{equation}
The per-step error is \emph{linear in the evicted mass} $m_t(E)$ and vanishes
when $m_t(E){=}0$. Summing over decoding (errors accumulate to first order as
each output feeds the next step), the end-to-end deviation is bounded by
$\sum_t m_t(E)\,\|\Delta\mu_t\|$, and the eviction \emph{time} sets the
summation range:
\begin{equation}
\mathcal{L}_{\snapkv}\!\approx\!\!\sum_{t=1}^{T}\! m_t(E_w)\|\Delta\mu_t\|
\ \ (\text{cut at }t{=}0),\qquad
\mathcal{L}_{\dg}\!\approx\!\!\sum_{t=k+1}^{T}\! m_t(E)\|\Delta\mu_t\|
\ \ (\text{cut at }t{=}k).
\label{eq:loss-range}
\end{equation}
Only the per-step bound \eqref{eq:evict-mass} is exact; three observations
then suggest an ordering.
\textbf{(i)~Termination.} A dataset-dependent fraction $\rho$ of answers
finishes within $k$ tokens ($T\!\le\!k$)---$11.2\%$ overall at $k{=}2$, up to
$92.5\%$ on passage\_count in \Cref{tab:term}; for those the sum in
$\mathcal{L}_{\dg}$ is empty, so $\dg$ and $\dg$-W equal FullKV \emph{exactly}.
\textbf{(ii)~Anchoring.} For $T{>}k$, the $k$ tokens $y_{1:k}$ are produced on
the full cache and their KV is \emph{kept} (recent), so later queries commit to
the context those tokens select and $m_t(E)$ decays for $t{>}k$; the deferred
terms in \eqref{eq:loss-range} are small even for the window-chosen set $E_w$
(this is $\dg$-W).
\textbf{(iii)~Signal.} $\dg$ further picks $E$ to minimize
$\sum_{t\le k} m_t(E)$ from the \emph{real} draft queries, which at $k{=}16$
(\dg$_{16}$) cover
$+16.5\%$ more held-out decode mass than the window estimate
(\S\ref{sec:audit-sel}); this shrinks $m_t(E)$ for $t{>}k$ a little more, but by
(i)--(ii) the residual is already small, so the gain over $\dg$-W is second
order. Together they suggest
$\mathcal{L}_{\snapkv}\ge\mathcal{L}_{\dg\text{-W}}\!\approx\!\mathcal{L}_{\dg}
\ge 0=\mathcal{L}_{\mathrm{FKV}}$, i.e.\ SKV $<$ $\dg$-W $\approx$ $\dg$
$\approx$ FullKV, consistent with the measured $42.2 < 46.17\!\approx\!46.19 < 46.24$
at $k{=}16$ (\dg{}-W$_{16}$ and \dg$_{16}$); at the headline $k{=}2$ both deferred rows edge past FullKV, by
$0.09$ and $0.12$ in \Cref{tab:ci}'s paired bootstrap, a margin it does not
resolve from zero.

\paragraph{Why $\dg$ and $\dg$-W barely differ.}
The two share every step and differ only in the evicted set, so their loss gap
is $\mathcal{L}_{\dg\text{-W}}-\mathcal{L}_{\dg}=\sum_{t>k}\bigl[m_t(E_w)-m_t(E)\bigr]\|\Delta\mu_t\|$.
Two effects make this doubly small. First, by anchoring~(ii) both $m_t(E_w)$
and $m_t(E)$ are already small for $t{>}k$, so each summand is a difference of
two small numbers. Second, the window set $E_w$ and the draft-query set $E$
\emph{largely overlap}: both keep the salient tokens and disagree only on the
residual $+16.5\%$ of decode mass ($k{=}16$), which anchoring has already made
negligible after step $k$. The gap is thus second order in the deferred mass,
and $\dg$-W tracks $\dg$ to within measurement noise ($46.17$ vs $46.19$
for \dg{}-W$_{16}$ and \dg$_{16}$; $\dg$-W even edges $\dg$ at $B{=}32$ and $k{=}2$, $46.40$ vs $46.39$ for \dg{}-W$_{2}$ and \dg$_{2}$). In one
line: once the cut is deferred, \emph{which} tokens are dropped barely
matters---the timing has already removed the error that selection was fighting over.

\section{Draft-like passes compared}
\label{app:draftlike}

Several training-free compression methods insert an extra forward pass
between prefill and generation because the observation window is a weak
estimate of the queries the answer will actually ask. \Cref{tab:draftlike}
compares them along three design choices: what the pass generates, whether
its compute survives into the output, and which axis its product feeds.
Repeat-prefill \citep{kim2025kvzip} replays the prompt and scores keys by
reconstruction; self-study \citep{zweiger2026fast} generates synthetic
Q\&A about the context to calibrate biases and values, at an up-front
per-context cost; Lookahead Q-Cache \citep{ge2025lookahead}
synthesizes $m$ pseudo future queries and re-scores retention with them.
All three discard the pass afterwards, and all three still cut at prefill
end: the pass only sharpens \emph{what} is kept. Speculative decoding
\citep{leviathan2023speculative} appears because \dg{} is mechanically
closer to it than to the eviction scorers---a draft generated by the model
itself, kept when it is right---but it targets latency, not memory.
LookaheadKV \citep{ahn2026lookaheadkv} is the trained sibling of the same
\emph{what}-axis move, predicting lookahead scores draft-free; we compare
training-free passes only.

\dg{} breaks both invariants at once. Its pass is the answer's own first
$k$ tokens, so the compute is not an overhead to amortize but output the
user receives either way; and its product is not a better score but a later
cut. The ablations say the second difference is the one that pays: with
\snapkv{}'s unmodified window scores, the timing-only \dg{}-W stays within
$0.44$ of full \dg{} in every cell of \Cref{tab:main}, while LAQ---a
strictly richer scoring signal evicting at the old time---saturates
$\ge$2 points below the pure timing change at its best dose in
\Cref{tab:lookahead}. Sharpening \emph{what} without moving \emph{when}
runs into a ceiling that one deferred step clears.

\begin{table}[htbp]
\centering\small
\caption{\textbf{Draft-like passes in training-free KV compression:}
repeat-prefill (KVzip; \citealp{kim2025kvzip}), self-study (FAST;
\citealp{zweiger2026fast}), Lookahead Q-Cache \citep{ge2025lookahead},
speculative decoding \citep{leviathan2023speculative}.}
\label{tab:draftlike}
\vspace{2pt}
\adjustbox{max width=\textwidth}{%
\begin{tabular}{l l l l l}
\toprule
Extra pass & Queries & Kept? & Evicts & Axis moved \\
\midrule
Repeat-prefill & reconstruction & no & prefill end & what (scores) \\
Self-study & generated Q\&A & no$^{a}$ & prefill end & what ($\beta$, values) \\
Lookahead Q-Cache & pseudo future & no & prefill end & what (scores) \\
Speculative decoding & draft tokens & if verified & --- & latency \\
\textbf{\dg{} (ours)} & first $k$ answer tokens & yes (answer prefix) & step $k$ & \textbf{when} (scorer: free slot) \\
\bottomrule
\end{tabular}}
\vspace{2pt}
\begin{flushleft}\footnotesize
$^{a}$Amortizable only under prefix-cache reuse.
\end{flushleft}
\end{table}

\section{Complexity and overheads}
\label{app:complexity}
\Cref{tab:complexity} reports, per method, when it evicts, the extra
scoring compute, peak/decode KV memory, and measured per-sample latency
relative to \snapkv{}. All evictors share $O(L)$ \emph{asymptotic} peak memory (the end-of-prefill
cache) and $O(B{+}t)$ decode memory; \dg{} changes only the eviction time,
holding the full cache for $k$ extra decode steps before dropping to the same
steady state.

\paragraph{Where the overhead actually goes.} To separate the method's
intrinsic cost from our prototype's, we ran a controlled microbenchmark
(Llama-3.1-8B, $B{=}128$, a fixed $7{,}762$-token prompt, $256$ new tokens,
$1$ warm-up $+$ $3$ timed repeats on one A100-class GPU; \Cref{tab:complexity}'s
Latency column comes from the separate LongBench probe described there and is
not directly comparable).
Three things follow.

\emph{(i) The dose is free.} Total latency is flat in $k$: $10.88$s, $10.49$s
and $10.49$s at $k{=}2,4,8$---a $3.7\%$ spread with no monotone trend, against
a run-to-run $\sigma$ of up to $0.23$s. Quadrupling the draft costs nothing
measurable, so the $k$ full-cache decode steps are \emph{not} the cost driver.
The overhead over \snapkv{} ($+0.96$s, $+9.9\%$ here) is therefore the
\emph{one-off} per-layer cache surgery.

\emph{(ii) TTFT improves.} Because the draft runs before any eviction, the
first token is produced on the uncompressed cache and the compression never
sits on the prefill critical path: TTFT is $0.93\times$ \snapkv{}'s, essentially
FullKV's ($0.91\times$). H2O is the opposite case---its $O(L^2)$ accumulator
runs during prefill, giving a $3.45\times$ TTFT that its total latency hides
whenever the generation is long ($2.15\times$ on short answers, $1.06\times$ on
long-form). LAQ is the same case for the same reason: its $m$ pseudo
queries come from an extra forward pass that is discarded, so the whole pass
lands before the first token. Measured under this protocol in LAQ's own code
base ($m{=}16$, $1$ warm-up $+$ $5$ repeats), it is $1.42\times$. That code
base's \snapkv{} is not ours ($1.068$ vs $1.170$\,s), so we anchor on FullKV,
unpatched in both and agreeing to $0.2\%$ ($1.0593$ vs $1.0616$\,s): LAQ is
$1.561\times$ FullKV there and FullKV is $0.907\times$ \snapkv{} here.
Dividing LAQ straight by our \snapkv{} gives $1.41\times$, so the anchoring
changes little. \dg{}'s draft is the answer's opening, and is not discarded.

\emph{(iii) The real price is memory, and it is layout, not deferral.}
Separating the $14.96$\,GiB of weights from the working set (KV-Cache $+$
activations), the peak is $5.69$\,GiB for \snapkv{}, $6.58$\,GiB for FullKV and
$13.13$\,GiB for \dg$_2$---$2.0\times$ FullKV's working set, and
dose-independent ($13.126$ vs $13.132$\,GiB at $k{=}2$ and $k{=}8$), confirming
(i): the cost is the one-off draft machinery, not holding the cache longer.
The dominant term is the \emph{cache layout}: to score eviction per head, this
codebase repeats KV to all $32$ query heads \emph{before} caching, so the
uncompressed prefill cache is $3.79$\,GiB where the native $8$-head GQA layout
stores $0.95$\,GiB. Every evictor pays this constant, but only \dg{} pays it at
full prompt length, because only \dg{} holds the cache uncompressed through the
draft.

LAQ shows the same cost from the same cause. Holding the generated count
fixed at $256$ on this prompt, so every method does equal work, its working
set is FullKV's, not \snapkv{}'s: $7.60$ against FullKV's $7.62$ and
\snapkv{}'s $5.68$\,GiB. The lookahead pass runs before any eviction, so the
peak is taken on the uncompressed cache---the same structural reason its TTFT
is $1.42\times$. (These three are measured with the count pinned, which
FullKV's uncapped cache grows with, so its $7.62$ is not the $6.58$ above;
\snapkv{}'s $5.68$ and \dg{}'s $13.11$ reproduce the $5.69$ and $13.13$
either way.)

\begin{figure}[htbp]
\centering
\begin{subfigure}{0.49\textwidth}
  \centering\includegraphics[width=\textwidth]{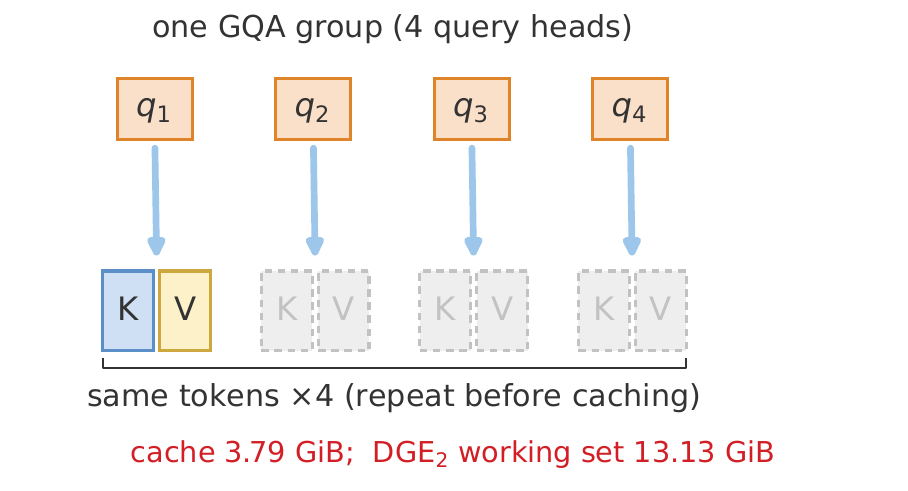}
  \caption{Per-head: KV repeated $\times4$ before caching.}
  \label{fig:kvlayout-perhead}
\end{subfigure}
\hfill
\begin{subfigure}{0.49\textwidth}
  \centering\includegraphics[width=\textwidth]{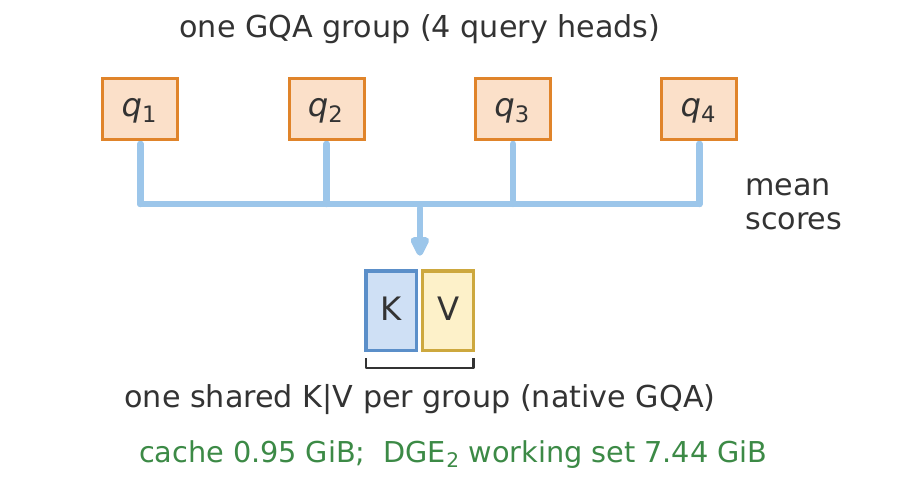}
  \caption{\texttt{kv\_head}: one shared K$|$V, mean scores.}
  \label{fig:kvlayout-kvhead}
\end{subfigure}
\caption{\textbf{The memory constant is a layout choice} (microbenchmark
numbers).}
\label{fig:kvlayout}
\end{figure}

\emph{Removing it, measured.} We implemented the fix sketched in
\Cref{fig:kvlayout}(b): cache in the native GQA
layout, score eviction with the captured query-head queries, and mean the
scores over each group of $4$ query heads (the codebase's
\texttt{gqa\_score\_agg} convention). \dg$_2$'s working set drops
$13.13\!\to\!7.44$\,GiB---from $2.0\times$ to $1.13\times$ FullKV---%
exceeding the $+2.84$\,GiB layout term because FlashAttention-2 consumes GQA
natively, so the $32$-head expansion disappears from the \emph{compute} path as
well as the cache. The result is base-independent (PKV$+$\dg$_2$: identical
$13.126$/$7.439$\,GiB in the two layouts) and score-neutral on a spot check
(Llama-3.1, $B{=}128$: Qasper $-0.06$, TREC $\pm0.00$, SAMSum $-0.34$---all
inside the $\pm0.5$ no-effect band). The full-suite numbers elsewhere in this
paper remain in the query-head layout, the one every baseline also uses, so all
comparisons stay matched; the asymptotics ($O(L)$ peak) were never at issue,
and the constant is now demonstrated---not merely argued---to be a layout
choice rather than a property of deferral.

\input{sections/table_complexity}

\section{Attention-matching loss}
\label{app:matchloss}

\paragraph{Definition.} Fix a layer and head. A compression keeps a retained
set $R$ of past keys and drops $E$; let $q$ range over the \emph{observation}
queries---in \snapkv{} the $w$ window queries at the end of the prompt, used as
an estimate of the not-yet-seen decode queries. Write the full-attention
log-partition and its retained restriction as
\begin{equation}
  \log Z(q) = \log\!\!\sum_{j\in R\cup E}\! e^{q^\top k_j/\sqrt d},
  \qquad
  \log \hat Z(q) = \log\!\!\sum_{j\in R}\! e^{q^\top k_j/\sqrt d}.
\end{equation}
The training-free attention-matching loss is the mean log-partition deficit,
which we measure directly at compression time,
\begin{equation}
  \mathcal{L}_{\mathrm{match}}
   \;=\; \mathbb{E}_q\,\bigl|\log Z(q)-\log \hat Z(q)\bigr|
   \;=\; \mathbb{E}_q\,\bigl[-\log\bigl(1-\rho_q\bigr)\bigr],
   \quad
   \rho_q=\frac{\sum_{j\in E} e^{q^\top k_j/\sqrt d}}
               {\sum_{j\in R\cup E} e^{q^\top k_j/\sqrt d}}
   \in[0,1),
   \label{eq:match}
\end{equation}
where $\rho_q$ is exactly the softmax mass the observation query places on the
evicted keys---the observation-query analogue of the evicted mass $m_t(E)$ of
\Cref{eq:evict-mass}. So the loss is a monotone $-\log(1-\cdot)$ transform of
the dropped attention mass: it is zero iff the retained set carries all of
$q$'s attention and diverges as the evicted mass approaches one. Minimizing it
over $R$ under a size budget is the window-query counterpart of FAST's
mass-matching objective \citep{zweiger2026fast}, the one its NNLS fit solves; \snapkv{}'s top-$B$ selection
on the pooled window-query scores is its greedy solution, and the per-token
bias $\beta$ that FAST fits to reweight the retained logits shifts $\log\hat Z$
by exactly the recoverable part of \eqref{eq:match} (a run with $\beta$ closes
the deficit to $\mathbb{E}_q|\log Z-\log(\hat Z\,e^{\beta})|$).

\begin{figure}[htbp]
\centering
\includegraphics[width=0.60\textwidth]{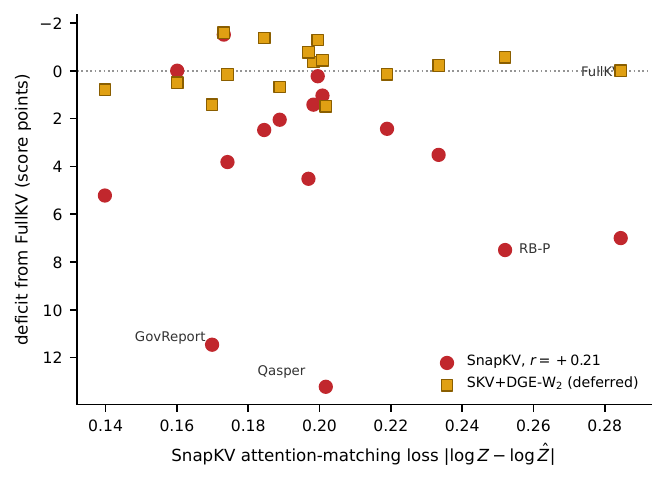}
\caption{\textbf{The attention-matching loss does not predict the downstream
deficit.} One point per LongBench set ($B{=}128$, Llama-3.1-8B), loss as in
\eqref{eq:match}; \dg{}-W$_2$ defers \snapkv{}'s identical selection by one
step.}
\label{fig:matchloss}
\end{figure}

\paragraph{The loss does not predict the downstream deficit.} The crucial point
for this paper is that \eqref{eq:match} is an \emph{observation-query, prefill-time}
quantity: it scores how well $R$ reconstructs attention for the window queries,
on the full cache, before any decoding. \Cref{eq:loss-range} shows the error
that actually accrues is $\sum_t m_t(E_t)\|\Delta\mu_t\|$ over the \emph{real}
decode queries $q_t$, whose weight is set by \emph{when} $E$ is dropped, not by
how well the prefill observations were matched. \Cref{fig:matchloss} makes this
quantitative at scale on Llama-3.1-8B ($B{=}128$). Across the sixteen sets the
loss varies over a narrow $2\times$ band ($0.14$--$0.28$) yet \snapkv{}'s
deficit from FullKV ranges from $-1.5$ to $13.2$ points, and the two are
essentially uncorrelated (Pearson $r{\approx}0.2$, i.e.\ the loss explains
$\sim4\%$ of the deficit variance): the sets \snapkv{} hits hardest---Qasper
($13.2$) and GovReport ($11.5$)---sit at \emph{low-to-mid} loss, while the
highest-loss set (TREC, $0.28$) loses only $7$. Keeping the identical selection
and evicting one decode step later (\dg{}-W$_2$) drives the deficit to $-0.09$ on
average (worst case $1.5$) at \emph{every} loss. The
quantity the \emph{what}-axis optimizes is orthogonal to the error the
\emph{when}-axis removes.

\section{Confidence intervals and significance}
\label{app:ci}
All decoding is greedy, so re-running a configuration reproduces it bit for bit
and there is no seed variance to report. The uncertainty that does exist is
sampling over the $3{,}750$ LongBench examples. We bootstrap it ($2000$
resamples): within each dataset we resample examples with replacement,
preserving dataset sizes, recompute each dataset score and macro-average over
the 16 sets---exactly how the headline number is formed.

Comparisons use a \emph{paired} bootstrap: both methods are scored on the same
resample, which is legitimate because every method here is run on identical
example sets, and is far sharper than comparing two marginal intervals. The
distinction matters: the marginal intervals are $\pm1.1$ (dominated by which
datasets are drawn), so read naively they would suggest \emph{nothing} in
\Cref{tab:main} is resolvable, whereas the paired intervals on differences are
$\sim\pm0.3$.

\begin{table}[htbp]
\centering\small
\setlength{\tabcolsep}{5pt}
\caption{\textbf{Paired bootstrap} ($2000$ resamples of the $3{,}750$
examples): Llama-3.1-8B at $B{=}128$ and Qwen2.5-14B at all three budgets
(the 14B SKV separations, $+15.8/{+}9.5/{+}5.8$, all resolve and are omitted
for space). SKV = \snapkv{} \citep{li2024snapkv}, FKV = FullKV;
\dg$_2$ and \dg{}-W$_2$ are the two variants \S\ref{sec:method} defines.
\emph{Resolvable} means the $95\%$ interval excludes zero. Shading here marks
the signal-only contrast this paper's attribution turns on, not the ``ours''
shading of the result tables.}
\label{tab:ci}
\vspace{2pt}
\begin{tabular}{lrr l}
\toprule
Comparison & $\Delta$ & 95\% CI & Resolvable? \\
\midrule
\multicolumn{4}{l}{\emph{Llama-3.1-8B, $B{=}128$}} \\
SKV+\dg$_2$ $-$ SKV            & $+4.15$ & $[+3.60,+4.70]$ & yes \\
SKV+\dg{}-W$_2$ $-$ SKV        & $+4.11$ & $[+3.61,+4.67]$ & yes \\
\hi SKV+\dg$_2$ $-$ SKV+\dg{}-W$_2$ & $+0.04$ & $[-0.19,+0.26]$ & \textbf{no} \\
SKV+\dg$_2$ $-$ FKV            & $+0.12$ & $[-0.16,+0.44]$ & no \\
SKV+\dg{}-W$_2$ $-$ FKV        & $+0.09$ & $[-0.19,+0.38]$ & no \\
\midrule
\multicolumn{4}{l}{\emph{Qwen2.5-14B, $B{=}32$}} \\
\hi SKV+\dg$_2$ $-$ SKV+\dg{}-W$_2$ & $-0.44$ & $[-0.76,-0.13]$ & yes (control ahead) \\
SKV+\dg$_2$ $-$ FKV            & $-0.30$ & $[-0.66,+0.06]$ & no \\
SKV+\dg{}-W$_2$ $-$ FKV        & $+0.13$ & $[-0.21,+0.47]$ & no \\
\midrule
\multicolumn{4}{l}{\emph{Qwen2.5-14B, $B{=}64$}} \\
\hi SKV+\dg$_2$ $-$ SKV+\dg{}-W$_2$ & $-0.37$ & $[-0.64,-0.11]$ & yes (control ahead) \\
SKV+\dg$_2$ $-$ FKV            & $-0.11$ & $[-0.45,+0.21]$ & no \\
SKV+\dg{}-W$_2$ $-$ FKV        & $+0.26$ & $[-0.06,+0.58]$ & no \\
\midrule
\multicolumn{4}{l}{\emph{Qwen2.5-14B, $B{=}128$}} \\
\hi SKV+\dg$_2$ $-$ SKV+\dg{}-W$_2$ & $+0.10$ & $[-0.18,+0.37]$ & \textbf{no} \\
SKV+\dg$_2$ $-$ FKV            & $+0.23$ & $[-0.10,+0.57]$ & no \\
SKV+\dg{}-W$_2$ $-$ FKV        & $+0.13$ & $[-0.20,+0.44]$ & no \\
\bottomrule
\end{tabular}
\end{table}

This sharpens the paper's central claim. On Llama-3.1 the paired test says
\dg$_2$ and \dg{}-W$_2$ are \emph{not distinguishable at all} on $3{,}750$
examples, while both separate from \snapkv{} by $\sim$$4.1$ with intervals
nowhere near zero. Qwen2.5-14B is stricter still: at $B{\le}64$ the
difference between the two deferred variants \emph{does} resolve, and it
favors the timing-only control by $0.4$---swapping \snapkv{}'s window
scores for real draft queries adds no signal and, on this backbone, costs
$0.4$---while both variants stay indistinguishable from FullKV at
every budget. The \emph{when} axis carries a gain that survives resampling;
the \emph{what} axis never produces one, on either backbone.

\section{Needle-in-a-Haystack retrieval}
\label{app:needle}
\begin{figure}[htbp]
\centering
\includegraphics[width=0.86\textwidth]{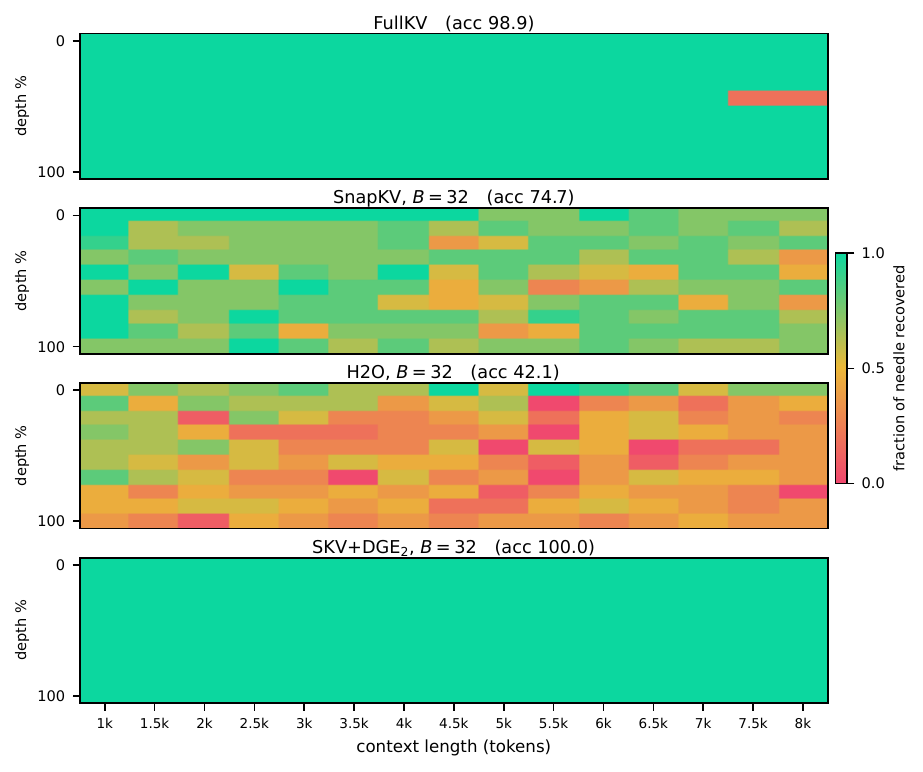}
\caption{\textbf{Needle-in-a-Haystack at $B{=}32$ (Qwen2.5-7B).} $150$ probes
per panel, with the retrieval question visible in the prompt at compression
time.}
\label{fig:niah}
\end{figure}

The LongBench averages aggregate generation quality; \Cref{fig:niah} isolates
\emph{retrieval} under the same tight budget. We run the standard
Needle-in-a-Haystack probe on Qwen2.5-7B-Instruct at $B{=}32$: the needle
sentence is planted at $10$ depths across $15$ context lengths ($1$k--$8$k), for
$150$ probes per method, and a cell is scored by the fraction of the expected
answer's words the model recovers.

This is the regime most favorable to prefill-time eviction: the retrieval
question is in the prompt, so \snapkv{}'s window queries \emph{can} see which
tokens matter, and a query-aware selector should simply keep the needle. It does
not. At $B{=}32$ \snapkv{} recovers only $74.7$ against FullKV's $98.9$---it
drops needles across the whole depth$\times$length grid---and H2O, whose
accumulated-attention rule is not query-aware, falls to $42.1$. Deferring the
same \snapkv{} selection by a single decode step (\dg$_2$) restores $100.0$:
every needle, at every depth and length. The $25.3$-point recovery is bought
without changing which tokens are scored or how, only \emph{when} they are
dropped---the \Cref{tab:main} story reproduced on a task where the
\emph{what}-axis was supposed to be sufficient.

\section{Full result tables}
\label{app:tables}
The full 16-dataset breakdown of the main results (summarized as averages in
\Cref{tab:main}) is split for readability into the prefill-time baselines
of \Cref{tab:main-full} and the deferred-eviction rows of \Cref{tab:main-defer};
the complete SnapKV-variant dose grid and the lookahead-query comparison
follow.
\input{sections/table_main}
\input{sections/table_defer}
\input{sections/table_ablation}
\input{sections/table_lookahead}
\Cref{tab:category} aggregates the 16 sets into the six standard LongBench
categories, and \Cref{tab:term} reports, per dataset, how often the answer
finishes inside the $k$-token draft---together with the sample-count check.
Every configuration in these tables is run at LongBench's full official sizes
($200$ per set, $150$ for multifieldqa\_en, $500$ for lcc and repobench-p;
$3{,}750$ samples per configuration) with no subsampling; the diagnostic
runs behind the grids of App.~\ref{app:comp} are the exception that section
records. The only truncation is the standard per-backbone context limit of
App.~\ref{app:expdetails}.
\input{sections/table_category}
\input{sections/table_term}

\section{Additional analysis}
\label{app:analysis}

\paragraph{Draft-termination fractions and the two regimes.}
When an answer finishes inside the $k$-token draft the cache is freed at answer
end, no eviction ever fires, and \dg{} is bit-identical to FullKV---so it is
worth asking how much of \dg{}'s gain this path accounts for. Measuring
it directly in \Cref{tab:term}, re-tokenizing every saved answer, shows it is a
small share: at the headline dose $k{=}2$ only $11.2\%$ of the $3{,}750$
samples terminate in the draft, and termination is confined to short-answer QA
($5$--$40\%$) and passage\_count ($92.5\%$), with Summarization, Few-shot and
Code at $0\%$. Even at $k{=}16$ the overall figure is $34.0\%$. The gain
therefore cannot be a draft-termination artifact: $\sim$89\% of samples at
$k{=}2$ genuinely decode on the compressed cache, and \dg$_2$ still matches
FullKV ($46.37$ vs $46.24$). The cleanest case is long-form generation, where
the answer far exceeds $k$: on gov\_report ($512$-token generations, $0\%$
termination) $\sim$99\% of tokens decode on the compressed cache, so the
recovery ($22.3\!\to\!32.6$ ROUGE at $B{=}128$) is best explained by
trajectory anchoring.

\paragraph{Composition across retention rules.}
The single full-cache decode step transfers to the three other base evictors
we test, not just
\snapkv{}: at $B{=}128$, PKV$+$\dg$_2$ reaches $46.21$ (from PyramidKV's
$41.64$), H2O$+$\dg$_2$ reaches $45.73$ ($+7.6$), and StreamingLLM$+$\dg$_2$
reaches $45.83$---a $+9.3$ jump for a rule that keeps four sinks and a recency
window \emph{without any attention scores at all}. The timing-only compositions,
run in the same sweep but not tabulated, land within $0.35$ of their
draft-scored counterparts (PKV$+$\dg{}-W$_2$ $46.38$, H2O$+$\dg{}-W$_2$
$45.39$), confirming that the recovery is a property
of the eviction \emph{event} and not of any particular retention rule. The
recovery also holds at the tightest budget: at $B{=}32$, PKV/H2O/StreamingLLM$+$\dg$_2$
reach $46.42/45.70/46.23$, a $+12.8$ jump for StreamingLLM from $33.44$. The
same composition holds on a second backbone: on Qwen2.5-7B at $B{=}128$,
PKV$+$\dg$_2$ reaches $45.03$ (from PyramidKV's $38.38$, $+6.7$),
H2O$+$\dg$_2$ reaches $44.67$ (from $35.66$, $+9.0$), and
StreamingLLM$+$\dg$_2$ reaches $44.86$ ($+14.7$)---all within $0.9$ of
Qwen's FullKV $45.54$, per \Cref{tab:main-defer}.

\paragraph{Lookahead-query dose saturation.}
LAQ's pseudo-query dose helps but saturates $\sim$2 points below a
pure timing change. At $B{=}128$ the $m{=}4/8/16/32$ averages are
$43.81/44.13/44.20/44.10$---the curve peaks at $m{=}16$ and is flat (even
slightly down) at $m{=}32$, so spending more pseudo queries does not close the
gap to FullKV ($46.24$) or to \dg{}-W$_2$ ($46.33$). The same ordering holds at
$B{=}64$. The deficit against the timing-only control also \emph{widens} as the
budget shrinks---$1.0$--$2.1$ points at $B{=}128$ (Qwen2.5-14B: $1.7$) vs
$3.0$--$3.9$ at $B{=}64$ and $5.8$--$9.1$ at $B{=}32$---so tighter budgets make
\emph{when} matter more, not less. Sharpening the \emph{what}-signal has a
ceiling that a single deferral step clears outright, as
\Cref{tab:lookahead} shows.

\paragraph{What LAQ's pseudo queries are actually worth.}
\label{app:laqcov}
The dose curve says the \emph{what}-signal saturates; this probe says why. We
replay LAQ's two stages inside the selection harness of
\Cref{app:diagnostics}---compress the prefill cache to $B$ with \snapkv{}
(window $32$), generate $m{=}16$ tokens on that compressed cache keeping their
post-RoPE queries, then score the full prompt keys with the last $8+m$ rows,
exactly as our matched runs configure LAQ---and score the resulting selection
against the same held-out decode attention every other rung of the ladder is
measured on. LAQ's pseudo queries buy $+1.2\%$ covered mass over \snapkv{}'s
window estimate at $B{=}128$ ($+1.3\%$ at $B{=}64$; $12$ prompts), and $-3.1\%$
when the window rows are removed and the pseudo queries score alone. The real
draft queries \dg{} captures at the same dose ($k{=}16$, i.e.\ \dg$_{16}$) are worth $+18.7\%$ on the same
$12$ prompts, and the oracle $+30.5\%$---the two rungs \Cref{tab:audit}(b)
reports as $+16.5$/$+31.8\%$ on its own diagnostic split, re-measured here so
all three are scored on identical prompts. Generating the lookahead on a cache already
compressed to $B$ costs essentially all of the signal the draft was supposed to
provide. The pseudo continuation also leaves the real answer early---it agrees
with the full-cache answer for $5.1$ tokens on average, and on
\texttt{gov\_report}, where the compression gap lives, it diverges after one to
eight. This is the measurement behind LAQ's position in
\Cref{fig:headline}(a): it improves its own objective by almost nothing, yet
still gains $+2.0$ LongBench---a second, opposite reading of the
miscalibration the analysis reports, and the reason the timing-only control beside
it, with no signal change at all, gains twice as much.

\paragraph{The minimal steering dose is capability-, not scale-, dependent.}
The dose $k$ that recovers FullKV is backbone-dependent: Llama-3.1-8B (both
$B{=}32$ and $128$), Mistral-7B, Qwen2.5-7B, and the smaller Llama-3.2-3B all
saturate at $k{=}2$, whereas the older, shorter-context Llama-3-8B climbs
gradually in \Cref{fig:dose}(a) and needs $k{\approx}8$, analyzed in
App.~\ref{app:llama3}. That the $3$B model
still needs only $k{=}2$ while an $8$B model needs $k{\approx}8$ shows the dose
tracks the decisiveness of a backbone's opening tokens (how quickly its
instruction tuning commits to a trajectory), not parameter count. Scaling up
does not shrink the problem either: on Qwen2.5-14B---the largest backbone we
run---\snapkv{} at $B{=}32$ posts its largest LongBench-average collapse in the paper
($29.94$ vs FullKV $46.05$, $-16.1$), and the same $k{=}2$ deferral recovers
$45.74$ in \Cref{tab:main-defer}, within $0.31$ of FullKV.

\paragraph{PyramidKV's non-monotone Qwen columns are a discrete-schedule artifact.}
In \Cref{tab:main} PKV on Qwen2.5-7B scores \emph{lower} at $B{=}64$ ($30.74$)
than at $B{=}32$ ($32.69$), and Qwen2.5-14B repeats the inversion ($34.64$
vs $35.07$); the runs reproduce bit-identically under greedy
decoding, so it is systematic, and the cause is arithmetic. PKV allocates the
per-layer past budget on a linear ramp from $\mathrm{max}=2c-\lfloor
c/20\rfloor$ down by $\mathrm{steps}=\lfloor(\mathrm{max}-\mathrm{min})/
(L{-}1)\rfloor$ per layer, where $c=B-w$ and $L$ is the layer count. With
Qwen's $L{=}28$ and $B{=}64$ ($c{=}56$) the floor lands on
$\mathrm{steps}{=}4$, so the deepest layers keep only $[14,10,6,2]$ past
tokens---the last layer retains $2$ tokens beside the $8$-token
window---whereas at $B{=}32$ ($\mathrm{steps}{=}1$) the same layers keep
$[23,22,21,20]$: the \emph{smaller} budget leaves \emph{more} cache in the
deep layers, inverting the column. Qwen2.5-14B ($L{=}48$) is the extreme
case: at $B{=}32$ the floor collapses to $\mathrm{steps}{=}0$, so the ramp
degenerates and \emph{every} layer keeps the flat $\mathrm{max}{=}47$ past
tokens---nearly twice the nominal budget---while at $B{=}64$
($\mathrm{steps}{=}2$) the deepest layer keeps only $16$. On $L{=}32$
backbones the floor lands on $\mathrm{steps}{=}3$ at $B{=}64$ (last layer
$17$), so no inversion appears. Deferral is indifferent to this:
SKV$+\dg_{2}$ is budget-invariant on Qwen (7B: $45.05/45.32/45.32$; 14B:
$45.74/45.94/46.27$), so the artifact affects only the prefill-time
schedule, not the timing result.

\paragraph{Per-dataset view of timing vs pseudo queries.}
\Cref{fig:dose}(b) plots, per dataset at $B{=}128$, \snapkv{}'s score against
each method's; points above $y{=}x$ beat \snapkv{}. Both timing methods
(SKV$+$\dg$_2$, SKV$+$\dg{}-W$_2$) sit well above the diagonal and gain most on
the hard, low-\snapkv{} summarization and code sets (clearing LAQ by $+8.8$ on GovReport),
which they lift to FullKV level; Lookahead Q-Cache ($m{=}16$, prefill-time) hugs
the diagonal on exactly those sets, so its improvement is concentrated where
\snapkv{} was already adequate.

\section{What the Diagnostics Established}
\label{app:negative}
%Each alternative below was settled by the held-out probes of \Cref{app:diagnostics} in under a day of compute, before any LongBench run. We log them because the \emph{pattern} is the point: every entry improves an in-window quantity and fails on held-out decode behavior---the same miscalibration \S\ref{sec:audit} measures.

Before any LongBench evaluation, we used the held-out probes of \Cref{app:diagnostics} to identify the design choices that most reliably predict decode-time behavior. The results consistently favor signals that match held-out decoding dynamics, and provide direct evidence for the selection, transport, allocation, and absorption components used in our method.

\begin{itemize}
%\item \textbf{Repeat-prefill selection.} Re-forwarding the context to harvest ``future-like'' queries (the KVzip recipe, as a selection signal): $-13\%$ held-out coverage with mean run length $1.2$---repetition queries attend to the locally-next token, shredding selection into singletons. Killed by the decode-replay coverage probe. 

\item \textbf{Decode-aligned selection.} Repeat-prefill queries, following the KVzip recipe, provide a convenient future-like selection signal but achieve $-13\%$ held-out coverage with mean run length $1.2$. In contrast, the decode-replay probe favors selection signals that capture contiguous spans relevant to future decoding. This supports our use of decode-aligned coverage for selection.

%\item \textbf{Key-kernel mass-flow transport.} Predicting where evicted attention mass re-flows from a key-similarity kernel (instead of the empirical co-attention $T$ of \Cref{app:comp}): correlation with the true decode-time flow $\approx 0$. Killed by the offline transport-validation replay of \Cref{app:comp}.

\item \textbf{Empirical mass-flow transport.} Predicting evicted attention mass with a key-similarity kernel yields approximately zero correlation with the true decode-time flow. The offline transport-validation replay of \Cref{app:comp} instead supports the empirical co-attention transport matrix $T$, motivating its use for modeling how attention mass redistributes after eviction.

%\item \textbf{Layer-only allocation.} Re-allocating budget across layers without per-head granularity collapses the achievable held-out coverage gain from $+8.0\%$ (free per-head) to $+2.7\%$: the variance that matters lives across heads, not layers. The free per-head variant is what \Cref{fig:headline}(a) plots on its \emph{allocate} axis, under the family label \emph{cross-layer allocation}: at its $+8.0\%$ coverage headroom a full sixteen-set run scores $+0.60$ LongBench over \snapkv{}, still under a point. Killed by the allocation headroom probe, which bounds the axis as a whole: on the same held-out split a better scoring signal is worth $4.4$ points of $\betadec$ against $0.7$ for reallocating the per-head budget.

\item \textbf{Per-head budget allocation.} Allowing budgets to vary across heads provides substantially more held-out coverage headroom than layer-only allocation: $+8.0\%$ versus $+2.7\%$. The variance relevant to allocation therefore occurs primarily across heads. The free per-head variant used in \Cref{fig:headline}(a), under the \emph{cross-layer allocation} family, achieves this $+8.0\%$ headroom and improves LongBench by $+0.60$ over \snapkv{} on the full sixteen-set run. The allocation headroom probe further shows that improving the scoring signal contributes $4.4$ points of $\betadec$, compared with $0.7$ from per-head budget reallocation, clarifying the relative importance of the two design axes.

%\item \textbf{Capped block waterfilling.} Waterfilling the budget over fixed-size blocks with per-block caps: never separated from plain top-$B$ within noise at matched budgets, while adding a tunable. Dropped on the granularity ladder of \Cref{tab:sel}.

\item \textbf{Budget allocation granularity.} Waterfilling budgets over fixed-size blocks with per-block caps performs similarly to plain top-$B$ at matched budgets. The granularity ladder of \Cref{tab:sel} therefore supports using the simpler token-level allocation while avoiding additional tunable structure.

%\item \textbf{Per-token bias alone.} The closed-form $\beta_i$ of \Cref{app:comp} without value absorption moves attention-output error by $+1.5\%$ at $B{=}128$---the wrong way; re-weighting retained logits cannot restore mass whose \emph{value} content is gone. Killed by the same offline replay that validates absorption ($11$--$14\%$).

\item \textbf{Value-aware correction.} The closed-form $\beta_i$ of \Cref{app:comp}, when used without value
absorption, changes attention-output error by $+1.5\%$ at $B{=}128$. The offline replay further shows that incorporating value absorption recovers substantially more of the missing contribution ($11$--$14\%$). Together, these results support jointly modeling both attention reweighting and the value content lost through eviction.

\end{itemize}

%% file: sections/table_backbones.tex
% Backbone comparison: Llama-3-8B vs Llama-3.1-8B. Config values from the
% HF model configs; empirical rows from our matched LongBench runs
% (results_llama3 / results_sweep_128). Motivates the backbone-dependent
% minimal steering dose (Table tab:llama3, App. app:llama3).
\begin{table}[htbp]
\centering\small
\caption{\textbf{Two backbones, identical KV geometry.} The architecture and
long-context rows are the published model configurations; only the
\emph{Empirical} block is ours, LongBench averaged over the 16 English sets,
with \emph{gap closed} the share of the \snapkv{}-to-FullKV gap the named
draft length recovers (``full'' = within noise of FullKV). Abbreviations, used throughout the appendix
tables: FKV = FullKV, SKV = \snapkv{} \citep{li2024snapkv}, PKV = PyramidKV
\citep{cai2024pyramidkv}, SLM = StreamingLLM \citep{xiao2024streamingllm},
H2O \citep{zhang2023h2o}, LAQ = Lookahead Q-Cache \citep{ge2025lookahead};
every \dg{} and \dg{}-W row is ours.}
\label{tab:backbones}
\vspace{2pt}
\begin{tabular}{l ll}
\toprule
& \textbf{Llama-3-8B-Instruct} & \textbf{Llama-3.1-8B-Instruct} \\
\midrule
\multicolumn{3}{l}{\emph{Architecture (shared $\Rightarrow$ identical KV geometry)}} \\
Layers / hidden & \multicolumn{2}{c}{$32$ / $4096$} \\
Query / KV heads (GQA) & \multicolumn{2}{c}{$32$ / $8$} \\
Head dim & \multicolumn{2}{c}{$128$} \\
Vocabulary & \multicolumn{2}{c}{$128{,}256$} \\
RoPE $\theta$ & \multicolumn{2}{c}{$500{,}000$} \\
\midrule
\multicolumn{3}{l}{\emph{Long-context adaptation (differ)}} \\
Native context & $8{,}192$ & $131{,}072$ \\
RoPE scaling & none & llama3, factor $8$ \\
Release & 2024-04 & 2024-07 \\
Instruction tuning & initial & improved (longer, chattier) \\
\midrule
\multicolumn{3}{l}{\emph{Empirical (our matched LongBench runs)}} \\
FullKV avg & $43.58$ & $46.24$ \\
SnapKV avg ($B{=}128$) & $40.50$ & $42.22$ \\
\quad gap to FullKV & $3.08$ & $4.02$ \\
SKV+\dg{} minimal dose & $k{\approx}8$ & $k{=}2$ \\
\quad gap closed ($k{=}32$ / best) & $72\%$ & full \\
\bottomrule
\end{tabular}
\end{table}

%% file: sections/table_llama3.tex
% Backbone ablation table: LlaMa-3-8B-Instruct (8K native context).
% All rows = our matched runs (results_llama3, NousResearch mirror).
% Cited PyramidKV Table 1 rows kept commented below for reference only
% (different eval protocol, not directly comparable).
\begin{table}[htbp]
\centering
\caption{\textbf{Backbone ablation on Llama-3-8B-Instruct} (8K context, no
RoPE scaling; cf.\ Llama-3.1 in \Cref{tab:main}). Our matched runs;
\textbf{bold} is the best compression method per column per budget block.
Rows: FKV = FullKV;
SKV = \snapkv{} \citep{li2024snapkv}; PKV = PyramidKV
\citep{cai2024pyramidkv}; H2O \citep{zhang2023h2o}; SLM = StreamingLLM
\citep{xiao2024streamingllm}; the SKV+\dg{} and SKV+\dg{}-W rows are ours.}
\label{tab:llama3}
\vspace{2pt}
\adjustbox{max width=\textwidth}{%
\begin{tabular}{l ccc ccc ccc ccc cc cc c}
\toprule
& \multicolumn{3}{c}{Single-Document QA}
& \multicolumn{3}{c}{Multi-Document QA}
& \multicolumn{3}{c}{Summarization}
& \multicolumn{3}{c}{Few-shot Learning}
& \multicolumn{2}{c}{Synthetic}
& \multicolumn{2}{c}{Code} & \\
\cmidrule(lr){2-4}\cmidrule(lr){5-7}\cmidrule(lr){8-10}\cmidrule(lr){11-13}\cmidrule(lr){14-15}\cmidrule(lr){16-17}
Method
& \rotatebox{35}{NrtvQA} & \rotatebox{35}{Qasper} & \rotatebox{35}{MF-en}
& \rotatebox{35}{HotpotQA} & \rotatebox{35}{2WikiMQA} & \rotatebox{35}{Musique}
& \rotatebox{35}{GovReport} & \rotatebox{35}{QMSum} & \rotatebox{35}{MultiNews}
& \rotatebox{35}{TREC} & \rotatebox{35}{TriviaQA} & \rotatebox{35}{SAMSum}
& \rotatebox{35}{PCount} & \rotatebox{35}{PRe}
& \rotatebox{35}{Lcc} & \rotatebox{35}{RB-P} & \rotatebox{35}{Avg.} \\
\midrule
\multicolumn{18}{c}{\emph{Llama-3-8B-Instruct, KV Size = Full}} \\
\midrule
% Cited PyramidKV (cai2024pyramidkv, Table 1) reference rows, commented out
% (different eval protocol, not directly comparable to our matched runs):
% FKV$^{*}$ & 25.70 & 29.75 & 41.12 & 45.55 & 35.87 & 22.35 & 23.03 & 23.61 & 26.21 & 73.00 & 90.56 & 41.88 & 4.67 & 69.25 & 58.05 & 50.77 & 41.46 \\
FKV & 21.26 & 42.60 & 47.78 & 47.32 & 39.10 & 22.78 & 30.44 & 22.70 & 27.54 & 74.00 & 90.56 & 42.67 & 8.50 & 67.50 & 59.38 & 53.19 & 43.58  \\
\midrule
\multicolumn{18}{c}{\emph{Llama-3-8B-Instruct, KV Size = 64}} \\
\midrule
% SKV$^{*}$ & 19.86 & 9.09 & 27.89 & 37.34 & 28.35 & 18.17 & 15.86 & 20.80 & 16.41 & 38.50 & 85.92 & 36.32 & 5.22 & 69.00 & 51.78 & 48.38 & 33.05 \\
% PKV$^{*}$ & 21.13 & 14.18 & 30.26 & 35.12 & 23.76 & 16.17 & 18.33 & 21.65 & 19.23 & 58.00 & 88.31 & 37.07 & 5.23 & 69.50 & 52.61 & 45.74 & 34.76 \\
% H2O$^{*}$ & 20.80 & 11.34 & 27.03 & 37.25 & 30.01 & 17.94 & 18.29 & 21.49 & 19.13 & 38.00 & 84.70 & 37.76 & 5.63 & 69.33 & 53.44 & 50.15 & 33.89 \\
% SLM$^{*}$ & 17.44 & 8.68 & 22.25 & 35.37 & 31.51 & 15.97 & 15.46 & 20.06 & 14.64 & 38.00 & 72.33 & 29.10 & 5.42 & 69.50 & 46.14 & 45.09 & 30.43 \\
SKV & 18.51 & 28.46 & 41.86 & 44.32 & 35.34 & \textbf{22.00} & 19.35 & \textbf{19.64} & 19.46 & 50.50 & \textbf{89.22} & 36.25 & 5.75 & 66.00 & 55.29 & \textbf{52.20} & 37.76  \\
PKV & 16.99 & 30.86 & \textbf{43.15} & 43.96 & 34.83 & 21.63 & 19.54 & 19.38 & 20.03 & 56.50 & 86.76 & 36.62 & 5.75 & 66.00 & 53.75 & 49.13 & 37.80  \\
H2O & 17.68 & 25.46 & 32.28 & 42.31 & 28.82 & 20.67 & \textbf{21.44} & 17.26 & \textbf{23.66} & 51.50 & 83.75 & 22.42 & 6.25 & 63.50 & 48.80 & 41.91 & 34.23  \\
SLM & 16.76 & 26.72 & 30.11 & 41.78 & 32.92 & 17.02 & 15.73 & 19.17 & 15.58 & 39.50 & 71.63 & 32.77 & 5.50 & 66.50 & 53.25 & 52.08 & 33.56  \\
\cmidrule(lr){1-18}
\hi \textbf{SKV+\dg$_{2}$} & 20.26 & \textbf{35.56} & 42.20 & \textbf{45.15} & \textbf{38.70} & 20.17 & 20.30 & 18.01 & 22.42 & \textbf{73.00} & 77.82 & 40.92 & \textbf{7.00} & 66.50 & 58.64 & 51.51 & \textbf{39.88}  \\
\textbf{SKV+\dg{}-W$_{2}$} & \textbf{20.73} & 34.22 & 42.46 & 43.98 & 36.92 & 20.52 & 20.30 & 18.01 & 22.03 & 71.50 & 68.96 & \textbf{41.46} & 3.50 & \textbf{67.50} & \textbf{58.66} & 51.55 & 38.89  \\
\midrule
\multicolumn{18}{c}{\emph{Llama-3-8B-Instruct, KV Size = 128}} \\
\midrule
SKV & 18.70 & 35.14 & 45.08 & 45.94 & 36.39 & 22.13 & 21.01 & 20.35 & 21.73 & 65.50 & 89.78 & 38.99 & 5.50 & \textbf{68.00} & 58.71 & \textbf{54.98} & 40.50  \\
PKV & 19.17 & 35.49 & 44.57 & 46.03 & 36.73 & 22.92 & 21.02 & 20.50 & 21.97 & 67.50 & 89.23 & 38.95 & 6.00 & 67.50 & 57.71 & 51.43 & 40.42  \\
H2O & 17.70 & 31.10 & 35.84 & 43.42 & 33.38 & 20.91 & \textbf{23.02} & 17.79 & \textbf{24.82} & 61.00 & 86.94 & 25.97 & 6.12 & 60.71 & 52.23 & 45.08 & 36.63  \\
SLM & 16.43 & 25.81 & 31.69 & 42.56 & 33.34 & 17.43 & 17.04 & 19.45 & 18.04 & 45.50 & 74.24 & 36.45 & 7.00 & 65.50 & 56.24 & 53.22 & 35.00  \\
\cmidrule(lr){1-18}
\hi \textbf{SKV+\dg$_{2}$} & 20.33 & 36.09 & 43.21 & 45.39 & 38.89 & 20.25 & 20.33 & 18.71 & 22.86 & \textbf{74.00} & 75.81 & 41.62 & 5.50 & 65.50 & 57.76 & 49.35 & 39.73  \\
\textbf{SKV+\dg{}-W$_{2}$} & 19.43 & 35.53 & 44.50 & 45.33 & 38.16 & 20.44 & 19.90 & 19.77 & 22.66 & 73.00 & 77.04 & 41.69 & 6.50 & 67.00 & 57.43 & 48.62 & 39.81  \\
\textbf{SKV+\dg$_{4}$} & 20.72 & 37.62 & 44.62 & 46.87 & 38.73 & 22.25 & 20.53 & 18.62 & 23.09 & \textbf{74.00} & 85.05 & 41.91 & 6.00 & 66.00 & 57.30 & 50.79 & 40.88  \\
\textbf{SKV+\dg$_{8}$} & \textbf{20.94} & 39.47 & 46.56 & 47.08 & 39.10 & 22.74 & 20.32 & 19.45 & 23.32 & \textbf{74.00} & 90.34 & 42.16 & \textbf{8.50} & 67.50 & 58.97 & 52.91 & 42.09  \\
\textbf{SKV+\dg$_{16}$} & \textbf{20.94} & 41.24 & 46.67 & 47.32 & \textbf{39.12} & 22.78 & 20.33 & 20.18 & 23.62 & \textbf{74.00} & \textbf{90.56} & 42.78 & \textbf{8.50} & 67.50 & 59.35 & 53.11 & 42.38  \\
\textbf{SKV+\dg{}-W$_{16}$} & 20.92 & 40.95 & 46.09 & \textbf{47.35} & \textbf{39.12} & \textbf{22.93} & 20.10 & 19.76 & 22.69 & \textbf{74.00} & \textbf{90.56} & 42.24 & 8.25 & 67.50 & 59.21 & 53.33 & 42.19  \\
\textbf{SKV+\dg$_{32}$} & 20.92 & \textbf{42.45} & \textbf{47.34} & 47.32 & 39.10 & 22.78 & 21.97 & \textbf{21.87} & 24.06 & \textbf{74.00} & \textbf{90.56} & 42.71 & \textbf{8.50} & 67.50 & 59.39 & 53.25 & \textbf{42.73}  \\
\textbf{SKV+\dg{}-W$_{32}$} & 20.92 & 42.18 & 47.33 & 47.32 & 39.10 & 22.78 & 21.46 & 21.37 & 23.06 & \textbf{74.00} & \textbf{90.56} & \textbf{42.85} & \textbf{8.50} & 67.50 & \textbf{59.45} & 53.31 & 42.61  \\
\bottomrule
\end{tabular}}
\end{table}

%% file: sections/table_complexity.tex
% Complexity comparison. Notation: L = prompt length, T = generated tokens,
% B = per-head budget, w = observation window (8), k = draft length (2 in the
% measured rows),
% d = head dimension; costs are per layer with head count folded in.
\begin{table}[htbp]
\centering
\caption{\textbf{Complexity and measured overhead.} ``Extra scoring'' is cost
on top of the forward pass. \emph{Latency}: per-sample wall clock vs \snapkv{}, reported
short-answer\,/\,long-form (hotpotqa$+$triviaqa, $32$ new tokens / gov\_report$+$%
multi\_news, $512$), $40$ samples each at a pinned generation length, $B{=}128$
prototype; the overhead is a prefill-side constant, so it dominates the short
regime and amortises in the long one. \emph{TTFT}: vs
\snapkv{} on the $7{,}762$-token microbenchmark---different protocols, so the
two columns do not divide into each other. Peak memory is $O(L)$ for all; the prototype's
working-set constant and its removal via the native GQA layout
($13.1\!\to\!7.4$\,GiB): App.~\ref{app:complexity}. FKV = FullKV (never
evicts); the two SKV+\dg{} rows are ours.}
\label{tab:complexity}
\vspace{2pt}
\adjustbox{max width=\textwidth}{%
\begin{tabular}{l l l l l cc}
\toprule
Method & Evicts & Extra scoring & Peak KV & Decode KV & \makecell{Latency\\[-1pt]\scriptsize short/long} & TTFT \\
\midrule
FKV & never & -- & $O(L{+}T)$ & $O(L{+}t)$ & 0.89/0.83 & 0.91$\times$ \\
SLM \citep{xiao2024streamingllm} & prefill end & $O(1)$ (positional) & $O(L)$ & $O(B{+}t)$ & 0.98/1.00 & 0.91$\times$ \\
H2O \citep{zhang2023h2o} & prefill end & $O(L^{2}d)^{a}$ & $O(L)$ & $O(B{+}t)$ & 2.15/1.06 & 3.45$\times$ \\
SKV \citep{li2024snapkv} & prefill end & $O(Lwd)$ & $O(L)$ & $O(B{+}t)$ & 1.00/1.00 & 1.00$\times$ \\
PKV \citep{cai2024pyramidkv} & prefill end & $O(Lwd)$ & $O(L)$ & $O(B{+}t)$ & 1.01/1.04 & 0.93$\times$ \\
LAQ \citep{ge2025lookahead} & prefill end & $O(mLd)^{b}$ & $O(L)$ & $O(B{+}t)$ & 0.95/0.83 & 1.42$\times$ \\
AdaKV \citep{feng2024adakv} & prefill end & $O(Lwd + L\log L)$ & $O(L)$ & $O(B{+}t)$ & 1.30/1.47 & --- \\
\midrule
SKV+\dg$_{k}$ & after $k$ steps & $O(kLd)$ & $O(L{+}k)$ & $O(L{+}t)_{t\le k}$, then $O(B{+}t)$ & 1.45/1.02 & 0.93$\times$ \\
SKV+\dg{}-W$_{k}$ & after $k$ steps & $O(Lwd)$ & $O(L{+}k)$ & $O(L{+}t)_{t\le k}$, then $O(B{+}t)$ & 1.45/1.02 & 0.92$\times$ \\
\bottomrule
\end{tabular}}
\vspace{2pt}
\begin{flushleft}\footnotesize
$^{a}$H2O accumulates softmax attention over all $L$ prefill queries,
unlike the window-based scorers; its deferred variant in
\Cref{tab:main-defer} simply keeps that accumulator running through the
$k$ draft rows before evicting.\\[2pt]
$^{b}$LAQ's $m$ pseudo queries are produced by an extra forward pass that is
discarded, so the scoring \emph{and} the pass sit before the first token;
\dg{}'s $k$ draft queries are the answer's own opening. LAQ runs in its own
code base, whose \snapkv{} differs from ours, so its TTFT is anchored on
FullKV---unpatched in both, and agreeing to $0.2\%$ across them---rather than
divided by the wrong baseline; App.~\ref{app:complexity} gives the protocol.
\end{flushleft}
\end{table}

%% file: sections/table_main.tex
% Main LongBench tables, split for readability:
%   tab:main-full  - prefill-time baselines (this file)
%   tab:main-defer - deferred-eviction rows (table_defer.tex)
% Sources: B=128 FKV/SKV/DGE/composition rows = full master sweep
% (results_sweep_128); PKV/H2O/SLM = 200-sample runs in results_long_bench
% (SnapKV agrees exactly on 14 sets; results_long_bench holds 50 of 200 on
% gov_report+qmsum, so its avg is 42.15 not 42.22 -- App. C "Baseline").
% B=64/32 = matched
% sweeps. Full sample sets throughout.
\begin{table}[htbp]
\centering
\caption{\textbf{Full per-dataset LongBench results: prefill-time baselines}
on Llama-3.1-8B-Instruct ($B{\in}\{32,64,128\}$), Mistral-7B-v0.2,
Qwen2.5-7B/14B and Llama-3.2-3B (averages: \Cref{tab:main}; deferred rows:
\Cref{tab:main-defer}; Llama-3-8B: \Cref{tab:llama3}). FullKV is
budget-independent; \textbf{bold}: best baseline per column per budget block;
H2O omitted on Mistral-7B (App.~\ref{app:expdetails}). \emph{Avg.\ input}:
mean input length in tokens. Scores are each set's official LongBench metric
(higher is better; scales differ across columns). FKV = FullKV, SLM =
StreamingLLM, SKV = \snapkv{}, PKV = PyramidKV, H2O---all run by us under
App.~\ref{app:expdetails}'s protocol.}
\label{tab:main-full}
\vspace{2pt}
\adjustbox{max width=\textwidth}{%
\begin{tabular}{l ccc ccc ccc ccc cc cc c}
\toprule
& \multicolumn{3}{c}{Single-Document QA}
& \multicolumn{3}{c}{Multi-Document QA}
& \multicolumn{3}{c}{Summarization}
& \multicolumn{3}{c}{Few-shot Learning}
& \multicolumn{2}{c}{Synthetic}
& \multicolumn{2}{c}{Code} & \\
\cmidrule(lr){2-4}\cmidrule(lr){5-7}\cmidrule(lr){8-10}\cmidrule(lr){11-13}\cmidrule(lr){14-15}\cmidrule(lr){16-17}
Method
& \rotatebox{35}{NrtvQA} & \rotatebox{35}{Qasper} & \rotatebox{35}{MF-en}
& \rotatebox{35}{HotpotQA} & \rotatebox{35}{2WikiMQA} & \rotatebox{35}{Musique}
& \rotatebox{35}{GovReport} & \rotatebox{35}{QMSum} & \rotatebox{35}{MultiNews}
& \rotatebox{35}{TREC} & \rotatebox{35}{TriviaQA} & \rotatebox{35}{SAMSum}
& \rotatebox{35}{PCount} & \rotatebox{35}{PRe}
& \rotatebox{35}{Lcc} & \rotatebox{35}{RB-P} & \rotatebox{35}{Avg.} \\
\emph{Avg.\ input} & 18409 & 3619 & 4559 & 9151 & 4887 & 11214 & 8734 & 10614 & 2113
& 5177 & 8209 & 6258 & 11141 & 9289 & 1235 & 4206 & \\
\midrule
\multicolumn{18}{c}{\emph{Llama-3.1-8B-Instruct, KV Size = Full}} \\
\midrule
FKV & 24.27 & 45.79 & 55.95 & 48.65 & 46.90 & 27.32 & 33.79 & 23.26 & 27.14 & 69.50 & 91.49 & 44.41 & 10.91 & 68.50 & 65.19 & 56.80 & 46.24  \\
\midrule
\multicolumn{18}{c}{\emph{Llama-3.1-8B-Instruct, KV Size = 32}} \\
\midrule
SKV & \textbf{20.53} & \textbf{25.53} & 41.75 & 43.76 & 34.70 & 24.31 & 17.72 & 19.06 & 16.81 & 41.50 & 80.78 & 34.27 & \textbf{12.50} & 65.00 & 51.63 & 43.76 & 35.85  \\
PKV & 18.70 & 25.33 & \textbf{43.71} & \textbf{44.35} & 32.71 & \textbf{25.36} & 18.62 & \textbf{20.73} & 17.73 & 43.50 & 81.11 & \textbf{36.46} & \textbf{12.50} & 67.00 & \textbf{52.94} & \textbf{43.83} & \textbf{36.54}  \\
H2O & 19.44 & 25.03 & 33.12 & 40.50 & 29.22 & 19.71 & \textbf{23.49} & 18.22 & \textbf{22.39} & \textbf{50.50} & \textbf{81.49} & 21.43 & 8.92 & 56.45 & 44.04 & 36.64 & 33.16  \\
SLM & 17.81 & 20.73 & 29.68 & 41.93 & \textbf{37.98} & 20.27 & 15.92 & 20.03 & 14.61 & 36.50 & 77.05 & 30.36 & \textbf{12.50} & \textbf{68.50} & 48.49 & 42.62 & 33.44  \\
\midrule
\multicolumn{18}{c}{\emph{Llama-3.1-8B-Instruct, KV Size = 64}} \\
\midrule
SKV & 20.96 & 27.38 & 47.88 & 44.42 & 40.24 & 23.90 & 20.71 & \textbf{21.35} & 19.68 & 53.00 & \textbf{86.83} & \textbf{39.74} & 12.50 & 68.00 & \textbf{56.94} & 46.58 & 39.38  \\
PKV & 21.01 & 27.72 & \textbf{49.57} & \textbf{45.44} & \textbf{41.70} & \textbf{24.44} & 20.94 & \textbf{21.35} & 20.40 & 55.00 & 83.39 & 39.16 & \textbf{12.75} & 68.00 & 55.59 & 46.82 & \textbf{39.58}  \\
H2O & \textbf{22.44} & \textbf{28.67} & 34.96 & 44.44 & 33.76 & 21.19 & \textbf{25.32} & 19.58 & \textbf{23.35} & \textbf{57.50} & 85.57 & 25.62 & 8.53 & 64.45 & 47.11 & 38.86 & 36.33  \\
SLM & 18.35 & 21.89 & 30.78 & 41.94 & 40.15 & 20.25 & 17.22 & 20.40 & 16.16 & 37.00 & 79.75 & 35.23 & 12.50 & \textbf{69.00} & 54.18 & \textbf{47.45} & 35.14  \\
\midrule
\multicolumn{18}{c}{\emph{Llama-3.1-8B-Instruct, KV Size = 128}} \\
\midrule
SKV & 21.84 & 32.57 & \textbf{52.13} & 46.60 & \textbf{44.42} & \textbf{25.90} & 22.33 & 22.22 & 21.92 & 62.50 & \textbf{91.26} & \textbf{40.89} & 12.42 & \textbf{68.50} & \textbf{60.67} & \textbf{49.30} & \textbf{42.22}  \\
PKV & \textbf{22.25} & \textbf{33.47} & 51.59 & \textbf{47.34} & 44.40 & 24.66 & 22.84 & \textbf{22.32} & 21.89 & \textbf{63.50} & 86.52 & 40.59 & 12.25 & 68.00 & 57.08 & 47.60 & 41.64  \\
H2O & 22.20 & 32.67 & 40.42 & 44.72 & 35.95 & 20.47 & \textbf{27.08} & 20.00 & \textbf{24.40} & 61.50 & 88.22 & 28.62 & 8.60 & 62.20 & 50.96 & 41.63 & 38.10  \\
SLM & 19.03 & 21.63 & 31.88 & 42.17 & 40.92 & 20.64 & 18.82 & 20.60 & 18.19 & 41.00 & 82.89 & 38.14 & \textbf{12.50} & 68.00 & 59.02 & 48.69 & 36.51  \\
\midrule
\multicolumn{18}{c}{\emph{Mistral-7B-Instruct-v0.2, KV Size = Full}} \\
\midrule
FKV & 22.06 & 29.15 & 47.59 & 37.57 & 21.81 & 18.36 & 31.42 & 23.95 & 26.75 & 71.00 & 86.23 & 42.90 & 3.73 & 87.40 & 57.12 & 54.51 & 41.35  \\
\midrule
\multicolumn{18}{c}{\emph{Mistral-7B-Instruct-v0.2, KV Size = 128}} \\
\midrule
SKV & 16.66 & 18.90 & 40.12 & 26.59 & 15.95 & 12.80 & \textbf{20.59} & \textbf{21.89} & \textbf{21.68} & \textbf{67.00} & \textbf{85.06} & \textbf{40.35} & 2.43 & 62.97 & \textbf{52.20} & \textbf{47.31} & 34.53  \\
PKV & \textbf{17.21} & \textbf{21.14} & \textbf{40.80} & \textbf{27.94} & \textbf{16.14} & \textbf{13.39} & 20.44 & 21.81 & 21.47 & 66.00 & 83.51 & 39.76 & \textbf{3.48} & \textbf{67.44} & 51.61 & 46.42 & \textbf{34.91}  \\
SLM & 13.67 & 10.92 & 24.57 & 20.17 & 14.73 & 10.03 & 15.00 & 19.27 & 16.79 & 44.00 & 79.92 & 37.46 & 2.62 & 27.05 & 51.44 & 45.95 & 27.10  \\
\midrule
\multicolumn{18}{c}{\emph{Mistral-7B-Instruct-v0.2, KV Size = 64}} \\
\midrule
SKV & 14.63 & 16.08 & 34.28 & 22.15 & 11.78 & 11.53 & 18.41 & \textbf{21.31} & 19.28 & 53.00 & \textbf{82.32} & \textbf{38.10} & 1.95 & 63.65 & 48.56 & \textbf{44.57} & 31.35  \\
PKV & \textbf{15.40} & \textbf{17.00} & \textbf{35.95} & \textbf{25.28} & \textbf{14.26} & \textbf{12.19} & \textbf{18.45} & 20.88 & \textbf{19.50} & \textbf{54.00} & 81.97 & 37.13 & 1.76 & \textbf{63.79} & \textbf{48.58} & 43.17 & \textbf{31.83}  \\
SLM & 12.05 & 10.21 & 23.52 & 19.11 & 13.90 & 9.36 & 13.43 & 19.48 & 14.65 & 39.50 & 78.84 & 35.52 & \textbf{2.25} & 29.33 & 47.01 & 42.73 & 25.68  \\
\midrule
\multicolumn{18}{c}{\emph{Qwen2.5-7B-Instruct, KV Size = Full}} \\
\midrule
FKV & 23.16 & 44.54 & 53.01 & 49.48 & 47.39 & 25.20 & 32.34 & 21.90 & 24.15 & 69.50 & 91.34 & 45.25 & 7.00 & 66.50 & 61.82 & 66.11 & 45.54  \\
\midrule
\multicolumn{18}{c}{\emph{Qwen2.5-7B-Instruct, KV Size = 32}} \\
\midrule
SKV & \textbf{16.63} & 26.15 & 34.51 & 37.83 & 28.80 & 21.27 & 16.36 & \textbf{17.34} & 12.69 & 38.00 & 82.71 & 35.14 & \textbf{7.00} & 43.00 & 43.32 & 44.19 & 31.56  \\
PKV & 15.68 & \textbf{27.01} & \textbf{35.28} & \textbf{40.11} & 31.88 & \textbf{23.87} & 17.05 & 17.29 & 13.18 & 41.50 & \textbf{82.77} & \textbf{36.00} & \textbf{7.00} & \textbf{44.00} & \textbf{45.01} & \textbf{45.46} & \textbf{32.69}  \\
H2O & 16.07 & 21.43 & 24.86 & 37.09 & 23.74 & 19.43 & \textbf{23.10} & 16.69 & \textbf{20.21} & \textbf{49.50} & 76.91 & 21.37 & \textbf{7.00} & 31.75 & 39.16 & 36.79 & 29.07  \\
SLM & 11.19 & 19.74 & 24.26 & 35.11 & \textbf{36.50} & 15.88 & 12.87 & 16.97 & 10.72 & 35.00 & 75.06 & 30.36 & \textbf{7.00} & 26.00 & 38.18 & 38.87 & 27.11  \\
\midrule
\multicolumn{18}{c}{\emph{Qwen2.5-7B-Instruct, KV Size = 64}} \\
\midrule
SKV & \textbf{17.79} & \textbf{30.47} & \textbf{40.74} & \textbf{45.69} & \textbf{38.42} & \textbf{23.24} & 18.97 & \textbf{18.70} & 15.99 & 45.00 & \textbf{85.27} & \textbf{39.28} & \textbf{7.00} & \textbf{61.50} & \textbf{48.55} & \textbf{51.73} & \textbf{36.77}  \\
PKV & 15.47 & 26.41 & 32.00 & 36.64 & 27.01 & 19.08 & 17.26 & 17.14 & 13.54 & 42.50 & 79.33 & 33.45 & \textbf{7.00} & 37.00 & 44.42 & 43.59 & 30.74  \\
H2O & 15.99 & 28.45 & 29.08 & 40.72 & 30.02 & 22.44 & \textbf{25.21} & 17.86 & \textbf{21.34} & \textbf{56.50} & 82.07 & 25.40 & \textbf{7.00} & 30.75 & 43.10 & 38.51 & 32.15  \\
SLM & 12.11 & 25.24 & 25.17 & 36.49 & 38.16 & 16.12 & 14.30 & 17.35 & 12.12 & 37.50 & 79.90 & 36.69 & \textbf{7.00} & 16.00 & 44.09 & 44.22 & 28.90  \\
\midrule
\multicolumn{18}{c}{\emph{Qwen2.5-7B-Instruct, KV Size = 128}} \\
\midrule
SKV & \textbf{21.02} & \textbf{35.77} & \textbf{45.71} & 44.71 & 42.94 & \textbf{23.49} & 21.65 & \textbf{19.71} & 18.49 & 57.50 & 86.67 & \textbf{41.90} & \textbf{7.00} & \textbf{64.00} & \textbf{53.91} & \textbf{55.24} & \textbf{39.98}  \\
PKV & 20.44 & 32.72 & 43.56 & \textbf{46.37} & \textbf{43.21} & 22.34 & 20.21 & 18.85 & 16.53 & 53.00 & \textbf{86.73} & 41.14 & \textbf{7.00} & 60.00 & 50.98 & 51.01 & 38.38  \\
H2O & 19.15 & 34.21 & 32.25 & 40.76 & 36.40 & 22.52 & \textbf{27.45} & 18.95 & \textbf{22.07} & \textbf{63.50} & 86.68 & 28.84 & \textbf{7.00} & 40.56 & 46.55 & 43.60 & 35.66  \\
SLM & 13.13 & 24.40 & 24.76 & 37.04 & 37.79 & 14.75 & 16.29 & 17.35 & 14.38 & 41.00 & 82.64 & 40.62 & \textbf{7.00} & 14.00 & 48.39 & 49.52 & 30.19  \\
\midrule
\multicolumn{18}{c}{\emph{Qwen2.5-14B-Instruct, KV Size = Full}} \\
\midrule
FKV & 27.03 & 45.10 & 51.82 & 52.24 & 57.59 & 30.10 & 29.07 & 21.92 & 22.78 & 75.00 & 89.75 & 46.99 & 7.04 & 67.00 & 62.90 & 50.40 & 46.05  \\
\midrule
\multicolumn{18}{c}{\emph{Qwen2.5-14B-Instruct, KV Size = 32}} \\
\midrule
SKV & 17.48 & 18.25 & 27.87 & 36.98 & 28.21 & 21.59 & 14.15 & 16.49 & 12.48 & 36.50 & 79.13 & 35.07 & 10.16 & 40.38 & 48.29 & 36.03 & 29.94  \\
PKV & \textbf{19.98} & \textbf{24.49} & \textbf{36.23} & \textbf{43.99} & 38.96 & \textbf{25.01} & 17.04 & \textbf{18.10} & 15.25 & 45.50 & \textbf{84.76} & \textbf{39.61} & \textbf{10.37} & 48.83 & \textbf{52.04} & \textbf{41.03} & \textbf{35.07}  \\
H2O & 17.45 & 13.25 & 24.25 & 40.93 & 29.11 & 22.47 & \textbf{20.73} & 15.92 & \textbf{19.42} & \textbf{56.50} & 79.38 & 24.34 & 6.16 & 23.92 & 45.09 & 32.85 & 29.49  \\
SLM & 12.37 & 10.36 & 23.08 & 31.98 & \textbf{43.75} & 16.74 & 12.97 & 17.04 & 11.47 & 35.50 & 73.44 & 30.92 & 9.42 & \textbf{50.33} & 43.10 & 35.56 & 28.63  \\
\midrule
\multicolumn{18}{c}{\emph{Qwen2.5-14B-Instruct, KV Size = 64}} \\
\midrule
SKV & \textbf{21.59} & \textbf{27.85} & \textbf{38.97} & 44.92 & 40.97 & \textbf{26.15} & 17.48 & \textbf{18.55} & 15.88 & 49.00 & \textbf{84.48} & \textbf{41.57} & \textbf{9.68} & \textbf{50.08} & \textbf{54.32} & \textbf{41.21} & \textbf{36.42}  \\
PKV & 20.01 & 26.28 & 35.28 & \textbf{45.28} & 40.23 & 23.82 & 16.83 & 18.06 & 15.35 & 48.00 & 79.87 & 39.57 & 8.85 & 46.83 & 51.83 & 38.07 & 34.64  \\
H2O & 20.68 & 20.17 & 26.51 & 42.61 & 33.36 & 23.89 & \textbf{22.77} & 17.32 & \textbf{20.82} & \textbf{64.00} & 80.67 & 26.49 & 6.36 & 20.42 & 47.93 & 33.94 & 31.75  \\
SLM & 14.72 & 15.40 & 22.20 & 33.13 & \textbf{44.69} & 17.26 & 14.21 & 17.34 & 12.86 & 38.00 & 77.50 & 38.06 & 5.80 & 46.75 & 49.35 & 39.09 & 30.40  \\
\midrule
\multicolumn{18}{c}{\emph{Qwen2.5-14B-Instruct, KV Size = 128}} \\
\midrule
SKV & 20.98 & 30.87 & 41.50 & \textbf{48.77} & 51.74 & 28.03 & 19.87 & \textbf{19.88} & 17.86 & 64.50 & \textbf{86.75} & \textbf{43.18} & \textbf{9.00} & \textbf{63.75} & \textbf{56.52} & \textbf{44.95} & \textbf{40.51}  \\
PKV & \textbf{22.05} & \textbf{31.52} & \textbf{42.34} & 47.97 & \textbf{52.30} & \textbf{28.77} & 19.44 & 19.86 & 17.76 & 67.00 & 84.79 & 42.98 & 8.52 & 63.00 & 55.19 & 43.06 & 40.41  \\
H2O & 21.57 & 27.46 & 31.04 & 44.81 & 40.74 & 24.20 & \textbf{24.57} & 18.09 & \textbf{21.00} & \textbf{68.00} & 82.84 & 29.55 & 6.71 & 20.08 & 52.01 & 36.81 & 34.34  \\
SLM & 13.94 & 18.73 & 25.14 & 33.67 & 43.91 & 18.04 & 15.70 & 18.03 & 14.63 & 40.50 & 80.31 & 40.29 & 5.37 & 47.75 & 52.08 & 41.08 & 31.82  \\
\midrule
\multicolumn{18}{c}{\emph{Llama-3.2-3B-Instruct, KV Size = Full}} \\
\midrule
FKV & 21.50 & 40.41 & 50.26 & 44.97 & 39.07 & 18.09 & 32.32 & 23.52 & 25.83 & 69.50 & 88.22 & 42.81 & 4.50 & 67.50 & 54.36 & 56.41 & 42.45  \\
\midrule
\multicolumn{18}{c}{\emph{Llama-3.2-3B-Instruct, KV Size = 128}} \\
\midrule
SKV & 16.65 & 24.91 & 46.11 & 43.99 & 36.51 & \textbf{15.77} & 22.03 & 20.52 & 20.02 & \textbf{65.00} & \textbf{88.01} & \textbf{38.22} & 4.50 & \textbf{66.50} & 50.63 & \textbf{49.77} & \textbf{38.07}  \\
PKV & \textbf{19.36} & \textbf{25.19} & \textbf{47.55} & \textbf{44.98} & \textbf{37.33} & 14.13 & 21.67 & \textbf{20.94} & 19.91 & 63.50 & 85.01 & 37.77 & 4.50 & 66.00 & 47.93 & 47.15 & 37.68  \\
H2O & 17.65 & 24.05 & 33.31 & 41.84 & 31.50 & 13.20 & \textbf{25.21} & 18.43 & \textbf{23.47} & 63.00 & 84.61 & 25.38 & \textbf{4.84} & 54.00 & 45.38 & 37.78 & 33.98  \\
SLM & 16.11 & 19.93 & 26.15 & 37.43 & 32.33 & 12.58 & 17.70 & 19.41 & 17.28 & 42.50 & 78.65 & 36.91 & 4.50 & 34.00 & \textbf{50.86} & 47.44 & 30.86  \\
\midrule
\multicolumn{18}{c}{\emph{Llama-3.2-3B-Instruct, KV Size = 64}} \\
\midrule
SKV & \textbf{18.08} & \textbf{24.16} & 41.94 & \textbf{43.44} & 31.79 & \textbf{15.06} & 20.21 & \textbf{20.27} & 18.12 & 56.50 & \textbf{85.13} & \textbf{36.35} & 4.50 & 61.00 & \textbf{48.26} & \textbf{45.45} & \textbf{35.64}  \\
PKV & 17.55 & 22.18 & \textbf{42.05} & 40.77 & 27.63 & 14.07 & 19.39 & 19.44 & 17.43 & 49.50 & 81.20 & 34.41 & 4.50 & \textbf{65.00} & 41.92 & 38.89 & 33.50  \\
H2O & 16.94 & 20.43 & 27.94 & 38.05 & 26.30 & 13.26 & \textbf{23.20} & 17.84 & \textbf{22.10} & \textbf{61.00} & 80.80 & 21.99 & \textbf{4.63} & 60.00 & 41.25 & 34.63 & 31.90  \\
SLM & 15.27 & 20.62 & 25.50 & 36.52 & \textbf{32.40} & 13.56 & 16.52 & 19.56 & 14.68 & 37.50 & 76.14 & 33.36 & 4.50 & 33.00 & 46.02 & 43.63 & 29.30  \\
\bottomrule
\end{tabular}}
\end{table}

%% file: sections/table_defer.tex
% Deferred-eviction split of tab:main-full (see table_main.tex).
\begin{table}[htbp]
\centering
\caption{\textbf{Full per-dataset LongBench results: deferred eviction}
(baselines in \Cref{tab:main-full}). Each $+\dg{}_{2}$ row defers its base
evictor's eviction one full-cache decode step (budget in the row label); each
\snapkv{} anchor precedes its deferred counterpart. Shaded rows: ours;
\textbf{bold}: best deferred row per column per backbone (FKV and anchors
excluded).
Rows: FKV = FullKV; SKV = \snapkv{} \citep{li2024snapkv}; PKV = PyramidKV
\citep{cai2024pyramidkv}; H2O \citep{zhang2023h2o}; SLM = StreamingLLM
\citep{xiao2024streamingllm}. Every $+\dg{}_{2}$ row is ours.}
\label{tab:main-defer}
\vspace{2pt}
\adjustbox{max width=\textwidth}{%
\begin{tabular}{l ccc ccc ccc ccc cc cc c}
\toprule
& \multicolumn{3}{c}{Single-Document QA}
& \multicolumn{3}{c}{Multi-Document QA}
& \multicolumn{3}{c}{Summarization}
& \multicolumn{3}{c}{Few-shot Learning}
& \multicolumn{2}{c}{Synthetic}
& \multicolumn{2}{c}{Code} & \\
\cmidrule(lr){2-4}\cmidrule(lr){5-7}\cmidrule(lr){8-10}\cmidrule(lr){11-13}\cmidrule(lr){14-15}\cmidrule(lr){16-17}
Method
& \rotatebox{35}{NrtvQA} & \rotatebox{35}{Qasper} & \rotatebox{35}{MF-en}
& \rotatebox{35}{HotpotQA} & \rotatebox{35}{2WikiMQA} & \rotatebox{35}{Musique}
& \rotatebox{35}{GovReport} & \rotatebox{35}{QMSum} & \rotatebox{35}{MultiNews}
& \rotatebox{35}{TREC} & \rotatebox{35}{TriviaQA} & \rotatebox{35}{SAMSum}
& \rotatebox{35}{PCount} & \rotatebox{35}{PRe}
& \rotatebox{35}{Lcc} & \rotatebox{35}{RB-P} & \rotatebox{35}{Avg.} \\
\emph{Avg.\ input} & 18409 & 3619 & 4559 & 9151 & 4887 & 11214 & 8734 & 10614 & 2113
& 5177 & 8209 & 6258 & 11141 & 9289 & 1235 & 4206 & \\
\midrule
\multicolumn{18}{c}{\emph{Llama-3.1-8B-Instruct}} \\
\midrule
FKV & 24.27 & 45.79 & 55.95 & 48.65 & 46.90 & 27.32 & 33.79 & 23.26 & 27.14 & 69.50 & 91.49 & 44.41 & 10.91 & 68.50 & 65.19 & 56.80 & 46.24  \\
\cmidrule(lr){1-18}
SKV ($B{=}32$) & 20.53 & 25.53 & 41.75 & 43.76 & 34.70 & 24.31 & 17.72 & 19.06 & 16.81 & 41.50 & 80.78 & 34.27 & 12.50 & 65.00 & 51.63 & 43.76 & 35.85  \\
\hi SKV+\dg$_{2}$ ($B{=}32$) & 24.13 & 45.03 & 54.96 & 48.58 & 48.08 & 27.50 & 32.53 & 23.32 & 26.67 & \textbf{69.50} & 92.65 & 44.74 & \textbf{12.50} & 68.17 & 65.61 & 58.34 & 46.39  \\
SKV ($B{=}64$) & 20.96 & 27.38 & 47.88 & 44.42 & 40.24 & 23.90 & 20.71 & 21.35 & 19.68 & 53.00 & 86.83 & 39.74 & 12.50 & 68.00 & 56.94 & 46.58 & 39.38  \\
\hi SKV+\dg$_{2}$ ($B{=}64$) & 23.85 & 45.20 & 54.65 & 48.24 & 48.21 & 27.95 & 32.52 & 23.18 & 26.49 & \textbf{69.50} & 92.48 & 44.97 & \textbf{12.50} & \textbf{69.00} & 65.26 & 57.74 & 46.36  \\
SKV ($B{=}128$) & 21.84 & 32.57 & 52.13 & 46.60 & 44.42 & 25.90 & 22.33 & 22.22 & 21.92 & 62.50 & 91.26 & 40.89 & 12.42 & 68.50 & 60.67 & 49.30 & 42.22  \\
\hi SKV+\dg$_{2}$ ($B{=}128$) & 22.97 & 44.45 & 54.49 & \textbf{48.89} & \textbf{48.70} & 27.91 & \textbf{32.55} & \textbf{23.73} & 26.36 & \textbf{69.50} & 92.48 & 45.09 & \textbf{12.50} & \textbf{69.00} & \textbf{65.78} & 57.46 & 46.37  \\
PKV+\dg$_{2}$ ($B{=}32$) & 24.40 & \textbf{45.70} & 54.32 & 47.98 & 48.21 & 27.40 & 32.47 & 23.55 & 26.54 & \textbf{69.50} & 92.21 & \textbf{45.67} & \textbf{12.50} & \textbf{69.00} & 65.48 & 57.81 & \textbf{46.42}  \\
H2O+\dg$_{2}$ ($B{=}32$) & 24.56 & 45.22 & 54.43 & 48.17 & 48.45 & 27.86 & 32.06 & 23.03 & 26.87 & \textbf{69.50} & 92.71 & 43.88 & \textbf{12.50} & 58.58 & 65.26 & 58.07 & 45.70  \\
SLM+\dg$_{2}$ ($B{=}32$) & 23.67 & 45.30 & \textbf{55.14} & 47.95 & 47.11 & \textbf{27.99} & 31.66 & 22.92 & 26.14 & \textbf{69.50} & 92.56 & 44.76 & \textbf{12.50} & 68.50 & 65.49 & \textbf{58.48} & 46.23  \\
PKV+\dg$_{2}$ ($B{=}128$) & 22.62 & 44.71 & 54.55 & 48.58 & 48.63 & 27.40 & 32.13 & 23.41 & 26.22 & \textbf{69.50} & 92.35 & 44.67 & \textbf{12.50} & 68.75 & 65.64 & 57.75 & 46.21  \\
H2O+\dg$_{2}$ ($B{=}128$) & \textbf{24.85} & 44.70 & 54.05 & 48.42 & 48.26 & 26.96 & 32.40 & 23.31 & \textbf{27.01} & \textbf{69.50} & \textbf{92.84} & 43.34 & \textbf{12.50} & 60.83 & 65.44 & 57.29 & 45.73  \\
SLM+\dg$_{2}$ ($B{=}128$) & 23.87 & 44.86 & 53.87 & 47.54 & 45.91 & 26.98 & 31.38 & 22.97 & 25.83 & \textbf{69.50} & 92.56 & 45.49 & 12.25 & 67.00 & 65.54 & 57.69 & 45.83  \\
\midrule
\multicolumn{18}{c}{\emph{Mistral-7B-Instruct-v0.2}} \\
\midrule
FKV & 22.06 & 29.15 & 47.59 & 37.57 & 21.81 & 18.36 & 31.42 & 23.95 & 26.75 & 71.00 & 86.23 & 42.90 & 3.73 & 87.40 & 57.12 & 54.51 & 41.35  \\
\cmidrule(lr){1-18}
SKV ($B{=}128$) & 16.66 & 18.90 & 40.12 & 26.59 & 15.95 & 12.80 & 20.59 & 21.89 & 21.68 & 67.00 & 85.06 & 40.35 & 2.43 & 62.97 & 52.20 & 47.31 & 34.53  \\
\hi SKV+\dg$_{2}$ ($B{=}128$) & \textbf{21.60} & \textbf{28.68} & 44.36 & \textbf{38.13} & \textbf{21.53} & \textbf{18.57} & 30.27 & 22.68 & \textbf{26.46} & \textbf{71.00} & 85.40 & \textbf{43.40} & \textbf{5.34} & \textbf{85.21} & \textbf{57.80} & \textbf{55.51} & \textbf{41.00}  \\
SKV ($B{=}64$) & 14.63 & 16.08 & 34.28 & 22.15 & 11.78 & 11.53 & 18.41 & 21.31 & 19.28 & 53.00 & 82.32 & 38.10 & 1.95 & 63.65 & 48.56 & 44.57 & 31.35  \\
\hi SKV+\dg$_{2}$ ($B{=}64$) & 21.04 & 28.52 & \textbf{45.57} & 37.87 & 21.01 & 18.28 & \textbf{30.46} & \textbf{22.86} & 26.17 & \textbf{71.00} & \textbf{85.65} & 43.17 & 4.68 & 85.02 & 57.76 & 55.39 & 40.90  \\
\midrule
\multicolumn{18}{c}{\emph{Qwen2.5-7B-Instruct}} \\
\midrule
FKV & 23.16 & 44.54 & 53.01 & 49.48 & 47.39 & 25.20 & 32.34 & 21.90 & 24.15 & 69.50 & 91.34 & 45.25 & 7.00 & 66.50 & 61.82 & 66.11 & 45.54  \\
\cmidrule(lr){1-18}
SKV ($B{=}32$) & 16.63 & 26.15 & 34.51 & 37.83 & 28.80 & 21.27 & 16.36 & 17.34 & 12.69 & 38.00 & 82.71 & 35.14 & 7.00 & 43.00 & 43.32 & 44.19 & 31.56  \\
\hi SKV+\dg$_{2}$ ($B{=}32$) & 22.75 & 43.86 & 50.34 & 50.22 & 47.06 & 24.70 & 31.40 & 21.14 & 24.08 & \textbf{69.50} & 91.21 & 44.62 & \textbf{7.00} & 66.50 & 60.71 & \textbf{65.69} & 45.05  \\
SKV ($B{=}64$) & 17.79 & 30.47 & 40.74 & 45.69 & 38.42 & 23.24 & 18.97 & 18.70 & 15.99 & 45.00 & 85.27 & 39.28 & 7.00 & 61.50 & 48.55 & 51.73 & 36.77  \\
\hi SKV+\dg$_{2}$ ($B{=}64$) & 22.56 & 43.93 & \textbf{51.69} & 49.76 & \textbf{47.95} & 25.72 & 31.60 & 21.69 & 23.95 & \textbf{69.50} & 91.71 & 44.78 & \textbf{7.00} & \textbf{67.00} & 60.78 & 65.52 & \textbf{45.32}  \\
SKV ($B{=}128$) & 21.02 & 35.77 & 45.71 & 44.71 & 42.94 & 23.49 & 21.65 & 19.71 & 18.49 & 57.50 & 86.67 & 41.90 & 7.00 & 64.00 & 53.91 & 55.24 & 39.98  \\
\hi SKV+\dg$_{2}$ ($B{=}128$) & 22.61 & 43.93 & 51.24 & \textbf{50.38} & 47.90 & 25.96 & \textbf{31.86} & 21.51 & 24.15 & \textbf{69.50} & 91.24 & \textbf{45.18} & \textbf{7.00} & 66.50 & 60.65 & 65.59 & \textbf{45.32}  \\
PKV+\dg$_{2}$ ($B{=}128$) & 22.10 & \textbf{44.30} & 50.17 & 50.12 & 47.49 & 24.91 & 31.11 & \textbf{21.91} & 23.79 & \textbf{69.50} & 91.19 & \textbf{45.18} & \textbf{7.00} & 66.50 & 59.64 & 65.64 & 45.03  \\
H2O+\dg$_{2}$ ($B{=}128$) & 21.96 & 43.05 & 49.37 & 48.90 & 46.83 & 25.47 & 30.88 & 21.85 & 24.20 & \textbf{69.50} & \textbf{91.92} & 43.88 & \textbf{7.00} & 65.50 & 59.66 & 64.82 & 44.67  \\
SLM+\dg$_{2}$ ($B{=}128$) & \textbf{24.21} & 43.72 & 48.88 & 49.45 & 46.42 & \textbf{26.05} & 30.95 & 21.29 & \textbf{24.52} & \textbf{69.50} & 91.45 & 43.34 & \textbf{7.00} & 64.00 & \textbf{61.47} & 65.46 & 44.86  \\
\midrule
\multicolumn{18}{c}{\emph{Qwen2.5-14B-Instruct}} \\
\midrule
FKV & 27.03 & 45.10 & 51.82 & 52.24 & 57.59 & 30.10 & 29.07 & 21.92 & 22.78 & 75.00 & 89.75 & 46.99 & 7.04 & 67.00 & 62.90 & 50.40 & 46.05  \\
\cmidrule(lr){1-18}
SKV ($B{=}32$) & 17.48 & 18.25 & 27.87 & 36.98 & 28.21 & 21.59 & 14.15 & 16.49 & 12.48 & 36.50 & 79.13 & 35.07 & 10.16 & 40.38 & 48.29 & 36.03 & 29.94  \\
\hi SKV+\dg$_{2}$ ($B{=}32$) & 25.55 & 44.66 & 51.00 & 51.18 & 57.15 & 30.24 & 29.08 & 21.96 & 22.81 & \textbf{75.00} & 90.73 & 46.96 & 8.20 & 62.00 & 63.06 & 52.31 & 45.74  \\
SKV+\dg{}-W$_{2}$ ($B{=}32$) & \textbf{27.29} & 44.90 & 51.17 & 51.46 & 57.70 & 30.06 & 29.34 & 21.90 & 22.65 & \textbf{75.00} & \textbf{91.28} & 45.79 & \textbf{8.93} & 66.00 & 63.67 & 51.71 & 46.18  \\
SKV ($B{=}64$) & 21.59 & 27.85 & 38.97 & 44.92 & 40.97 & 26.15 & 17.48 & 18.55 & 15.88 & 49.00 & 84.48 & 41.57 & 9.68 & 50.08 & 54.32 & 41.21 & 36.42  \\
\hi SKV+\dg$_{2}$ ($B{=}64$) & 25.57 & 43.80 & 50.82 & \textbf{51.77} & 57.07 & 30.24 & \textbf{29.48} & 22.27 & \textbf{22.89} & \textbf{75.00} & 90.43 & 46.94 & 8.23 & 65.25 & 63.48 & 51.80 & 45.94  \\
SKV+\dg{}-W$_{2}$ ($B{=}64$) & 26.38 & \textbf{45.64} & \textbf{51.37} & 51.12 & \textbf{57.81} & 30.11 & 29.36 & 21.72 & 22.62 & \textbf{75.00} & 90.82 & 46.14 & 8.70 & \textbf{67.00} & 64.54 & 52.56 & \textbf{46.31}  \\
SKV ($B{=}128$) & 20.98 & 30.87 & 41.50 & 48.77 & 51.74 & 28.03 & 19.87 & 19.88 & 17.86 & 64.50 & 86.75 & 43.18 & 9.00 & 63.75 & 56.52 & 44.95 & 40.51  \\
\hi SKV+\dg$_{2}$ ($B{=}128$) & 25.86 & 44.79 & 50.77 & 51.59 & 57.26 & \textbf{30.64} & 29.43 & \textbf{22.66} & 22.64 & \textbf{75.00} & 90.48 & \textbf{47.51} & 8.71 & 65.75 & \textbf{64.76} & 52.52 & 46.27  \\
SKV+\dg{}-W$_{2}$ ($B{=}128$) & 25.91 & 45.42 & 50.28 & 51.45 & 57.78 & 30.38 & 29.36 & 21.83 & 22.49 & \textbf{75.00} & 90.17 & 46.90 & 8.39 & 66.50 & 64.03 & \textbf{52.94} & 46.18  \\
\midrule
\multicolumn{18}{c}{\emph{Llama-3.2-3B-Instruct}} \\
\midrule
FKV & 21.50 & 40.41 & 50.26 & 44.97 & 39.07 & 18.09 & 32.32 & 23.52 & 25.83 & 69.50 & 88.22 & 42.81 & 4.50 & 67.50 & 54.36 & 56.41 & 42.45  \\
\cmidrule(lr){1-18}
SKV ($B{=}128$) & 16.65 & 24.91 & 46.11 & 43.99 & 36.51 & 15.77 & 22.03 & 20.52 & 20.02 & 65.00 & 88.01 & 38.22 & 4.50 & 66.50 & 50.63 & 49.77 & 38.07  \\
\hi SKV+\dg$_{2}$ ($B{=}128$) & \textbf{22.19} & \textbf{40.19} & \textbf{51.60} & \textbf{45.46} & \textbf{38.87} & 18.34 & \textbf{31.56} & 22.48 & \textbf{25.18} & \textbf{69.50} & \textbf{88.26} & \textbf{43.09} & \textbf{4.50} & \textbf{66.50} & \textbf{55.99} & 57.31 & \textbf{42.56}  \\
SKV ($B{=}64$) & 18.08 & 24.16 & 41.94 & 43.44 & 31.79 & 15.06 & 20.21 & 20.27 & 18.12 & 56.50 & 85.13 & 36.35 & 4.50 & 61.00 & 48.26 & 45.45 & 35.64  \\
\hi SKV+\dg$_{2}$ ($B{=}64$) & 22.13 & 39.83 & 51.11 & 45.29 & 38.79 & \textbf{18.36} & 31.34 & \textbf{22.62} & 25.10 & \textbf{69.50} & 87.81 & 42.79 & \textbf{4.50} & 65.50 & 55.75 & \textbf{57.33} & 42.36  \\
\bottomrule
\end{tabular}}
\end{table}

%% file: sections/table_ablation.tex
% SnapKV-variant ablation grid: draft-length dose-response (DGE k=2..32)
% and eviction signal (DGE = draft queries, DGE-W = SnapKV's unmodified
% window scores). FKV/SKV repeated as reference rows. The B=128 block is the
% full master sweep (results_sweep_128); the B=32 block is results_sweep_32.
% Full evaluation sets throughout.
\begin{table}[htbp]
\centering
\caption{\textbf{SnapKV-variant ablation} on Llama-3.1-8B at $B{=}128$
(full dose grid) and $32$ (sampled). \textbf{Bold}: best compression method
per column per budget
block (FullKV excluded). Axes: draft length $k$ and eviction signal (\dg{}:
the $k$ real decode queries; \dg{}-W: \snapkv{}'s unmodified window scores,
i.e.\ timing only). On this backbone both signals are flat from $k{=}2$:
every point is $\ge3.9$ above SKV and within $0.16$ of FullKV. FKV = FullKV, SKV = \snapkv{} \citep{li2024snapkv};
every SKV+\dg{} and SKV+\dg{}-W row is ours.}
\label{tab:ablation}
\vspace{2pt}
\adjustbox{max width=\textwidth}{%
\begin{tabular}{l ccc ccc ccc ccc cc cc c}
\toprule
& \multicolumn{3}{c}{Single-Document QA}
& \multicolumn{3}{c}{Multi-Document QA}
& \multicolumn{3}{c}{Summarization}
& \multicolumn{3}{c}{Few-shot Learning}
& \multicolumn{2}{c}{Synthetic}
& \multicolumn{2}{c}{Code} & \\
\cmidrule(lr){2-4}\cmidrule(lr){5-7}\cmidrule(lr){8-10}\cmidrule(lr){11-13}\cmidrule(lr){14-15}\cmidrule(lr){16-17}
Method
& \rotatebox{35}{NrtvQA} & \rotatebox{35}{Qasper} & \rotatebox{35}{MF-en}
& \rotatebox{35}{HotpotQA} & \rotatebox{35}{2WikiMQA} & \rotatebox{35}{Musique}
& \rotatebox{35}{GovReport} & \rotatebox{35}{QMSum} & \rotatebox{35}{MultiNews}
& \rotatebox{35}{TREC} & \rotatebox{35}{TriviaQA} & \rotatebox{35}{SAMSum}
& \rotatebox{35}{PCount} & \rotatebox{35}{PRe}
& \rotatebox{35}{Lcc} & \rotatebox{35}{RB-P} & \rotatebox{35}{Avg.} \\
\midrule
\multicolumn{18}{c}{\emph{KV Size = Full (budget-independent reference)}} \\
\midrule
FKV & 24.27 & 45.79 & 55.95 & 48.65 & 46.90 & 27.32 & 33.79 & 23.26 & 27.14 & 69.50 & 91.49 & 44.41 & 10.91 & 68.50 & 65.19 & 56.80 & 46.24  \\
\midrule
\multicolumn{18}{c}{\emph{KV Size = 32 (sampled dose grid: $k{=}2,8,32$)}} \\
\midrule
SKV & 20.53 & 25.53 & 41.75 & 43.76 & 34.70 & 24.31 & 17.72 & 19.06 & 16.81 & 41.50 & 80.78 & 34.27 & \textbf{12.50} & 65.00 & 51.63 & 43.76 & 35.85  \\
\cmidrule(lr){1-18}
\hi SKV+\dg$_{2}$ & 24.13 & 45.03 & 54.96 & 48.58 & \textbf{48.08} & 27.50 & \textbf{32.53} & 23.32 & \textbf{26.67} & \textbf{69.50} & \textbf{92.65} & 44.74 & \textbf{12.50} & 68.17 & 65.61 & \textbf{58.34} & 46.39  \\
SKV+\dg$_{8}$ & 24.28 & \textbf{45.80} & 55.23 & \textbf{48.68} & 47.46 & \textbf{27.60} & 31.96 & \textbf{23.49} & 26.42 & \textbf{69.50} & 91.18 & 44.86 & 11.33 & \textbf{68.50} & 65.20 & 56.67 & 46.14  \\
SKV+\dg$_{32}$ & 24.27 & 45.79 & 55.89 & 48.65 & 46.85 & 27.32 & 31.82 & 23.31 & 26.55 & \textbf{69.50} & 91.49 & 44.42 & 10.91 & \textbf{68.50} & 65.26 & 56.82 & 46.08  \\
SKV+\dg{}-W$_{2}$ & \textbf{24.47} & 45.32 & \textbf{56.39} & 48.18 & 47.76 & 27.15 & 31.98 & 23.23 & 26.49 & \textbf{69.50} & 92.53 & \textbf{45.11} & \textbf{12.50} & 68.00 & \textbf{66.21} & 57.55 & \textbf{46.40}  \\
\midrule
\multicolumn{18}{c}{\emph{KV Size = 128 (full dose grid)}} \\
\midrule
SKV & 21.84 & 32.57 & 52.13 & 46.60 & 44.42 & 25.90 & 22.33 & 22.22 & 21.92 & 62.50 & 91.26 & 40.89 & 12.42 & 68.50 & 60.67 & 49.30 & 42.22  \\
\cmidrule(lr){1-18}
\hi SKV+\dg$_{2}$ & 22.97 & 44.45 & 54.49 & \textbf{48.89} & \textbf{48.70} & \textbf{27.91} & 32.55 & 23.73 & 26.36 & \textbf{69.50} & 92.48 & 45.09 & \textbf{12.50} & \textbf{69.00} & 65.78 & \textbf{57.46} & \textbf{46.37}  \\
SKV+\dg$_{4}$ & 23.95 & 44.42 & 55.86 & 48.65 & 48.24 & 27.44 & 32.49 & 23.47 & 26.75 & \textbf{69.50} & 92.29 & 44.97 & 12.17 & 68.50 & 65.22 & 56.77 & 46.29  \\
SKV+\dg$_{8}$ & 24.18 & 45.32 & 55.70 & 48.63 & 47.23 & 27.42 & 32.26 & 23.35 & 26.65 & \textbf{69.50} & 91.25 & 44.68 & 11.33 & 68.50 & 65.19 & 56.73 & 46.12  \\
SKV+\dg$_{16}$ & \textbf{24.31} & 45.76 & 55.63 & 48.72 & 46.86 & 27.34 & \textbf{32.56} & 23.92 & 26.62 & \textbf{69.50} & 91.56 & 44.93 & 10.91 & 68.50 & 65.23 & 56.72 & 46.19  \\
SKV+\dg$_{32}$ & 24.27 & 45.81 & 55.84 & 48.65 & 46.85 & 27.32 & 32.36 & 23.42 & \textbf{26.93} & \textbf{69.50} & 91.49 & 44.28 & 10.91 & 68.50 & 65.13 & 56.82 & 46.13  \\
SKV+\dg{}-W$_{2}$ & 24.12 & 44.30 & 55.79 & 47.97 & 48.27 & 27.69 & 32.37 & 23.70 & 26.35 & \textbf{69.50} & \textbf{92.77} & 44.63 & \textbf{12.50} & 68.00 & \textbf{65.96} & 57.36 & 46.33  \\
SKV+\dg{}-W$_{4}$ & 23.64 & 45.30 & \textbf{56.13} & 48.58 & 47.69 & 27.49 & 32.20 & \textbf{24.04} & 26.05 & \textbf{69.50} & 92.20 & \textbf{45.40} & 12.17 & 68.50 & 65.21 & 56.74 & 46.30  \\
SKV+\dg{}-W$_{8}$ & 24.25 & 46.01 & 55.99 & 48.62 & 47.23 & 27.45 & 31.79 & 23.82 & 26.13 & \textbf{69.50} & 91.28 & 44.95 & 11.17 & 68.50 & 65.14 & 56.42 & 46.14  \\
SKV+\dg{}-W$_{16}$ & 24.29 & \textbf{46.20} & 55.86 & 48.72 & 46.86 & 27.34 & 31.55 & \textbf{24.04} & 26.62 & \textbf{69.50} & 91.48 & 44.80 & 11.21 & 68.50 & 65.04 & 56.69 & 46.17  \\
\bottomrule
\end{tabular}}
\end{table}

%% file: sections/table_lookahead.tex
% Lookahead Q-Cache (LAQ) comparison. Same backbone (Llama-3.1-8B-Instruct),
% same 16 LongBench sets, same truncation (7500) and greedy decoding, run in
% the *original* LAQ code base with our chat template / stop-token protocol
% ported in for a matched comparison. LAQ improves the WHAT axis (pseudo
% lookahead queries -> better importance scores) but still evicts at prefill
% end; DGE-W keeps SnapKV's unmodified window scores (no WHAT change) but
% defers eviction by one decode step. Avg over the 16 sets.
\begin{table}[htbp]
\centering
\caption{\textbf{Lookahead-query eviction vs deferred eviction}
(Llama-3.1-8B, 16 LongBench sets, matched protocol; LAQ run in its own
code). LAQ \citep{ge2025lookahead} sharpens the \emph{what} axis with $m$
pseudo future queries but still evicts at prefill end. \dg{}-W keeps
\snapkv{}'s window scores and only defers eviction one decode step; \dg{}
also swaps in the real decode queries. \textbf{Bold}: best per budget
(FullKV excluded).}
\label{tab:lookahead}
\vspace{2pt}
\adjustbox{max width=0.85\textwidth}{%
\begin{tabular}{l l cc}
\toprule
& Eviction time / signal & $B{=}64$ & $B{=}128$ \\
\midrule
FullKV & --- (budget-independent) & 46.24 & 46.24 \\
\midrule
\snapkv{} \citep{li2024snapkv}
  & prefill end / window & 39.38 & 42.22 \\
\midrule
\multicolumn{4}{l}{\emph{Lookahead Q-Cache} \citep{ge2025lookahead}
  --- prefill end / pseudo-query scores ($m$ lookahead queries)} \\
\quad LAQ ($m{=}4$)   & prefill end / pseudo & 42.75 & 43.81 \\
\quad LAQ ($m{=}8$)   & prefill end / pseudo & 43.11 & 44.13 \\
\quad LAQ ($m{=}16$)  & prefill end / pseudo & 43.17 & 44.20 \\
\quad LAQ ($m{=}32$)  & prefill end / pseudo & --- & 44.10 \\
\midrule
\multicolumn{4}{l}{\emph{Deferred eviction (ours) --- one full-cache decode
  step, then evict}} \\
\quad \snapkv{}+\dg{}-W$_{2}$ & \textbf{decode step 2} / window & 46.20 & 46.33 \\
\quad \textbf{\snapkv{}+\dg$_{2}$} & \textbf{decode step 2} / draft queries
  & \textbf{46.36} & \textbf{46.37} \\
\bottomrule
\end{tabular}}
\end{table}

%% file: sections/table_category.tex
% Per-category LongBench breakdown (Llama-3.1-8B), aggregated from the
% 16-dataset results of tab:main-full into the six standard LongBench groups.
% Appendix conventions: no "(Ours)" tags; \cmidrule separates deferred rows.
\begin{table}[htbp]
\centering
\caption{\textbf{LongBench per-category averages} (Llama-3.1-8B; FullKV is
budget-independent). \snapkv{}'s damage concentrates in Few-shot Learning,
Code, Single-Doc QA and Summarization ($10$--$16$ points at $B{=}32$), while
Synthetic---essentially retrieval, so the window queries see what
matters---already matches FullKV.
One deferred decode step restores \emph{every} category at both budgets; at
$B{=}32$ it matches FullKV category-for-category where \snapkv{} is $10.4$
points down overall and as much as $16.3$ on Few-shot Learning.
\textbf{Bold} marks the best compression method per column within each budget
block. FKV = FullKV, SKV = \snapkv{} \citep{li2024snapkv};
the SKV+\dg{} and SKV+\dg{}-W rows are ours.}
\label{tab:category}
\vspace{2pt}
\setlength{\tabcolsep}{3pt}
\footnotesize
\begin{tabular}{ll cccccc c}
\toprule
& Method & \makecell{Single-Doc\\QA} & \makecell{Multi-Doc\\QA}
& \makecell{Summar-\\ization} & \makecell{Few-shot\\Learning}
& Synthetic & Code & All \\
\midrule
& FKV & 42.00 & 40.96 & 28.06 & 68.47 & 39.70 & 60.99 & 46.24 \\
\midrule
\multirow{2}{*}{$B{=}32$}
& SKV & 29.27 & 34.26 & 17.86 & 52.18 & 38.75 & 47.70 & 35.85 \\
\cmidrule(lr){2-9}
\hi & SKV+\dg$_{2}$ & \textbf{41.37} & \textbf{41.39} & \textbf{27.51} & \textbf{68.96} & \textbf{40.34} & \textbf{61.98} & \textbf{46.39} \\
\midrule
\multirow{3}{*}{$B{=}128$}
& SKV & 35.51 & 38.97 & 22.16 & 64.88 & 40.46 & 54.98 & 42.22 \\
\cmidrule(lr){2-9}
& SKV+\dg{}-W$_{2}$ & \textbf{41.40} & 41.31 & 27.47 & 68.97 & 40.25 & \textbf{61.66} & 46.33 \\
\hi & SKV+\dg$_{2}$ & 40.64 & \textbf{41.83} & \textbf{27.55} & \textbf{69.02} & \textbf{40.75} & 61.62 & \textbf{46.37} \\
\bottomrule
\end{tabular}
\end{table}

%% file: sections/table_term.tex
% Draft-termination accounting: how often the answer finishes inside the k-token
% draft (so DGE never evicts and is bit-identical to FullKV), plus the
% sample-count audit. Measured on the saved SKV+DGE_2 predictions,
% Llama-3.1-8B B=128, by re-tokenizing each answer.
\begin{table}[htbp]
\centering
\caption{\textbf{Draft-termination fractions and sample accounting}
(Llama-3.1-8B, $B{=}128$). \emph{term$_k$} = share of answers finishing within
the $k$-token draft, where \dg{} is bit-identical to FullKV. All 16 sets at
full official sizes.}
\label{tab:term}
\vspace{2pt}
\setlength{\tabcolsep}{4pt}
\small
\adjustbox{max width=\textwidth}{%
\begin{tabular}{ll rrr r}
\toprule
Category & Dataset & $n$ & term$_{k=2}$ & term$_{k=16}$ & \makecell{median answer\\(tokens)} \\
\midrule
\multirow{3}{*}{Single-Doc QA}
& narrativeqa          & 200 & 5.0 & 87.0 & 7 \\
& qasper               & 200 & 12.5 & 48.0 & 17 \\
& multifieldqa\_en     & 150 & 13.3 & 58.7 & 11 \\
\midrule
\multirow{3}{*}{Multi-Doc QA}
& hotpotqa             & 200 & 39.5 & 89.5 & 3 \\
& 2wikimqa             & 200 & 24.0 & 96.0 & 4 \\
& musique              & 200 & 26.0 & 81.5 & 4 \\
\midrule
\multirow{3}{*}{Summarization}
& gov\_report          & 200 & 0.0 & 0.0 & 415 \\
& qmsum                & 200 & 0.0 & 1.0 & 87 \\
& multi\_news          & 200 & 0.0 & 0.0 & 471 \\
\midrule
\multirow{3}{*}{Few-shot}
& trec                 & 200 & 0.0 & 0.0 & 64 \\
& triviaqa             & 200 & 0.0 & 0.5 & 32 \\
& samsum               & 200 & 0.0 & 0.5 & 128 \\
\midrule
\multirow{2}{*}{Synthetic}
& passage\_count       & 200 & 92.5 & 98.0 & 1 \\
& passage\_retrieval\_en & 200 & 0.0 & 91.0 & 3 \\
\midrule
\multirow{2}{*}{Code}
& lcc                  & 500 & 0.0 & 0.0 & 64 \\
& repobench-p          & 500 & 0.0 & 0.0 & 64 \\
\midrule
\multicolumn{2}{l}{\textbf{Overall}} & \textbf{3750} & \textbf{11.2} & \textbf{34.0} & --- \\
\bottomrule
\end{tabular}}
\end{table}